\documentclass{article}
\usepackage{arxiv}
\PassOptionsToPackage{table}{xcolor}
\usepackage[utf8]{inputenc}
\usepackage[T1]{fontenc}
\usepackage[colorlinks=true,linkcolor=blue,citecolor=blue,urlcolor=blue]{hyperref}
\usepackage{url}
\usepackage{booktabs}
\usepackage{array}
\usepackage{enumitem}
\usepackage{amsfonts}
\usepackage{amsmath}
\usepackage{amssymb}
\usepackage{mathtools}
\usepackage{amsthm}
\usepackage{xcolor}
\usepackage{microtype}
\usepackage{graphicx}
\usepackage{subcaption}
\usepackage[authoryear,round]{natbib}
\usepackage{algorithm}
\usepackage{algorithmic}
\usepackage[capitalize,noabbrev]{cleveref}
\usepackage[textsize=tiny]{todonotes}

\usepackage[most]{tcolorbox}
\usepackage{etoolbox}
\usepackage{titlesec}
\definecolor{iapoprimary}{HTML}{1B3A6B}
\definecolor{iapoteal}{HTML}{3B6FD4}
\definecolor{iapotint}{HTML}{EEF3FB}
\definecolor{iapopale}{HTML}{F5F8FD}
\hypersetup{linkcolor=iapoprimary,citecolor=iapoteal,urlcolor=iapoteal}
\titleformat{\section}{\large\bfseries\color{iapoprimary}}{\thesection}{0.6em}{}[{\color{iapoteal}\titlerule[0.9pt]}]
\titleformat{\subsection}{\normalsize\bfseries\color{iapoprimary}}{\thesubsection}{0.6em}{}
\titleformat{\subsubsection}{\normalsize\bfseries\color{iapoprimary}}{\thesubsubsection}{0.6em}{}
\titleformat{\paragraph}[runin]{\normalsize\bfseries\color{iapoteal}}{\theparagraph}{0pt}{}
\titlespacing*{\section}{0pt}{2.4ex plus 0.5ex minus 0.2ex}{1.2ex plus 0.2ex}
\titlespacing*{\subsection}{0pt}{2.0ex plus 0.4ex minus 0.2ex}{0.8ex plus 0.2ex}
\titlespacing*{\paragraph}{0pt}{1.4ex plus 0.4ex minus 0.2ex}{0.8em}
\makeatletter
\renewcommand{\@toptitlebar}{\textcolor{iapoteal}{\hrule height 3pt}\vskip .15in}
\renewcommand{\@bottomtitlebar}{\vskip .15in\textcolor{iapoteal}{\hrule height 1pt}\vskip .1in}
\makeatother
\renewenvironment{abstract}
{\begin{tcolorbox}[enhanced, breakable, colback=iapotint, colframe=iapoprimary, boxrule=0.6pt, arc=3pt,
  left=9pt, right=9pt, top=10pt, bottom=7pt, before skip=6pt, after skip=10pt,
  title=Abstract, fonttitle=\sffamily\bfseries\small, coltitle=white, colbacktitle=iapoprimary,
  attach boxed title to top left={yshift=-2.5mm, xshift=5mm}, boxed title style={arc=2pt, boxrule=0pt}]\small}
{\end{tcolorbox}}
\tcbset{thmstyle/.style={enhanced, breakable, colback=iapopale, colframe=iapoteal, boxrule=0pt, frame hidden,
  arc=2pt, left=8pt, right=8pt, top=3pt, bottom=3pt,
  before skip=7pt, after skip=7pt}}
\newcommand{\hlt}[1]{\textcolor{iapoteal}{\textbf{#1}}}
\BeforeBeginEnvironment{theorem}{\begin{tcolorbox}[thmstyle]}
\AfterEndEnvironment{theorem}{\end{tcolorbox}}
\BeforeBeginEnvironment{proposition}{\begin{tcolorbox}[thmstyle]}
\AfterEndEnvironment{proposition}{\end{tcolorbox}}
\BeforeBeginEnvironment{corollary}{\begin{tcolorbox}[thmstyle]}
\AfterEndEnvironment{corollary}{\end{tcolorbox}}
\theoremstyle{plain}
\newtheorem{theorem}{Theorem}[section]
\newtheorem{proposition}[theorem]{Proposition}
\newtheorem{lemma}[theorem]{Lemma}
\newtheorem{corollary}[theorem]{Corollary}
\theoremstyle{definition}

\newtheorem{assumption}[theorem]{Assumption}
\theoremstyle{remark}
\newtheorem{remark}[theorem]{Remark}

\title{Learning Multi-Agent Orchestration:\\Higher-Order Interactions in Policy Improvement}
\author{
  Kaixuan Liu \\
  Department of Computer Science\\
  Emory University\\
  Atlanta, GA 30322 \\
  \texttt{kaixuan.liu@emory.edu} \\
  \And
  Guojun Xiong\thanks{Corresponding author.} \\
  John Hopcroft Center for Computer Science\\
  Shanghai Jiao Tong University\\
  Shanghai, China \\
  \texttt{gjxiong@sjtu.edu.cn} \\
}

\begin{document}
\maketitle
\begin{abstract}
Learned orchestration has emerged as a promising paradigm for large language
model (LLM)-based multi-agent systems, in which a central orchestrator
selects agents and controls the information they receive throughout a task.
Despite rapid empirical progress, existing methods train the orchestrator by
crediting each decision with its own advantage, implicitly assuming that the
other decisions in the episode stay fixed. This assumption fails in orchestration, where the contribution of each decision
to the final outcome depends on the decisions preceding it, so the gain of a
joint update is not the sum of its per-decision effects. To address this challenge, we introduce \textbf{IAPO}, an interaction-aware
policy optimization framework that explicitly accounts for interactions among
decisions. Specifically, IAPO \hlt{exactly decomposes the performance difference between two
policies into the effects of individual decisions and their interactions},
\hlt{which we prove can account for half of the attainable gain}, and then optimizes
both from sampled trajectories while \hlt{certifying that every update is an
improvement}. On long-input versions of GSM8K, MATH, DROP, MBPP and HumanEval, against fifteen published multi-agent and workflow-search systems on the same 7B executor, IAPO's deployed plans \hlt{reach the strongest system's accuracy with fewer tokens}, and \hlt{removing the interaction term costs 7 to 17 points of accuracy on the two code benchmarks}.
\par\smallskip\noindent{\textbf{\textcolor{iapoprimary}{Keywords:}} multi-agent orchestration, policy improvement, interaction effects, certified updates, LLM agents}
\end{abstract}

\section{Introduction}
\label{sec:intro}

Large language model (LLM)-based agents have achieved strong results in
multi-step reasoning \citep{yao2022react}, code generation
\citep{shinn2023reflexion}, and tool use \citep{schick2023toolformer}.
Organizing several such agents into a multi-agent system extends these
capabilities further, as agents that debate \citep{du2023improving}, role-play
\citep{li2023camel}, or converse \citep{wu2023autogen} often outperform each of
their members on complex tasks. This benefit, however, depends on
how the collaboration is organized: which agent acts next, which artifacts it
may read, and when the process terminates \citep{zhang2026reinforcement}. What an agent reads also determines how much it influences the others, and
hence what a population of interacting agents comes to believe collectively
\citep{liu2026social}.

\begin{figure*}[t]
\centering
\includegraphics[width=\textwidth]{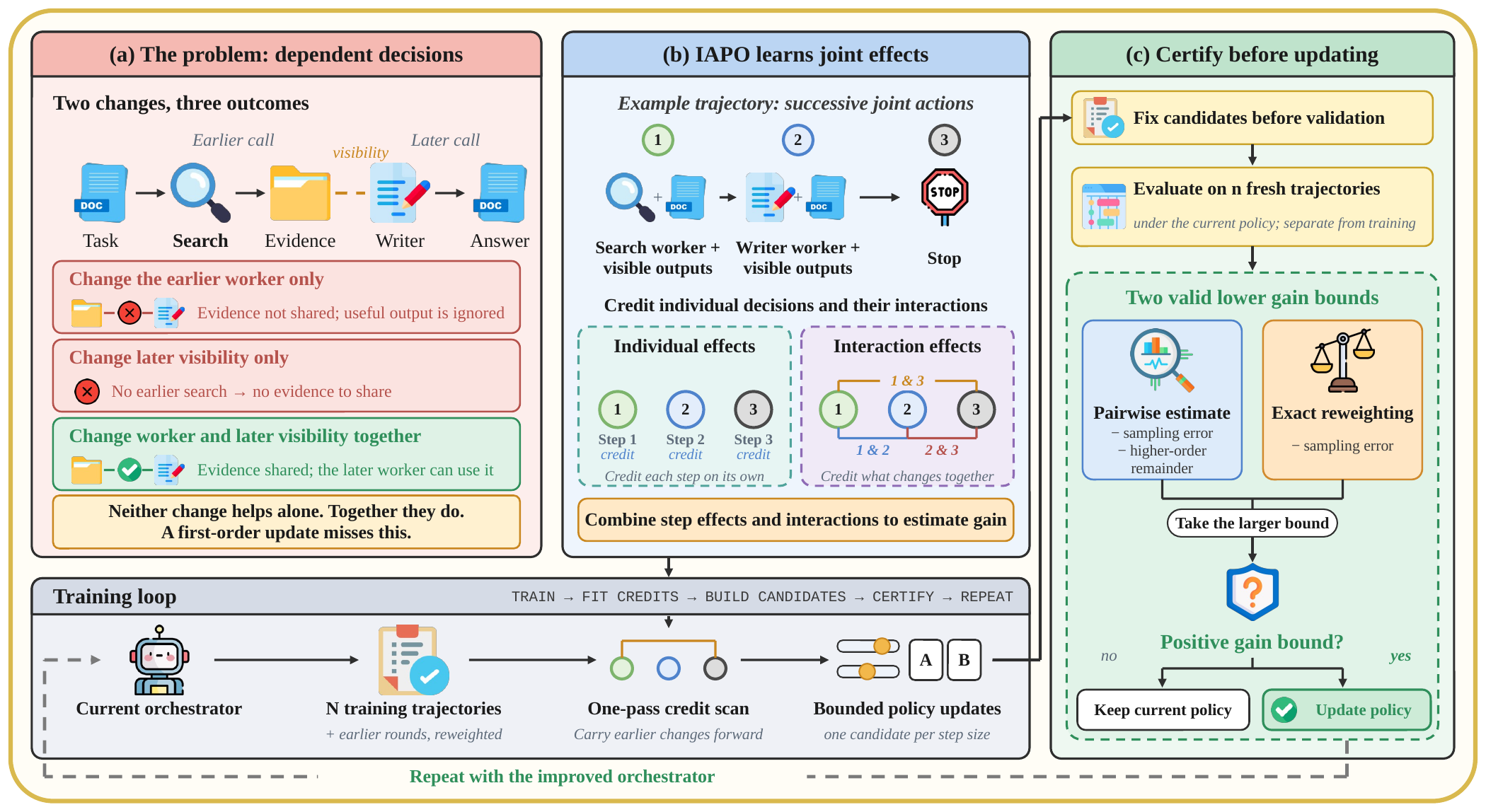}
\caption{Overview of IAPO. (a) Orchestration decisions depend on one
another: a new earlier worker helps only if a later call may read its output,
so neither change helps alone. (b) IAPO credits each orchestrator step and
every pair of steps in one pass over each training trajectory, and fits one
bounded update per step size. (c) With the candidates fixed, IAPO computes two
lower bounds on the gain from fresh trajectories, keeps the larger one, and
updates the policy only if it is positive.}
\label{fig:overview}
\end{figure*}

Early multi-agent systems specify this organization manually through fixed role
assignments and communication pipelines \citep{hong2024metagpt,wu2023autogen},
which limits their transfer across tasks and domains. Subsequent work automates
parts of the design: GPTSwarm \citep{zhuge2024language} and G-Designer
\citep{zhang2024g} optimize communication topologies, MasRouter
\citep{yue2025masrouter} learns to route queries among agents, and ADAS
\citep{hu2025automated}, AFlow \citep{zhang2025aflow}, and MaAS
\citep{zhang2025multi} search over entire workflows. Nevertheless, these methods
largely fix the organization before execution and cannot adapt it to the
intermediate results produced during a task. More recently, orchestration has
been cast as a sequential decision problem, in which a central orchestrator
observes the evolving state of a task and is trained by reinforcement learning
(RL) to decide how to proceed at every step
\citep{dang2026multi,nielsen2026learning}. The same principle has been extended
to training the coordinated agents jointly with the orchestrator
\citep{liu2026llm,zhao2026stronger}, producing systems that surpass the
strongest individual agents at reduced inference cost
\citep{dang2026multi,nielsen2026learning}.

Although this paradigm has shown clear empirical gains, its training
objectives are inherited directly from single-agent policy optimization
\citep{schulman2017proximal,shao2024deepseekmath}, which credits each decision
with its own advantage while treating all other decisions in the episode as
fixed. This approximation is justified by trust-region arguments that bound its error
through the per-step policy divergence
\citep{kakade2002approximately,schulman2015trust}, and multi-agent extensions
control the error in the same per-step manner
\citep{kuba2021trust,zhong2024heterogeneous}. Orchestration violates the
premise of these arguments in two ways. First, the decisions within an episode
depend on one another, since the effect of each decision on the outcome
changes with the decisions made before it. A joint change to several
decisions can therefore improve performance even when every single change
degrades it, and an objective that adds up per-decision credits cannot detect
such joint improvements. Second, an orchestrator makes many decisions in one
episode. Even if each decision changes only slightly, the errors add up across
decisions, so a per-step trust region no longer keeps the
approximation accurate over the length of a whole episode.

To address these challenges, we develop an interaction theory of policy
improvement for orchestration and propose \textbf{IAPO} (Interaction-Aware
Policy Optimization). We show that the performance difference between two
policies is a finite sum of interaction terms among decisions. Each term can be
estimated from trajectories without a critic, and the existing objective is the
first-order term. This objective has the correct gradient but misses all
interactions between organizational decisions, such as which worker to call
and which outputs it may read, so any update rule that treats these decisions
separately can lose half of the possible gain. We further bound each
interaction term by how often the decisions involved are made, which implies
that a trust region should limit the policy change weighted by visit counts. Based on
this theory, IAPO optimizes the first- and second-order terms in one pass over
each trajectory and accepts an update only if a high-confidence lower bound on
its gain is positive. We prove that every accepted update improves performance
with high probability. On controlled orchestration tasks, IAPO reaches the best
possible gain on all three tasks, while the existing objective fails on two,
and IAPO never accepts a harmful update. On long-input versions of five benchmarks, against fifteen published multi-agent and workflow-search systems on the same 7B executor, IAPO's deployed plans reach the strongest system's accuracy with fewer tokens in every environment and, at that system's token budget, exceed it significantly on GSM8K, MATH and DROP under at least one cost weight; removing the interaction term costs 7 to 17 points of accuracy on code.

\begin{itemize}
\item \textbf{Exact Interaction Theory:} We derive an exact, finite expansion
of the performance difference between two orchestration policies, which
identifies existing objectives as its single-decision terms and characterizes
the omitted interactions through their curvature, visit-count bounds, and
sample complexity.
\item \textbf{Practical Algorithm:} We propose IAPO, which jointly optimizes
the single-decision and pairwise terms in one pass per trajectory and accepts
an update only when its gain is certified with high confidence. Its
analysis extends to jointly trained agents.
\item \textbf{Theoretical and Empirical Validation:} We prove that ignoring
interactions can forfeit half of the attainable gain and that IAPO accepts
only improving updates with high probability, and we confirm on controlled
tasks that IAPO recovers gains missed by existing objectives without accepting
harmful updates, and on five LLM orchestration benchmarks that it matches or exceeds the strongest of fifteen baselines at lower cost.
\end{itemize}
\section{Problem Formulation}
\label{sec:formulation}
In this section, we formulate multi-agent orchestration as an episodic MDP and
define our optimization objective.

\subsection{Multi-Agent Orchestration}
\label{sec:orchestration}

Let $\mathcal{Q}$ be a finite set of tasks and $\mathcal{Y}$ a finite set of
outputs, such as text responses. A task $q\in\mathcal{Q}$ is drawn from a
distribution $\mathcal{D}$ and solved by the agents
$\mathcal{N}=\{0,1,\ldots,M\}$, where agent $0$ is the orchestrator and agents
$j\in[M]$ are workers, such as LLMs, tools, and verifiers. The orchestrator
solves the task through a sequence of worker calls. Let
$y_1,\dots,y_k\in\mathcal{Y}$ be the outputs of the first $k$ calls. In the
$(k+1)$-th call, the orchestrator chooses a worker $j\in[M]$ and an input set
$\mathcal{E}\subseteq\{y_1,\dots,y_k\}$ of earlier outputs, and worker $j$
reads task $q$ and $\mathcal{E}$ and returns a new output
$y_{k+1}\in\mathcal{Y}$. Each call to worker $j$ costs $c_j\ge0$, and at most
$K$ calls are allowed. Instead of calling a worker, the orchestrator may stop,
and the task and the outputs collected so far are scored by a task reward
$R\in[0,1]$.

\subsection{Orchestration MDP}
\label{sec:mdp}

We model multi-agent orchestration as an episodic MDP
$\mathcal{M}(\mathcal{S},\mathcal{A},H,\mathbb{P},r,d_0)$, where
$\mathcal{S}$ is the state space, $\mathcal{A}$ is the action space, $H$ is
the number of steps in each episode, $\mathbb{P}=\{P_t\}_{t=1}^H$ with
$P_t(\cdot\mid s,a)$ the distribution of the next state when action $a$ is
taken at state $s$ in step $t$, $r=\{r_t\}_{t=1}^H$ with
$r_t:\mathcal{S}\times\mathcal{A}\to\mathbb{R}$ the reward function at step
$t$, and $d_0=\mathcal{D}$ the initial state distribution.

\paragraph{Episode dynamics.}
An episode starts from the task, $s_1=q\sim\mathcal{D}$, and each agent
$i\in\mathcal{N}$ selects actions with its own policy $\pi^i$. After $k$
calls, the interaction history
$h_k=\big((j_1,\mathcal{E}_1,y_1),\dots,(j_k,\mathcal{E}_k,y_k)\big)$ records
each call as a triple: in the $\ell$-th call, worker $j_\ell\in[M]$ receives
the set $\mathcal{E}_\ell\subseteq\mathcal{Y}_{\ell-1}$ of earlier outputs and
returns $y_\ell\in\mathcal{Y}$, where
$\mathcal{Y}_k=\{y_1,\dots,y_k\}$. Each call takes two steps.
\begin{enumerate}
\item \textbf{Orchestrator step.} At state $s_t=(q,h_k)$, the orchestrator
acts, $i_t=0$. It decides whether to continue, and if so, which worker
$j\in[M]$ to call and which outputs $\mathcal{E}\subseteq\mathcal{Y}_k$ of
earlier workers to give it. Formally, it samples
$a_t\sim\pi^0(\cdot\mid s_t)$ from:
\begin{equation*}
\mathcal{A}(s_t)=\{(j,\mathcal{E}):j\in[M],\,\mathcal{E}\subseteq\mathcal{Y}_k\}\cup\{\mathsf{stop}\}.
\end{equation*}
A call $(j,\mathcal{E})$ receives reward $r_t=-\lambda c_j$ with cost
coefficient $\lambda\ge0$ and moves the state to
$s_{t+1}=(q,h_k,j,\mathcal{E})$, and $\mathsf{stop}$ receives the task reward
$r_t=R(q,h_k)$ and ends the episode.

\item \textbf{Worker step.} Worker $j$ then receives the call from the
orchestrator at state $s_{t+1}=(q,h_k,j,\mathcal{E})$, so $i_{t+1}=j$. It
reads the task $q$ and the outputs in $\mathcal{E}$ and produces a new output. Formally, it samples
$a_{t+1}=y\sim\pi^j(\cdot\mid q,\mathcal{E})$ from $\mathcal{Y}$ and receives
$r_{t+1}=0$. The output is added to the history,
$h_{k+1}=(h_k,(j,\mathcal{E},y))$, and control returns to the orchestrator at
state $s_{t+2}=(q,h_{k+1})$.
\end{enumerate}
The joint policy is therefore $\pi_t(a\mid s_t)=\pi^{i_t}(a\mid s_t)$, and
every trajectory $(s_1,a_1,\dots,s_H,a_H)$ is generated by $\pi$.

\paragraph{Policies and values.}
A policy $\pi=\{\pi_t\}_{t=1}^H$ maps each state $s$ to a distribution
$\pi_t(\cdot\mid s)$ over $\mathcal{A}(s)$. The value and action-value
functions of $\pi$ are:
\begin{align*}
V_t^\pi(s)&=\mathbb{E}_\pi\Big[\sum_{t'=t}^Hr_{t'}(s_{t'},a_{t'})\,\Big|\,s_t=s\Big],\\
Q_t^\pi(s,a)&=\mathbb{E}_\pi\Big[\sum_{t'=t}^Hr_{t'}(s_{t'},a_{t'})\,\Big|\,s_t=s,\,a_t=a\Big],
\end{align*}
and the advantage function is $A_t^\pi=Q_t^\pi-V_t^\pi$. The return-to-go of a
trajectory from step $t$ is $G_t=\sum_{t'=t}^Hr_{t'}(s_{t'},a_{t'})$.

\subsection{Learning Objective}
\label{sec:objective}

Our goal is to solve each task with high reward at low cost. We keep the worker
policies $\pi^1,\dots,\pi^M$ fixed and optimize the orchestrator policy
$\pi^0$ to maximize the expected task reward minus the cost of calls:
\begin{equation}
\max_{\pi^0}J(\pi):=
\mathbb{E}_\pi\Big[R(q,h)-\lambda\sum_{(j,\mathcal{E},y)\in h}c_j\Big],
\label{eq:obj}
\end{equation}
where $h$ is the interaction history when the episode ends. By the rewards
defined in Section~\ref{sec:mdp}, $J(\pi)=\mathbb{E}_\pi[\sum_{t=1}^H
r_t(s_t,a_t)]$ is the expected return of the MDP, so \eqref{eq:obj} is a
standard RL problem in $\pi^0$. Our analysis also covers the case where some
workers are trained.
\section{IAPO}
\label{sec:methods}

To improve the reward--cost objective~\eqref{eq:obj}, IAPO
constructs policy updates from an exact interaction expansion
(Section~\ref{sec:expansion}), controls their approximation error
(Section~\ref{sec:algorithm}), and certifies their performance
using independent data (Section~\ref{sec:certificate}).
Figure~\ref{fig:overview} gives an overview.

\subsection{An Interaction View of Policy Improvement}
\label{sec:expansion}

An orchestrator decision affects subsequent worker outputs,
actions, and call costs. To analyze the resulting interactions,
let $\pi$ be the reference policy and $\pi'$ a candidate that
updates only the orchestrator. We impose the following condition
to express $J(\pi')-J(\pi)$ using trajectories from $\pi$.

\begin{assumption}[Overlap]
\label{ass:overlap}
There exists $W\ge1$ such that
$\pi'_t(a\mid s)\le W\pi_t(a\mid s)$ for every
$t\in[H]$, every state $s$ reachable under $\pi$,
and every $a\in\mathcal A(s)$.
\end{assumption}

For a trajectory sampled from $\pi$, define the
likelihood ratio
$\rho_t=\pi'_t(a_t\mid s_t)/\pi_t(a_t\mid s_t)$.
Since worker policies and transitions are fixed, their ratios
are $1$, and the likelihood ratio of a trajectory prefix
through $t$ is $\prod_{t'=1}^t\rho_{t'}$. Importance sampling gives
the exact improvement in $J$:
\begin{equation}
J(\pi')-J(\pi)
=\mathbb E_\pi\Big[
\sum_{t=1}^H
\Big(\prod_{t'=1}^t\rho_{t'}-1\Big)r_t(s_t,a_t)
\Big].
\label{eq:is}
\end{equation}

Expanding the likelihood-ratio product in \eqref{eq:is},
with all summation indices ranging over $[t]$, gives:
\begin{equation*}
\begin{aligned}
\prod_{t'=1}^{t}\rho_{t'}-1
&=\sum_{t_1}(\rho_{t_1}-1)
+\sum_{t_1<t_2}(\rho_{t_1}-1)(\rho_{t_2}-1)\\
&\quad+\sum_{t_1<t_2<t_3}(\rho_{t_1}-1)(\rho_{t_2}-1)(\rho_{t_3}-1)\\
&\quad+\sum_{m=4}^{t}\sum_{t_1<\cdots<t_m}
(\rho_{t_1}-1)\cdots(\rho_{t_m}-1).
\end{aligned}
\end{equation*}
Each factor $\rho_{t'}-1$ represents a decision-level policy
change, while products capture interactions across decisions.
Rearranging \eqref{eq:is} by interaction order and replacing
returns by advantages yields the exact decomposition below.

\begin{theorem}[Interaction expansion]
\label{thm:expansion}
Under Assumption~\ref{ass:overlap}:
\begin{equation}
J(\pi')-J(\pi)=\sum_{m=1}^H J^{(m)},
\end{equation}
where the order-$m$ interaction is:
\begin{equation}
\begin{aligned}
J^{(m)}=\mathbb E_\pi\Big[&\sum_{1\le t_1<\cdots<t_m\le H}
(\rho_{t_1}-1)\cdots(\rho_{t_m}-1)\\
&\quad\quad\quad\quad\quad\quad\quad\quad\times A_{t_m}^\pi(s_{t_m},a_{t_m})\Big].
\end{aligned}
\label{eq:order-m}
\end{equation}
Moreover, each $J^{(m)}$ remains unchanged
if $A_{t_m}^\pi$ is replaced by
$G_{t_m}-b(s_{t_m})$ for any bounded function $b$.
\end{theorem}

The first-order term $J^{(1)}$ recovers the
advantage-weighted objective underlying PPO-style optimization
\citep{schulman2017proximal}, and higher orders capture the effects
of changing several decisions along the same trajectory.

Frozen workers contribute no likelihood-ratio changes,
but repeated orchestrator decisions still interact.
Their contribution appears in policy curvature.

\begin{corollary}[Interaction-induced curvature]
\label{cor:curvature}
Let $\pi_\theta$ be a twice continuously differentiable
orchestrator policy, with $\pi_{\theta^0}=\pi$ having
full support and all worker policies fixed.
Let $J^{(m)}(\theta)$ denote the interaction terms
for $\pi'=\pi_\theta$. At $\theta=\theta^0$:
\begin{align*}
&\nabla_\theta J(\pi_\theta)
=\nabla_\theta J^{(1)}(\theta),\\
&\nabla_\theta^2 J(\pi_\theta)
-\nabla_\theta^2 J^{(1)}(\theta)
=\nabla_\theta^2 J^{(2)}(\theta).
\end{align*}
\end{corollary}

Thus, the first-order objective has the correct local gradient
but misses interaction-induced curvature that can affect the
quality of finite policy updates.

\subsection{Interaction-Aware Policy Optimization}
\label{sec:algorithm}
\label{sec:visit}

\paragraph{Second-order truncation.}
To obtain a tractable update, we retain the first two orders
and bound the remainder. With only the orchestrator updated, define:
\[
\beta=2(K+1)\max_s
\mathrm{TV}\big(\pi'^0(\cdot\mid s),\pi^0(\cdot\mid s)\big),
\]
which upper-bounds the cumulative magnitude of policy changes
across repeated orchestrator decisions.
Let $\bar G=1+\lambda K\max_jc_j$ bound the total
absolute reward of a trajectory.

\begin{theorem}[Visit-count bound]
\label{thm:visit}
Under Assumption~\ref{ass:overlap}, with workers fixed,
for every integer $r\ge1$:
\begin{equation}
\Big|J(\pi')-J(\pi)-\sum_{m=1}^rJ^{(m)}\Big|
\le\bar G\sum_{m>r}\frac{\beta^m}{m!}.
\label{eq:tail}
\end{equation}
\end{theorem}

Taking $r=2$ yields:
\begin{equation}
J(\pi')-J(\pi)\ge J^{(1)}+J^{(2)}
-\bar G\Big(e^\beta-1-\beta-\frac{\beta^2}{2}\Big).
\label{eq:lower-bound}
\end{equation}

First-order truncation incurs $O(\bar G\beta^2)$ error, and
second-order truncation reduces this to $O(\bar G\beta^3)$.
We therefore maximize $J^{(1)}+J^{(2)}$ while controlling
$\beta$, following trust-region policy
improvement \citep{schulman2015trust}.

\paragraph{Bounded policy update.}
Clipping the objective does not bound the policy itself.
We instead enforce the visit-count constraint by construction.

For a learned score $f_\theta(s,a)\in[-1,1]$,
define its reference-policy-centered form:
\[
u_\theta(s,a)=\frac12\Big(
f_\theta(s,a)
-\sum_{a'}\pi(a'\mid s)f_\theta(s,a')
\Big).
\]
This gives $\sum_a\pi(a\mid s)u_\theta(s,a)=0$ and
$|u_\theta(s,a)|\le1$. For $\eta\in[0,1]$, let:
\begin{equation}
\pi'_{\theta,\eta}(a\mid s)
=\pi(a\mid s)\big(1+\eta u_\theta(s,a)\big).
\label{eq:bounded-policy}
\end{equation}
Only the orchestrator policy is updated,
and every worker policy remains fixed.

\begin{proposition}[Bounded ratios]
\label{prop:bounded}
For any $\theta$ and $\eta\in[0,1]$,
\eqref{eq:bounded-policy} defines a valid
action distribution at every orchestrator
state $s$, satisfying
$\pi'_{\theta,\eta}(a\mid s)\ge\pi(a\mid s)^2$.
Moreover, on the support of $\pi$, its
likelihood ratio satisfies
$|\rho_t-1|\le\eta$, and its total variation
distance from $\pi(\cdot\mid s)$ is at most
$\eta/4$.
\end{proposition}

Thus, overlap holds with $W=1+\eta$ and,
with at most $K+1$ orchestrator decisions,
$\beta\le(K+1)\eta/2$.

\begin{algorithm}[t]
\caption{Interaction-Aware Policy Optimization}
\label{alg:iapo}
\begin{algorithmic}[1]
\REQUIRE Initial policy $\pi$, score family
$f_\theta$, grid $\Lambda$, batch sizes
$N,n\ge2$, confidence level $\delta$
\FOR{$\kappa=1,2,\ldots$}
\STATE Collect $N$ training trajectories
$\mathcal T$ under $\pi$, and add those of earlier
rounds reweighted to $\pi$
\STATE $b\gets$ mean of $G_1$ over the $N$ new trajectories
\FOR{each $\eta\in\Lambda$}
\STATE Fit $\theta_\eta$ by approximately maximizing
the empirical mean of $Z_2$ with baseline $b$
on $\mathcal T$
\STATE $\pi'_\eta\gets
\pi'_{\theta_\eta,\eta}$
\ENDFOR
\STATE Collect $n$ fresh validation
trajectories $\mathcal V$ under $\pi$
\STATE $\delta_\kappa\gets\delta/2^\kappa$
\FOR{each $\eta\in\Lambda$}
\STATE Compute $L_\eta$ on $\mathcal V$
using \eqref{eq:certificate}, with the
fixed $b$, at level $\delta_\kappa$
\ENDFOR
\STATE $\eta^\star\gets
\arg\max_{\eta\in\Lambda}L_\eta$
\IF{$L_{\eta^\star}\le0$}
\STATE \textbf{return} $\pi$
\ENDIF
\STATE $\pi\gets\pi'_{\eta^\star}$
\ENDFOR
\end{algorithmic}
\end{algorithm}

\paragraph{Interaction-aware objective.}
Let $u_t=u_\theta(s_t,a_t)$ at orchestrator steps
and $0$ at worker steps. Since $\rho_t-1=\eta u_t$,
Theorem~\ref{thm:expansion} gives:
\begin{equation}
J^{(1)}+J^{(2)}
=\eta\mu_1(\theta)+\eta^2\mu_2(\theta),
\label{eq:quadratic}
\end{equation}
where:
\begin{align*}
\mu_1(\theta)
=\mathbb E_\pi\Big[\sum_t u_tG_t\Big],\quad
\mu_2(\theta)
&=\mathbb E_\pi\Big[\sum_{t'<t}u_{t'}u_tG_t\Big].
\end{align*}
Both terms can be estimated from reference-policy trajectories.
We subtract a scalar baseline $b$, estimated from training
data and held fixed during validation. By
Theorem~\ref{thm:expansion}, it does not alter the expectation.
Grouping pairwise terms by their later decision gives:
\begin{equation}
Z_2=\sum_{t=1}^H(\rho_t-1)
\Big(1+\sum_{t'<t}(\rho_{t'}-1)\Big)(G_t-b).
\label{eq:z2}
\end{equation}
For fixed $b$, $\mathbb E_\pi[Z_2]=J^{(1)}+J^{(2)}$.
In particular, the return-to-go $G_t$ includes the terminal
task reward and all subsequent worker-call costs through the
MDP rewards in Section~\ref{sec:mdp}. Thus, optimizing $Z_2$
approximates improvement in the original reward--cost
objective~\eqref{eq:obj}, not a separate task-accuracy objective.
A prefix sum computes $Z_2$ and its gradients in $O(H)$
trajectory operations (Algorithm~\ref{alg:onepass} in Appendix~\ref{app:onepass}).

The bounded update admits the explicit truncation bound:
\begin{equation}
\epsilon(\eta)=\bar G\sum_{m=3}^{K+1}
\binom{K+1}{m}\Big(\frac{\eta}{2}\Big)^m,
\label{eq:remainder}
\end{equation}
so that:
\[
J(\pi'_{\theta,\eta})-J(\pi)
\ge J^{(1)}+J^{(2)}-\epsilon(\eta).
\]
For fixed $\eta$, maximizing the second-order objective
therefore maximizes a valid lower bound on policy improvement.

\begin{theorem}[Existence of improving steps]
\label{thm:improving}
Fix $\theta$ and write
$\mu_1=\mu_1(\theta)$ and
$\mu_2=\mu_2(\theta)$.

\emph{(i) First-order gain.}
If $\mu_1>0$, then any $\eta\in(0,1]$ with
$\eta(|\mu_2|+\epsilon(1))\le\mu_1/2$ guarantees:
\[
J(\pi'_{\theta,\eta})-J(\pi)
\ge\eta\mu_1/2.
\]

\emph{(ii) Second-order gain.}
If $\mu_1=0<\mu_2$, then any $\eta\in(0,1]$
with $\eta\epsilon(1)\le\mu_2/2$ guarantees:
\[
J(\pi'_{\theta,\eta})-J(\pi)
\ge\eta^2\mu_2/2.
\]

Furthermore, if the score class can
represent
$f_\theta(s_t,a)=Q_t^\pi(s_t,a)/\bar G$,
then:
\[
\mu_1=\frac1{2\bar G}
\mathbb E_\pi\Big[
\sum_{t:\,i_t=0}
A_t^\pi(s_t,a_t)^2
\Big],
\]
which is positive whenever the advantage
is nonzero with positive probability
at an orchestrator decision.
\end{theorem}

Pairwise interactions can yield improvement even when
first-order gain vanishes (Appendix~\ref{app:trap}).
Let $\Lambda\subset(0,1]$ be a finite step-size grid.
IAPO optimizes $\theta$ separately for each $\eta\in\Lambda$
using empirical $Z_2$, then independently validates
the resulting candidates. The fit may also reuse the trajectories of
earlier rounds, reweighted to $\pi$ by their known likelihood ratios,
since the certificate uses only fresh validation trajectories.

\subsection{Certified Update Selection}
\label{sec:certificate}

Sampling error and higher-order truncation can make an
apparently promising update harmful. For each
$\eta\in\Lambda$, let $\pi'_\eta$ be the trained candidate.
After fixing all candidates, we draw $n$ independent
validation trajectories under the reference policy $\pi$.

Following high-confidence policy improvement
\citep{thomas2015high}, we use $Z_2$ with truncation error
$\epsilon(\eta)$ and an exact importance-sampling statistic.
Using the same fixed baseline $b$, \eqref{eq:is} gives:
\begin{equation}
Z_{\mathrm{IS}}
=\sum_{t=1}^H
\Big(\prod_{t'\le t}\rho_{t'}-1\Big)
r_t(s_t,a_t)-b\Big(\prod_{t=1}^H\rho_t-1\Big).
\label{eq:zis}
\end{equation}
The baseline correction has zero expectation, so
$\mathbb E_\pi[Z_{\mathrm{IS}}]=J(\pi'_\eta)-J(\pi)$,
though cumulative importance weights may increase its variance, which the validation sample size must absorb.

Lemma~\ref{lem:ranges} bounds both statistics:
\begin{align*}
B_{2,\eta}
&=(\bar G+|b|)\Big((K+1)\eta
+\binom{K+1}{2}\eta^2\Big),\\
B_{\mathrm{IS},\eta}
&=(\bar G+|b|)\big((1+\eta)^{K+1}-1\big).
\end{align*}
Since $b$ is fixed before validation, these bounds apply
conditionally on training. For either statistic $Z$, let
$\bar Z$ and $\hat\sigma_Z$ be its validation mean and sample
standard deviation, and $B_{Z,\eta}$ its absolute bound.
The empirical Bernstein inequality \citep{maurer2009empirical}
gives the confidence radius:
\begin{equation}
\xi_\eta(Z)=\hat\sigma_Z\sqrt{\frac{2\log(8|\Lambda|/\delta)}{n}}+\frac{14B_{Z,\eta}\log(8|\Lambda|/\delta)}{3(n-1)}.
\label{eq:radius}
\end{equation}

We take the larger of the two valid lower confidence bounds:
\begin{equation}
L_\eta=\max\big\{\bar Z_2-\xi_\eta(Z_2)-\epsilon(\eta),\ \bar Z_{\mathrm{IS}}-\xi_\eta(Z_{\mathrm{IS}})\big\}.
\label{eq:certificate}
\end{equation}
The first accounts for truncation error, while the second
uses unbiased importance sampling. Their maximum remains
a lower confidence bound on true policy improvement.

IAPO selects the largest positive certificate, retaining
$\pi$ if all certificates are nonpositive. Let $\hat\pi$
be the selected policy and define:
\[
c_\eta=\min\big\{
\xi_\eta(Z_2)+\epsilon(\eta),
\xi_\eta(Z_{\mathrm{IS}})
\big\}.
\]

\begin{theorem}[Certified selection]
\label{thm:certificate}
With probability at least $1-\delta$,
the following statements hold simultaneously.

\emph{(i) Safe selection.}
For every $\eta\in\Lambda$:
\[
L_\eta\le J(\pi'_\eta)-J(\pi).
\]
Consequently:
\[
J(\hat\pi)\ge J(\pi).
\]

\emph{(ii) Oracle inequality.}
The selected policy satisfies:
\[
\begin{aligned}
J(\hat\pi)-J(\pi)
\ge\max\Big\{0,\,
\max_{\eta\in\Lambda}\big[
J(\pi'_\eta)-J(\pi)-2c_\eta
\big]\Big\}.
\end{aligned}
\]

\emph{(iii) Detection of improvement.}
If some $\eta\in\Lambda$ satisfies
$J(\pi'_\eta)>J(\pi)$ and
\[
n-1>
\frac{256B_{\mathrm{IS},\eta}^2
\log(8|\Lambda|/\delta)}
{[J(\pi'_\eta)-J(\pi)]^2},
\]
then $\hat\pi\neq\pi$.
\end{theorem}

Across successive rounds $\kappa$, assigning level
$\delta/2^\kappa$ ensures joint coverage by a union bound.
Algorithm~\ref{alg:iapo} summarizes the full update procedure,
and the trajectory-level computation is given in
Appendix~\ref{app:onepass}. The betting confidence bound of
\citet{waudbysmith2024estimating} can replace \eqref{eq:radius} (Appendix~\ref{app:betting}), and it admits a per-problem baseline (Appendix~\ref{app:exp-cert-details}). Appendix~\ref{app:select} extends IAPO to
policies far from $\pi$, which gain only through coordinated changes.

\definecolor{ink}{HTML}{3F5F78}\definecolor{acc}{HTML}{1E9E8F}\definecolor{accl}{HTML}{E4F6F2}
\definecolor{upc}{HTML}{1E9E8F}\definecolor{dnc}{HTML}{E07A8E}
\newcommand{\up}[1]{\textcolor{upc}{\scriptsize$\uparrow$#1}}
\newcommand{\dn}[1]{\textcolor{dnc}{\scriptsize$\downarrow$#1}}
\section{Experiments}
\label{sec:experiments}

We evaluate IAPO in two settings. Controlled tasks, whose trajectories can be
enumerated exactly, check the theory (Section~\ref{sec:exp-controlled}).
Orchestration over LLM workers on five long-input benchmarks compares accuracy
and cost with published systems and ablates the interaction term
(Sections~\ref{sec:exp-setup} to~\ref{sec:exp-deployed}).

\subsection{Setup}
\label{sec:exp-setup}

\paragraph{Environments.} We build long-input versions of GSM8K
\citep{cobbe2021training}, MATH \citep{hendrycks2021measuring}, DROP
\citep{dua2019drop}, MBPP \citep{austin2021program} and HumanEval
\citep{chen2021evaluating}. A problem's own unit (its statements, passage or
function specification) is placed among 30 to 90 units of other problems, so one
worker reading everything is not already as good as any team. A document holds
81, 91, 31, 61 and 61 units, and the test splits have 790, 220, 800, 333 and 131
problems (Appendix~\ref{app:envs}). These are our own constructions, not the standard benchmarks, so absolute accuracies are not comparable with published numbers on the originals; every system is evaluated on the same constructed inputs. Every system receives the same question text,
which includes the opening words of the target unit. The baselines also read the
whole document; the orchestrated team reaches it only through an evidence finder,
whose BM25 retriever the baselines lack, or a generalist that reads all of it. All
workers are Qwen2.5-7B-Instruct \citep{qwen2025qwen25} served by vLLM
\citep{kwon2023efficient} at temperature~1, following the protocol of MaAS
\citep{zhang2025multi}.

\paragraph{Orchestration task.} At each of at most five decisions (four on code)
the orchestrator calls one of eight roles (planner, evidence finder, analyst,
second answerer, calculator, checker, selector) or stops; code adds MaAS's
test-and-repair operator. It observes an embedding of the task and a few state
flags, never a score. The reward is accuracy minus $\lambda$ times tokens per
$32{,}768$. We train with $\lambda{=}0.076$, the protocol's value (IAPO), and with
$\lambda{=}1$, which trades accuracy for tokens (IAPO-Lite). Our trajectories are
capped at $32{,}768$ tokens; the baselines, which read the whole document, run
with $131{,}072$ or $393{,}216$ tokens.

\paragraph{Training and certification.} IAPO runs one round from the uniform
policy. It fits one bounded update for each step in $\Lambda{=}\{0.25,0.5,1\}$ on
$6{,}400$ proposal episodes ($200$ training problems, $32$ repeats each), adds
three certified leaps (Appendix~\ref{app:leap}), and certifies every candidate
with betting bounds at total level $\delta{=}0.05$ on $8{,}000$ validation episodes
drawn from the remaining training problems. Test problems are never used for
fitting or selection. The certificate covers net return on the validation
distribution, so test accuracy is an empirical question. Search and training tokens
are excluded from the per-problem costs; one MBPP run uses about $1.7\times10^{8}$
in total (Appendix~\ref{app:exp-cert}). Each configuration is trained twice (three
times for MATH, IAPO), results are averaged over runs, and all rules were fixed
before the runs.

\paragraph{Evaluation.} Each test problem is run four times, and we report accuracy
and tokens per problem for one episode. Since the baselines spend more tokens, we
also evaluate at \emph{matched budget}: a majority vote (on code, MaAS's selector
over programs, with its tokens counted) over the most episodes whose tokens fit the
strongest baseline's. The number of episodes never depends on accuracy. Differences
are paired by problem and called significant when $z>2$ (Appendix~\ref{app:exp-cert}).

\paragraph{Baselines.} We run fifteen published systems on the same executor and
splits, each as released, so none has a retriever. The single-agent group is CoT
\citep{wei2022chain}, ComplexCoT \citep{fu2023complexity} and self-consistency
\citep{wang2023selfconsistency}. The hand-crafted multi-agent group is MultiPersona
\citep{wang2024unleashing}, LLM-Debate \citep{du2023improving}, LLM-Blender
\citep{jiang2023llm}, DyLAN \citep{liu2023dynamic}, AgentVerse
\citep{chen2023agentverse} and MacNet \citep{qian2025scaling}. The automated group is
AutoAgents \citep{chen2024autoagents}, GPTSwarm \citep{zhuge2024language}, ADAS
\citep{hu2025automated}, AFlow \citep{zhang2025aflow}, MAS-Zero \citep{ke2025mas}
and MaAS \citep{zhang2025multi}. Not every system runs on every benchmark, and
Appendix~\ref{app:baselines} lists the setup of each.

\subsection{Accuracy and Cost}
\label{sec:exp-llm}

\begin{table}[t]
\centering\scriptsize
\newcommand{\tk}[1]{\textcolor{tkc}{#1}}
\definecolor{tkc}{HTML}{6B8294}
\caption{Accuracy (\%) with its change from Vanilla, and cost (grey, thousand tokens per problem; one episode for baselines, all voted episodes for IAPO)
with Qwen2.5-7B-Instruct on long-input versions of five benchmarks. \textbf{Bold}: best; \underline{underline}: runner-up.
IAPO and IAPO-Lite (trained with $\lambda{=}1$) vote (on code, select programs) within the best baseline's tokens per column.
Avg.\ covers GSM8K, MATH, MBPP and HumanEval. Two-seed baselines are averaged; ``--'': not run; $^\dagger$better accuracy of the upstream and diagnostic arms.}
\label{tab:main}
\setlength{\tabcolsep}{1.6pt}
\renewcommand{\arraystretch}{1.04}
\resizebox{\textwidth}{!}{%
\begin{tabular}{l>{\hspace{7pt}}rl>{\hspace{3pt}}r>{\hspace{7pt}}rl>{\hspace{3pt}}r>{\hspace{7pt}}rl>{\hspace{3pt}}r>{\hspace{7pt}}rl>{\hspace{3pt}}r>{\hspace{7pt}}rl>{\hspace{3pt}}r>{\hspace{10pt}}r>{\hspace{3pt}}r}
\toprule
\textcolor{ink}{\textbf{Method}} & \multicolumn{3}{c}{\textcolor{ink}{\textbf{GSM8K}}} & \multicolumn{3}{c}{\textcolor{ink}{\textbf{MATH}}} & \multicolumn{3}{c}{\textcolor{ink}{\textbf{DROP}}} & \multicolumn{3}{c}{\textcolor{ink}{\textbf{MBPP}}} & \multicolumn{3}{c}{\textcolor{ink}{\textbf{HumanEval}}} & \multicolumn{2}{c}{\textcolor{ink}{\textbf{Avg.}}} \\
\cmidrule(lr){2-4}\cmidrule(lr){5-7}\cmidrule(lr){8-10}\cmidrule(lr){11-13}\cmidrule(lr){14-16}\cmidrule(lr){17-18}
 & \multicolumn{2}{c}{Acc.} & Tok. & \multicolumn{2}{c}{Acc.} & Tok. & \multicolumn{2}{c}{Acc.} & Tok. & \multicolumn{2}{c}{Acc.} & Tok. & \multicolumn{2}{c}{Acc.} & Tok. & Acc. & Tok. \\
\midrule
\multicolumn{18}{l}{\textsc{\textcolor{ink}{Single-agent}}} \\[0pt]
\hspace{2mm}Vanilla & 61.9 &  & \tk{3.8} & 36.7 &  & \tk{5.5} & 42.6 &  & \tk{9.1} & 67.5 &  & \tk{7.7} & 72.9 &  & \tk{8.8} & 59.8 & \tk{6.5} \\
\hspace{2mm}CoT \citep{wei2022chain} & 67.1 & \up{5.2} & \tk{4.4} & 16.4 & \dn{20.3} & \tk{6.1} & 64.3 & \up{21.7} & \tk{9.3} & 67.0 & \dn{0.5} & \tk{8.1} & 71.8 & \dn{1.1} & \tk{9.1} & 55.6 & \tk{6.9} \\
\hspace{2mm}ComplexCoT \citep{fu2023complexity} & 45.3 & \dn{16.6} & \tk{6.7} & 34.4 & \dn{2.3} & \tk{12.9} & -- &  & -- & 69.2 & \up{1.7} & \tk{9.5} & 73.1 & \up{0.2} & \tk{11.9} & 55.5 & \tk{10.2} \\
\hspace{2mm}SC (CoT$\times$5) \citep{wang2023selfconsistency} & 74.2 & \up{12.3} & \tk{26.9} & 19.8 & \dn{16.9} & \tk{39.3} & 67.4 & \up{24.8} & \tk{56.5} & 71.8 & \up{4.3} & \tk{48.5} & 73.3 & \up{0.4} & \tk{55.2} & 59.8 & \tk{42.5} \\
\noalign{\vskip 2pt}
\multicolumn{18}{l}{\textsc{\textcolor{ink}{Hand-crafted multi-agent}}} \\[0pt]
\hspace{2mm}MultiPersona \citep{wang2024unleashing} & 62.8 & \up{0.9} & \tk{4.5} & 12.4 & \dn{24.3} & \tk{6.3} & 66.7 & \up{24.1} & \tk{9.4} & 47.4 & \dn{20.1} & \tk{8.8} & 70.4 & \dn{2.5} & \tk{9.2} & 48.3 & \tk{7.2} \\
\hspace{2mm}LLM-Debate \citep{du2023improving} & 79.5 & \up{17.6} & \tk{28.6} & 27.8 & \dn{8.9} & \tk{43.5} & \underline{70.6} & \up{28.0} & \tk{57.1} & 69.9 & \up{2.4} & \tk{49.8} & 75.6 & \up{2.7} & \tk{56.7} & 63.2 & \tk{44.6} \\
\hspace{2mm}LLM-Blender \citep{jiang2023llm} & 68.2 & \up{6.3} & \tk{26.6} & 17.8 & \dn{18.9} & \tk{38.6} & 70.1 & \up{27.5} & \tk{56.2} & 69.5 & \up{2.0} & \tk{50.1} & 73.6 & \up{0.7} & \tk{56.3} & 57.3 & \tk{42.9} \\
\hspace{2mm}DyLAN \citep{liu2023dynamic} & 71.5 & \up{9.6} & \tk{29.8} & 48.3 & \up{11.6} & \tk{54.3} & -- &  & -- & 71.2 & \up{3.7} & \tk{122} & 41.2 & \dn{31.7} & \tk{125} & 58.1 & \tk{82.8} \\
\hspace{2mm}AgentVerse \citep{chen2023agentverse} & 51.8 & \dn{10.1} & \tk{26.6} & 20.1 & \dn{16.6} & \tk{38.9} & -- &  & -- & 56.2 & \dn{11.3} & \tk{94.5} & 62.6 & \dn{10.3} & \tk{95.5} & 47.7 & \tk{63.9} \\
\hspace{2mm}MacNet$^\dagger$ \citep{qian2025scaling} & 43.3 & \dn{18.6} & \tk{70.3} & 1.4 & \dn{35.3} & \tk{126} & -- &  & -- & 15.9 & \dn{51.6} & \tk{118} & 12.2 & \dn{60.7} & \tk{141} & 18.2 & \tk{114} \\
\noalign{\vskip 2pt}
\multicolumn{18}{l}{\textsc{\textcolor{ink}{Automated agentic systems}}} \\[0pt]
\hspace{2mm}AutoAgents \citep{chen2024autoagents} & 18.7 & \dn{43.2} & \tk{49.7} & 0.9 & \dn{35.8} & \tk{89.3} & 14.2 & \dn{28.4} & \tk{94.2} & 7.8 & \dn{59.7} & \tk{82.8} & 13.0 & \dn{59.9} & \tk{79.7} & 10.1 & \tk{75.4} \\
\hspace{2mm}GPTSwarm \citep{zhuge2024language} & 69.3 & \up{7.4} & \tk{23.0} & 39.8 & \up{3.1} & \tk{25.5} & 62.3 & \up{19.7} & \tk{37.6} & 66.1 & \dn{1.4} & \tk{23.1} & 66.4 & \dn{6.5} & \tk{71.1} & 60.4 & \tk{35.7} \\
\hspace{2mm}ADAS \citep{hu2025automated} & 71.3 & \up{9.4} & \tk{27.8} & 21.8 & \dn{14.9} & \tk{38.8} & 64.0 & \up{21.4} & \tk{65.5} & 61.2 & \dn{6.3} & \tk{15.0} & 49.4 & \dn{23.5} & \tk{43.5} & 50.9 & \tk{31.3} \\
\hspace{2mm}AFlow \citep{zhang2025aflow} & 61.1 & \dn{0.8} & \tk{3.8} & 37.4 & \up{0.7} & \tk{36.7} & 42.8 & \up{0.2} & \tk{9.1} & 67.4 & \dn{0.1} & \tk{7.5} & 66.8 & \dn{6.1} & \tk{8.6} & 58.2 & \tk{14.2} \\
\hspace{2mm}MAS-Zero \citep{ke2025mas} & 70.9 & \up{9.0} & \tk{558} & 30.0 & \dn{6.7} & \tk{729} & 65.1 & \up{22.5} & \tk{916} & 67.6 & \up{0.1} & \tk{789} & 64.9 & \dn{8.0} & \tk{1092} & 58.4 & \tk{792} \\
\hspace{2mm}MaAS \citep{zhang2025multi} & 66.0 & \up{4.1} & \tk{20.1} & 29.4 & \dn{7.3} & \tk{50.3} & 54.5 & \up{11.9} & \tk{50.9} & 64.0 & \dn{3.5} & \tk{18.4} & \underline{83.8} & \up{10.9} & \tk{26.5} & 60.8 & \tk{28.8} \\
\noalign{\vskip 2pt}
\midrule
\rowcolor{accl}\textbf{IAPO} & \underline{81.8} & \up{19.9} & \tk{25.1} & \underline{48.6} & \up{11.9} & \tk{50.8} & \textbf{74.2} & \up{31.6} & \tk{46.0} & \textbf{80.9} & \up{13.4} & \tk{33.1} & \textbf{84.3} & \up{11.4} & \tk{20.4} & \textbf{73.9} & \tk{32.4} \\
\rowcolor{accl}\textbf{IAPO-Lite} & \textbf{85.1} & \up{23.2} & \tk{25.8} & \textbf{54.3} & \up{17.6} & \tk{51.1} & 66.1 & \up{23.5} & \tk{51.7} & \underline{78.3} & \up{10.8} & \tk{29.0} & 75.7 & \up{2.8} & \tk{21.6} & \underline{73.4} & \tk{31.9} \\
\bottomrule
\end{tabular}}
\end{table}

\paragraph{Accuracy at matched compute.} At the strongest baseline's budget
(Table~\ref{tab:main}), IAPO leads by $2.3$ points on GSM8K, $3.6$ on DROP and $9.7$
on MBPP ($z{=}6.3$, against DyLAN), and is level with the strongest baseline on MATH
($+0.3$) and HumanEval ($+0.5$). The GSM8K and DROP differences are significant
(Appendix~\ref{app:results}). IAPO-Lite, trained with the higher token price, goes
further on GSM8K ($+5.6$), MATH ($+6.0$, $z{=}2.6$) and MBPP ($+7.2$, $z{=}4.6$),
and gives up accuracy on DROP ($-4.5$) and HumanEval ($-8.1$) in exchange for fewer
tokens.

\begin{figure}[t]
\centering
\includegraphics[width=\textwidth]{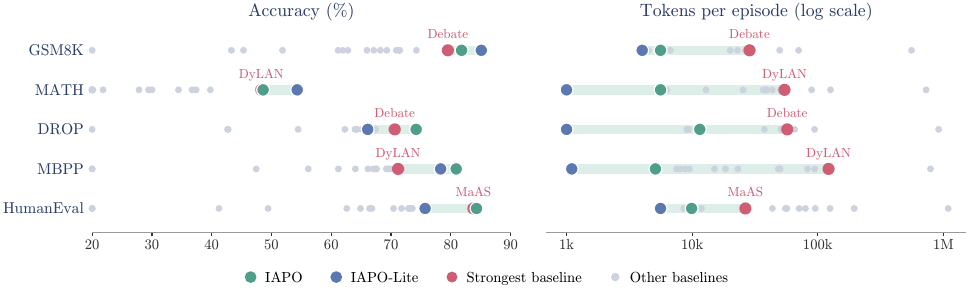}
\caption{Accuracy (left) and tokens per episode (right, log scale) of IAPO, IAPO-Lite,
the strongest baseline of each benchmark and the other baselines. Accuracy is that of the
voted plan within the strongest baseline's budget; tokens are those of a single episode.}
\label{fig:cost}
\end{figure}

\paragraph{Cost.} One deployed IAPO episode costs 5.1k to 11.5k tokens, against
26.5k to 121.9k for the strongest baselines, that is 2.7 to 24 times fewer
(Figure~\ref{fig:cost}); IAPO-Lite costs 1.0k to 5.6k. On code a single episode
already matches or beats the strongest baseline: 81.1\% on MBPP at 5.1k tokens
against DyLAN's 71.2\% at 121.9k, and 83.7\% on HumanEval at 9.9k against MaAS's
83.8\% at 26.5k. The matched-budget plans spend 20.4k to 50.8k tokens per problem,
each within the strongest baseline's budget.

\subsection{Ablation on the Interaction Term}
\label{sec:exp-ablation}

\begin{figure}[t]
\centering
\includegraphics[width=\textwidth]{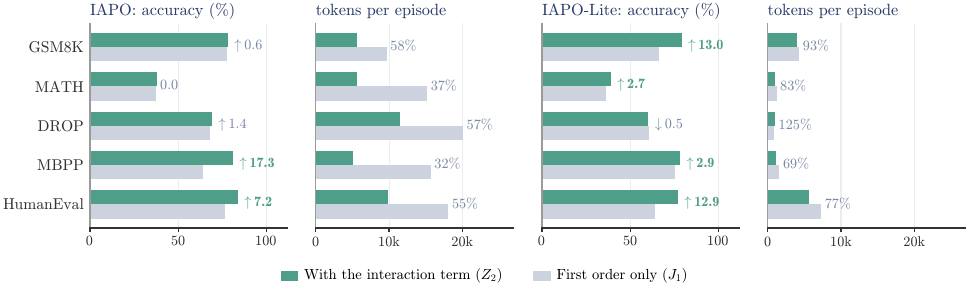}
\caption{With the interaction term ($Z_2$) and with first order only ($J_1$), for IAPO
(left) and IAPO-Lite (right): accuracy with its gain in points (bold: $z{>}2$), and
tokens per episode with IAPO's tokens as a share of the first-order variant's.}
\label{fig:ablation}
\end{figure}

The first-order variant removes $J^{(2)}$ from fitting and plan ranking;
validation, certification, retriever and roles are the same, so its gap to IAPO
comes from the interaction term alone. For IAPO, the interaction term raises
accuracy by $17.3$ points on MBPP and $7.2$ on HumanEval (both $z{>}2$), by $1.4$
on DROP and $0.6$ on GSM8K, and leaves MATH unchanged, while IAPO uses $32\%$ to
$58\%$ of the first-order variant's tokens on every benchmark
(Figure~\ref{fig:ablation}). For IAPO-Lite it raises accuracy by $13.0$ points on
GSM8K, $12.9$ on HumanEval, $2.9$ on MBPP and $2.7$ on MATH (all $z{>}2$) and lowers
it by $0.5$ on DROP, where its tokens rise to $125\%$. The paired differences and
$z$-values are in Appendix~\ref{app:results}.

\subsection{What the Policies Deploy}
\label{sec:exp-deployed}

The deployed plans show where the interaction term pays (Appendix~\ref{app:results}).
Test-and-repair helps only after a program exists, a pair of decisions. IAPO deploys
it in all four code runs; the first-order variant never proposes it, fills the slots
with a second programmer, and loses $17$ points on MBPP and $7$ on HumanEval at
$3.1$ and $1.8$ times the tokens. On question answering and mathematics, all seven
IAPO runs follow the evidence finder with the analyst, usually with a checker and the
second answerer after it, while the first-order variant never calls the analyst and
reads the whole document with the second answerer instead. The analyst alone loses
points on every benchmark, so a first-order score never selects it, yet following the
evidence finder with the analyst gains (Appendix~\ref{app:results}). 

\subsection{Controlled Tasks}
\label{sec:exp-controlled}

One probability $q$ is shared by $N$ decisions with frozen binary workers, and all
$2^N$ trajectories are enumerated (Appendix~\ref{app:exp-controlled}).
Complementarity is the trap of Proposition~\ref{prop:trap}, Redundancy has
substitutes, and Evidence threshold has $N{=}6$.

The first-order score is zero for every step on Complementarity (oracle gain $0.148$) and
overshoots on Redundancy ($0.006$ against $0.010$), while the second-order score picks
the oracle step on both. On Evidence threshold the truncation bound exceeds the
second-order gain, so $Z_2$ alone deploys a gain of $0.087$, and adding
$Z_{\mathrm{IS}}$ restores $0.356$. Across $6{,}000$ validation runs, no candidate with negative gain
was accepted.

\bibliographystyle{plainnat}
\bibliography{Reference}

\clearpage
\appendix
\section{Assumptions, Conventions, and Notation}
\label{app:setup}

Throughout, $\pi$ is the current policy, which generates the data, $\pi'$ is a
candidate policy, $\mathbb{E}_\pi$ and $\Pr_\pi$ are over trajectories of
$\pi$, and all other quantities are as in Sections~\ref{sec:formulation}
and~\ref{sec:methods}.

\subsection{The Overlap Assumption}

\begin{trivlist}\item[]\textbf{Assumption~\ref{ass:overlap} (Overlap).}\ 
There exists $W\ge1$ such that $\pi'_t(a\mid s)\le W\pi_t(a\mid s)$ for every
$t\in[H]$, every state $s$ reachable under $\pi$, and every
$a\in\mathcal A(s)$.
\end{trivlist}

By Lemma~\ref{lem:reweight}, Assumption~\ref{ass:overlap} gives
$\Pr_{\pi'}(\tau_t)>0\Rightarrow\Pr_\pi(\tau_t)>0$, and $\rho_{t'}\le W$
gives $\prod_{t'\le t}\rho_{t'}\le W^t$. Moreover:
\begin{itemize}
\item $W$ enters only the estimators, through $\rho_t\in[0,W]$; the expansion
of Theorem~\ref{thm:expansion} is an identity for every $\pi'$, and overlap
only writes each term as an expectation under $\pi$
(Remark~\ref{rem:no-overlap}).
\item The bounded update \eqref{eq:bounded-policy} satisfies
Assumption~\ref{ass:overlap} with $W=1+\eta$
(Proposition~\ref{prop:bounded}); PPO clipping bounds $\rho_t$ only on the
sampled actions and does not imply Assumption~\ref{ass:overlap} for the
returned policy.
\item For estimation the assumption cannot be dropped:
$\pi'_t(a\mid s)>0=\pi_t(a\mid s)$ at a reachable $s$ implies
$\Pr_\pi(s_t=s,a_t=a)=0$, so no number of trajectories of $\pi$ identifies the
reward that $\pi'$ collects after $(s,a)$.
\end{itemize}

\subsection{Conventions}

\paragraph{Finite trajectories.}
$|\mathcal{Q}|,|\mathcal{Y}|,H<\infty$, so all state, action, and trajectory
sets are finite, every expectation below is a finite sum, and sums,
expectations, and derivatives commute.

\paragraph{Fixed horizon.}
An episode that stops before step $H$ stays until step $H$ in an absorbing
state with a single action and zero reward, attributed to no agent in
$\mathcal{N}$. There $\rho_t=1$, so these steps enter no term below and no
visit count. With at most $K$ calls of two steps each and one stop, $H=2K+1$
and the orchestrator acts at most $K+1$ times.

\paragraph{The state determines the history.}
Let $\tau_t\coloneqq(s_1,a_1,\dots,s_t,a_t)$ for $t\ge1$ and $\tau_0$ be empty.
An orchestrator state $(q,h_k)$ and a worker state $(q,h_k,j,\mathcal{E})$
record the task, the earlier calls $(j_\ell,\mathcal{E}_\ell,y_\ell)$, and the
current call, which are the earlier actions; the earlier states are functions
of these, so $s_t$ determines $\tau_{t-1}$. The proofs use only this property
and hold for every finite-horizon MDP that has it.

\paragraph{Acting agents.}
At every non-absorbing step, $i_t=i(s_t)$ and
$\pi_t(a\mid s)=\pi^{i(s)}(a\mid s)$. A worker policy
$\pi^j(\cdot\mid q,\mathcal{E})$ depends on $s$ only through
$(q,\mathcal{E})$, a function of $s$, so we write it as $\pi_t(\cdot\mid s)$.

\paragraph{Reward bounds.}
Let $\bar r_t,\bar G\ge0$ satisfy $|r_t(s,a)|\le\bar r_t$ for all $(s,a)$ and
$\sum_{t=1}^H|r_t(s_t,a_t)|\le\bar G$ on every trajectory. In the
orchestration MDP, a trajectory has at most $K$ calls, each with reward
$-\lambda c_j$, and one stop, with reward $R\in[0,1]$, so:
\begin{equation*}
\sum_{t=1}^H|r_t(s_t,a_t)|\le\lambda\sum_{\text{calls}}c_j+R\le\lambda K\max_jc_j+1.
\end{equation*}
Hence $\bar G=1+\lambda K\max_jc_j$ and $\bar r_t=\max\{1,\lambda\max_jc_j\}$
are valid.

\subsection{Notation}

For distributions $p,p'$ on a finite set,
$\mathrm{TV}(p,p')\coloneqq\frac12\sum_a|p(a)-p'(a)|$. On a trajectory of
$\pi$:
\begin{equation*}
\rho_t\coloneqq\frac{\pi'_t(a_t\mid s_t)}{\pi_t(a_t\mid s_t)},\qquad
G_t\coloneqq\sum_{t'=t}^Hr_{t'}(s_{t'},a_{t'}),\qquad G_{H+1}\coloneqq0.
\end{equation*}
The policy change of agent $i$ is
$\alpha_i\coloneqq\max_{s:\,i(s)=i}\mathrm{TV}(\pi'^i(\cdot\mid s),\pi^i(\cdot\mid s))$.
Its visit count is $n_{i,t}\coloneqq\max|\{t'\le t:i_{t'}=i\}|$, the maximum
over trajectories of $\pi$, and $n_i\coloneqq n_{i,H}$; in the orchestration
MDP, $n_0\le K+1$.

For $S\subseteq[H]$, the \emph{agent-count vector}
$\mathbf{k}(S)\in\mathbb{N}^{\mathcal{N}}$ has entries
$k_i(S)\coloneqq|\{t\in S:i_t=i\}|$. For $\mathbf{k}\in\mathbb{N}^{\mathcal{N}}$,
$|\mathbf{k}|\coloneqq\sum_ik_i$, and $\mathbf{e}_i$ is the unit vector of
agent $i$. The \emph{interaction coefficient} of $\mathbf{k}\neq\mathbf{0}$
is:
\begin{equation}
J_{\mathbf{k}}\coloneqq\mathbb{E}_\pi\Big[\sum_{S:\,\mathbf{k}(S)=\mathbf{k}}\ \prod_{t\in S}(\rho_t-1)\,A_{t_S}^\pi(s_{t_S},a_{t_S})\Big],
\qquad t_S\coloneqq\max S.
\label{eq:app-coef}
\end{equation}
Every $S$ that contains an absorbing step has $\prod_{t\in S}(\rho_t-1)=0$,
and every other $S$ has $|\mathbf{k}(S)|=|S|$, so grouping the sets of size
$m$ by their counts gives:
\begin{equation}
J^{(m)}=\sum_{|\mathbf{k}|=m}J_{\mathbf{k}}.
\label{eq:order-by-counts}
\end{equation}
The first-order term $J^{(1)}$ is the objective of PPO. We split it and the
second-order term by the acting agents: $L_i$ is the part of $J^{(1)}$ from
the steps of agent $i$, and:
\begin{equation*}
I_{i\to i'}\coloneqq\mathbb{E}_\pi\Big[\sum_{\substack{t'<t:\,i_{t'}=i,\\ i_t=i'}}
(\rho_{t'}-1)(\rho_t-1)A_t^\pi(s_t,a_t)\Big],
\end{equation*}
so that $J^{(1)}=\sum_iL_i$ and $J^{(2)}=\sum_{i,i'}I_{i\to i'}$.

\section{Proof of Theorem~\ref{thm:expansion} (Exact Interaction Expansion)}
\label{app:expansion}

We prove a general form that also covers mixtures of policies, vanishing
coefficients, and the first two orders.

\subsection{The Trajectory Law and the Value Functions}
\label{app:law}

\begin{lemma}[Trajectory law]
\label{lem:prefix}
For every policy $\pi$ and every $t\in[H]$:
\begin{equation}
\Pr_\pi(\tau_t)=d_0(s_1)\prod_{t'=1}^t\pi_{t'}(a_{t'}\mid s_{t'})\prod_{t'=1}^{t-1}P_{t'}(s_{t'+1}\mid s_{t'},a_{t'}),
\label{eq:law}
\end{equation}
and the objective is
$J(\pi)=\sum_{t=1}^H\sum_{\tau_t}\Pr_\pi(\tau_t)\,r_t(s_t,a_t)$.
\end{lemma}

\begin{proof}
Splitting $\Pr_\pi(\tau_t)$ in time order:
\begin{align*}
\Pr_\pi(\tau_t)
&=\Pr(s_1)\prod_{t'=1}^t\Pr_\pi(a_{t'}\mid\tau_{t'-1},s_{t'})\prod_{t'=1}^{t-1}\Pr(s_{t'+1}\mid\tau_{t'})\\
&\overset{(a)}{=}d_0(s_1)\prod_{t'=1}^t\pi_{t'}(a_{t'}\mid s_{t'})\prod_{t'=1}^{t-1}P_{t'}(s_{t'+1}\mid s_{t'},a_{t'}),
\end{align*}
where (a) uses $\Pr(s_1)=d_0(s_1)$,
$\Pr_\pi(a_{t'}\mid\tau_{t'-1},s_{t'})=\pi_{t'}(a_{t'}\mid s_{t'})$ since
$s_{t'}$ determines $\tau_{t'-1}$, and
$\Pr(s_{t'+1}\mid\tau_{t'})=P_{t'}(s_{t'+1}\mid s_{t'},a_{t'})$ since $P_{t'}$
depends on neither $\pi$ nor $\tau_{t'-1}$. This is \eqref{eq:law}. Since
$r_t(s_t,a_t)$ is a function of $\tau_t$, linearity gives:
\begin{equation*}
J(\pi)=\sum_{t=1}^H\mathbb{E}_\pi[r_t(s_t,a_t)]=\sum_{t=1}^H\sum_{\tau_t}\Pr_\pi(\tau_t)\,r_t(s_t,a_t).
\end{equation*}
\end{proof}

\begin{lemma}[Value functions]
\label{lem:bellman}
For every step $t$, every prefix $\tau_t$ with $\Pr_\pi(\tau_t)>0$,
and every state $s$:
\begin{enumerate}
\item[(a)] $\mathbb{E}_\pi[G_t\mid\tau_t]=Q_t^\pi(s_t,a_t)$;
\item[(b)] $V_t^\pi(s)=\sum_a\pi_t(a\mid s)Q_t^\pi(s,a)$, and hence
$\sum_a\pi_t(a\mid s)A_t^\pi(s,a)=0$;
\item[(c)] $|Q_t^\pi(s,a)|\le\min\big\{\sum_{t'=t}^H\bar r_{t'},\,\bar G\big\}$.
\end{enumerate}
\end{lemma}

\begin{proof}
For a completion $\tau_H$ of $\tau_t$, dividing \eqref{eq:law} at $H$ by
\eqref{eq:law} at $t$:
\begin{equation*}
\frac{\Pr_\pi(\tau_H)}{\Pr_\pi(\tau_t)}
=\prod_{t'=t}^{H-1}P_{t'}(s_{t'+1}\mid s_{t'},a_{t'})\,\pi_{t'+1}(a_{t'+1}\mid s_{t'+1}),
\end{equation*}
which depends on $\tau_t$ only through $(s_t,a_t)$.

For (a), $G_t$ is a function of $(s_t,a_t,\dots,s_H,a_H)$, so its law given
$\tau_t$ equals its law given $(s_t,a_t)$:
\begin{equation*}
\mathbb{E}_\pi[G_t\mid\tau_t]=\mathbb{E}_\pi[G_t\mid s_t,a_t]=Q_t^\pi(s_t,a_t),
\end{equation*}
where the last equality is the definition of $Q_t^\pi$
(Section~\ref{sec:mdp}).

For (b), by the tower rule over $a_t\sim\pi_t(\cdot\mid s)$ and
$\sum_a\pi_t(a\mid s)=1$:
\begin{align*}
V_t^\pi(s)&=\mathbb{E}_\pi[G_t\mid s_t=s]
=\sum_a\pi_t(a\mid s)\,\mathbb{E}_\pi[G_t\mid s_t=s,a_t=a]
=\sum_a\pi_t(a\mid s)\,Q_t^\pi(s,a),\\
\sum_a\pi_t(a\mid s)A_t^\pi(s,a)&=\sum_a\pi_t(a\mid s)Q_t^\pi(s,a)-V_t^\pi(s)\sum_a\pi_t(a\mid s)=V_t^\pi(s)-V_t^\pi(s)=0.
\end{align*}

For (c), since $Q_t^\pi(s,a)$ is a conditional mean of $G_t$ and
$|G_t|\le\sum_{t'=t}^H|r_{t'}(s_{t'},a_{t'})|$:
\begin{equation*}
|Q_t^\pi(s,a)|\le\mathbb{E}_\pi\Big[\sum_{t'=t}^H|r_{t'}(s_{t'},a_{t'})|\,\Big|\,s_t=s,a_t=a\Big]\le\min\Big\{\sum_{t'=t}^H\bar r_{t'},\,\bar G\Big\}.
\end{equation*}
\end{proof}

\subsection{Reweighting and Centering}
\label{app:reweight}

\begin{lemma}[Reweighting]
\label{lem:reweight}
Under Assumption~\ref{ass:overlap}, for every step $t$ and every function $f$
of $\tau_t$:
\begin{equation}
\mathbb{E}_{\pi'}[f(\tau_t)]=\mathbb{E}_\pi\Big[\prod_{t'\le t}\rho_{t'}\,f(\tau_t)\Big].
\label{eq:reweight}
\end{equation}
\end{lemma}

\begin{proof}
Let $\Pr_{\pi'}(\tau_t)>0$. We show $\Pr_\pi(\tau_t)>0$ by induction over
$t'$: $d_0(s_1)>0$, so $s_1$ is reachable under $\pi$, and if $s_{t'}$ is
reachable under $\pi$, Assumption~\ref{ass:overlap} gives:
\begin{equation*}
0<\pi'_{t'}(a_{t'}\mid s_{t'})\le W\pi_{t'}(a_{t'}\mid s_{t'}),
\end{equation*}
so $\pi_{t'}(a_{t'}\mid s_{t'})>0$, and
$P_{t'}(s_{t'+1}\mid s_{t'},a_{t'})>0$ makes $s_{t'+1}$ reachable under
$\pi$. Hence all factors of \eqref{eq:law} under $\pi$ are positive, and the
shared initial and transition factors cancel:
\begin{equation*}
\frac{\Pr_{\pi'}(\tau_t)}{\Pr_\pi(\tau_t)}
=\frac{d_0(s_1)\prod_{t'}\pi'_{t'}(a_{t'}\mid s_{t'})\prod_{t'}P_{t'}(s_{t'+1}\mid s_{t'},a_{t'})}
{d_0(s_1)\prod_{t'}\pi_{t'}(a_{t'}\mid s_{t'})\prod_{t'}P_{t'}(s_{t'+1}\mid s_{t'},a_{t'})}
=\prod_{t'\le t}\rho_{t'}.
\end{equation*}
Therefore:
\begin{align*}
\mathbb{E}_{\pi'}[f(\tau_t)]
&=\sum_{\tau_t:\,\Pr_{\pi'}(\tau_t)>0}\Pr_{\pi'}(\tau_t)f(\tau_t)
=\sum_{\tau_t:\,\Pr_{\pi'}(\tau_t)>0}\Pr_\pi(\tau_t)\prod_{t'\le t}\rho_{t'}\,f(\tau_t)\\
&\overset{(a)}{=}\sum_{\tau_t:\,\Pr_\pi(\tau_t)>0}\Pr_\pi(\tau_t)\prod_{t'\le t}\rho_{t'}\,f(\tau_t)
=\mathbb{E}_\pi\Big[\prod_{t'\le t}\rho_{t'}\,f(\tau_t)\Big],
\end{align*}
where (a) holds since $\Pr_\pi(\tau_t)>0=\Pr_{\pi'}(\tau_t)$ forces a factor
$\pi'_{t'}(a_{t'}\mid s_{t'})=0$ in \eqref{eq:law}, so
$\prod_{t'\le t}\rho_{t'}=0$.
\end{proof}

\begin{lemma}[Centering]
\label{lem:center}
Under Assumption~\ref{ass:overlap}, for every step $t$:
\begin{equation*}
\mathbb{E}_\pi[\rho_t-1\mid\tau_{t-1},s_t]=0,\qquad
\mathbb{E}_\pi\big[|\rho_t-1|\,\big|\,\tau_{t-1},s_t\big]=2\,\mathrm{TV}\big(\pi'_t(\cdot\mid s_t),\pi_t(\cdot\mid s_t)\big)\le2\alpha_{i_t}.
\end{equation*}
with the right side read as $0$ at absorbing steps, where $\rho_t=1$.
\end{lemma}

\begin{proof}
Given $(\tau_{t-1},s_t)$, $a_t\sim\pi_t(\cdot\mid s_t)$. Write
$p_a\coloneqq\pi_t(a\mid s_t)$ and $p'_a\coloneqq\pi'_t(a\mid s_t)$. By
Assumption~\ref{ass:overlap}, $p_a=0\Rightarrow p'_a=0$, so
$\sum_{a:\,p_a>0}p_a=\sum_{a:\,p_a>0}p'_a=1$ and:
\begin{align*}
\mathbb{E}_\pi[\rho_t-1\mid\tau_{t-1},s_t]&=\sum_{a:\,p_a>0}p_a\Big(\frac{p'_a}{p_a}-1\Big)
=\sum_{a:\,p_a>0}p'_a-\sum_{a:\,p_a>0}p_a=1-1=0,\\
\mathbb{E}_\pi\big[|\rho_t-1|\,\big|\,\tau_{t-1},s_t\big]&=\sum_{a:\,p_a>0}p_a\Big|\frac{p'_a}{p_a}-1\Big|
\overset{(a)}{=}\sum_a|p'_a-p_a|=2\,\mathrm{TV}(p',p)\overset{(b)}{\le}2\alpha_{i_t},
\end{align*}
where (a) uses $p_a=0\Rightarrow p'_a=0$, and (b) is the definition of
$\alpha_{i_t}$, since $i(s_t)=i_t$.
\end{proof}

\begin{lemma}[Expanding a product]
\label{lem:product}
For real numbers $x_1,\dots,x_t$:
\begin{equation*}
\prod_{t'=1}^t(1+x_{t'})=\sum_{S\subseteq[t]}\prod_{t'\in S}x_{t'},
\end{equation*}
where the empty product equals $1$.
\end{lemma}

\begin{proof}
Induction on $t$. For $t=1$,
$1+x_1=\prod_{t'\in\varnothing}x_{t'}+\prod_{t'\in\{1\}}x_{t'}$. If the
identity holds for $t-1$, multiplying by $1+x_t$:
\begin{equation*}
\prod_{t'=1}^t(1+x_{t'})=\sum_{S\subseteq[t-1]}\prod_{t'\in S}x_{t'}+\sum_{S\subseteq[t-1]}x_t\prod_{t'\in S}x_{t'}
=\sum_{S\subseteq[t],\,t\notin S}\prod_{t'\in S}x_{t'}+\sum_{S\subseteq[t],\,t\in S}\prod_{t'\in S}x_{t'},
\end{equation*}
which is the sum over all $S\subseteq[t]$. For example,
$(1+x_1)(1+x_2)=1+x_1+x_2+x_1x_2$.
\end{proof}

\subsection{The Expansion}
\label{app:the-expansion}

\begin{trivlist}\item[]\textbf{Theorem~\ref{thm:expansion} (Interaction expansion).}\ \itshape
Under Assumption~\ref{ass:overlap}:
\begin{equation*}
J(\pi')-J(\pi)=\sum_{m=1}^H J^{(m)},
\end{equation*}
where the order-$m$ interaction is:
\begin{equation*}
J^{(m)}
=\mathbb E_\pi\Big[
\sum_{1\le t_1<\cdots<t_m\le H}
(\rho_{t_1}-1)\cdots(\rho_{t_m}-1)\,
A_{t_m}^\pi(s_{t_m},a_{t_m})
\Big].
\tag{\ref{eq:order-m}}
\end{equation*}
Moreover, each $J^{(m)}$ remains unchanged
if $A_{t_m}^\pi$ is replaced by
$G_{t_m}-b(s_{t_m})$ for any bounded function $b$.
\end{trivlist}

By item (ii) below, $J_{\mathbf{k}}$ is the coefficient of
$\prod_i\omega_i^{k_i}$ in a polynomial in the mixing weights of the agents.

\begin{theorem}[Exact interaction expansion, general form]
\label{thm:app-expansion}
Under Assumption~\ref{ass:overlap}, the following hold.
\begin{enumerate}
\item[(i)] $J(\pi')-J(\pi)=\sum_{\mathbf{k}\neq\mathbf{0}}J_{\mathbf{k}}=\sum_{m=1}^HJ^{(m)}$.
\item[(ii)] For $\omega\in[0,1]^{\mathcal{N}}$, let $\pi_\omega$ use
$\pi^i+\omega_i(\pi'^i-\pi^i)$ for every agent $i$. Then:
\begin{equation*}
J(\pi_\omega)=J(\pi)+\sum_{\mathbf{k}\neq\mathbf{0}}J_{\mathbf{k}}\prod_{i\in\mathcal{N}}\omega_i^{k_i}.
\end{equation*}
\item[(iii)] The advantage in \eqref{eq:app-coef} can be replaced by
$G_{t_S}-\psi(s_{t_S})$ for any bounded function $\psi$.
\item[(iv)] $J_{\mathbf{k}}=0$ if $k_i>0$ for some unchanged agent $i$, or if
no trajectory of $\pi$ contains a set of steps of changed agents with the counts
$\mathbf{k}$ in question.
\item[(v)] $J_{\mathbf{e}_i}=L_i$, $J_{2\mathbf{e}_i}=I_{i\to i}$, and
$J_{\mathbf{e}_i+\mathbf{e}_{i'}}=I_{i\to i'}+I_{i'\to i}$ for $i\neq i'$.
\end{enumerate}
\end{theorem}

\begin{proof}
By Lemma~\ref{lem:prefix} and Lemma~\ref{lem:reweight} with
$f=r_t(s_t,a_t)$, we obtain \eqref{eq:is}:
\begin{equation}
J(\pi')-J(\pi)=\sum_{t=1}^H\Big(\mathbb{E}_{\pi'}[r_t(s_t,a_t)]-\mathbb{E}_\pi[r_t(s_t,a_t)]\Big)
=\sum_{t=1}^H\mathbb{E}_\pi\Big[\Big(\prod_{t'\le t}\rho_{t'}-1\Big)r_t(s_t,a_t)\Big].
\label{eq:pf-exp-is}
\end{equation}
By Lemma~\ref{lem:product} with $x_{t'}=\rho_{t'}-1$, where $S=\varnothing$
contributes $1$ and cancels the $-1$:
\begin{equation*}
\prod_{t'\le t}\rho_{t'}-1=\prod_{t'\le t}\big(1+(\rho_{t'}-1)\big)-1=\sum_{\varnothing\neq S\subseteq[t]}\ \prod_{t'\in S}(\rho_{t'}-1).
\end{equation*}
Inserting this into \eqref{eq:pf-exp-is}, dropping the sets $S$ that contain
an absorbing step (their products vanish), and splitting the others by
$\mathbf{k}(S)$ gives the \emph{reward form}, used again in
Appendix~\ref{app:visit}:
\begin{equation}
J(\pi')-J(\pi)=\sum_{\mathbf{k}\neq\mathbf{0}}\ \sum_{t=1}^H\mathbb{E}_\pi\Big[\sum_{S\subseteq[t]:\,\mathbf{k}(S)=\mathbf{k}}\ \prod_{t'\in S}(\rho_{t'}-1)\ r_t(s_t,a_t)\Big].
\label{eq:reward-form}
\end{equation}
With $S$ running over $\{S:\mathbf{k}(S)=\mathbf{k}\}$ in the inner sums:
\begin{align}
J(\pi')-J(\pi)
&\overset{(a)}{=}\sum_{\mathbf{k}\neq\mathbf{0}}\mathbb{E}_\pi\Big[\sum_S\prod_{t'\in S}(\rho_{t'}-1)\sum_{t=t_S}^Hr_t(s_t,a_t)\Big]
\overset{(b)}{=}\sum_{\mathbf{k}\neq\mathbf{0}}\mathbb{E}_\pi\Big[\sum_S\prod_{t'\in S}(\rho_{t'}-1)\,G_{t_S}\Big],
\label{eq:pf-exp-return}
\end{align}
where (a) exchanges the finite sums using
$\{(t,S):S\subseteq[t]\}=\{(t,S):t\ge t_S\}$, and (b) is the definition of
$G_{t_S}$.

Fix $S$, let $t=t_S$, and write
$w_S\coloneqq\prod_{t'\in S}(\rho_{t'}-1)=w_S^-\,(\rho_t-1)$ with
$w_S^-\coloneqq\prod_{t'\in S,\,t'<t}(\rho_{t'}-1)$, so $w_S$ is a function of
$\tau_t$ and $w_S^-$ of $\tau_{t-1}$. For every bounded $\psi$:
\begin{align}
\mathbb{E}_\pi[w_S\,G_t]&\overset{(a)}{=}\mathbb{E}_\pi\big[w_S\,\mathbb{E}_\pi[G_t\mid\tau_t]\big]\overset{(b)}{=}\mathbb{E}_\pi\big[w_S\,Q_t^\pi(s_t,a_t)\big],
\label{eq:pf-exp-q}\\
\mathbb{E}_\pi[w_S\,\psi(s_t)]&\overset{(c)}{=}\mathbb{E}_\pi\Big[w_S^-\,\psi(s_t)\,\mathbb{E}_\pi[\rho_t-1\mid\tau_{t-1},s_t]\Big]\overset{(d)}{=}\mathbb{E}_\pi\big[w_S^-\,\psi(s_t)\cdot0\big]=0,
\label{eq:pf-exp-baseline}
\end{align}
where (a) is the tower rule, (b) uses Lemma~\ref{lem:bellman}(a), (c)
conditions on $(\tau_{t-1},s_t)$, and (d) uses Lemma~\ref{lem:center}.

For (i) and (iii), subtracting \eqref{eq:pf-exp-baseline} with $\psi=V_t^\pi$
from \eqref{eq:pf-exp-q} and using $A_t^\pi=Q_t^\pi-V_t^\pi$:
\begin{equation*}
\mathbb{E}_\pi[w_S\,G_{t_S}]=\mathbb{E}_\pi\big[w_S\,A_{t_S}^\pi(s_{t_S},a_{t_S})\big].
\end{equation*}
In \eqref{eq:pf-exp-return}, the sum over $S$ then gives $J_{\mathbf{k}}$ in
\eqref{eq:app-coef}, and the sum over $\mathbf{k}$ gives the first equality
in (i); the second is \eqref{eq:order-by-counts}. For an arbitrary bounded
$\psi$, \eqref{eq:pf-exp-baseline} gives
$\mathbb{E}_\pi[w_SG_{t_S}]=\mathbb{E}_\pi[w_S(G_{t_S}-\psi(s_{t_S}))]$, which
is (iii).

For (ii), $\pi^i+\omega_i(\pi'^i-\pi^i)=(1-\omega_i)\pi^i+\omega_i\pi'^i$ is a
convex combination of distributions, and on reachable states:
\begin{equation*}
(1-\omega_i)\pi^i+\omega_i\pi'^i\le(1-\omega_i)\pi^i+\omega_iW\pi^i\le W\pi^i,
\end{equation*}
so $\pi_\omega$ satisfies Assumption~\ref{ass:overlap} with the same $W$, and
its ratio at step $t'$ is:
\begin{equation*}
\frac{\pi_{t'}(a_{t'}\mid s_{t'})+\omega_{i_{t'}}\big(\pi'_{t'}(a_{t'}\mid s_{t'})-\pi_{t'}(a_{t'}\mid s_{t'})\big)}{\pi_{t'}(a_{t'}\mid s_{t'})}=1+\omega_{i_{t'}}(\rho_{t'}-1).
\end{equation*}
Apply (i) to $\pi_\omega$, with coefficients under the same $\pi$ and
$A^\pi$. Since $|\{t'\in S:i_{t'}=i\}|=k_i(S)$:
\begin{equation*}
\prod_{t'\in S}\omega_{i_{t'}}(\rho_{t'}-1)=\Big(\prod_{t'\in S}\omega_{i_{t'}}\Big)\prod_{t'\in S}(\rho_{t'}-1)=\Big(\prod_{i\in\mathcal{N}}\omega_i^{k_i(S)}\Big)\prod_{t'\in S}(\rho_{t'}-1).
\end{equation*}
The factor $\prod_i\omega_i^{k_i(S)}$ depends on $S$ only through
$\mathbf{k}(S)$, so it leaves the expectation, and
$J(\pi_\omega)-J(\pi)=\sum_{\mathbf{k}}J_{\mathbf{k}}\prod_i\omega_i^{k_i}$.

For (iv), $\pi'^i=\pi^i\Rightarrow\rho_{t'}=1$ for all $t'$ with $i_{t'}=i$,
so $k_i(S)>0\Rightarrow\prod_{t'\in S}(\rho_{t'}-1)=0$. If no trajectory of
$\pi$ has steps of changed agents with counts $\mathbf{k}$,
every $S$ with $\mathbf{k}(S)=\mathbf{k}$ contains a step of an unchanged
agent, so its product is zero and the term vanishes.

For (v), $|\mathbf{k}|=1$ gives $S=\{t\}$ and
$\mathbf{k}(S)=\mathbf{e}_i\iff i_t=i$, so $J_{\mathbf{e}_i}=L_i$.
$|\mathbf{k}|=2$ gives $S=\{t',t\}$ with $t'<t=t_S$, and the term
$(\rho_{t'}-1)(\rho_t-1)A_t^\pi(s_t,a_t)$ belongs to $I_{i_{t'}\to i_t}$. Since
$\mathbf{k}(S)=2\mathbf{e}_i\iff i_{t'}=i_t=i$ and
$\mathbf{k}(S)=\mathbf{e}_i+\mathbf{e}_{i'}\iff(i_{t'},i_t)\in\{(i,i'),(i',i)\}$,
we get $J_{2\mathbf{e}_i}=I_{i\to i}$ and
$J_{\mathbf{e}_i+\mathbf{e}_{i'}}=I_{i\to i'}+I_{i'\to i}$.
\end{proof}

\begin{proof}[Proof of Theorem~\ref{thm:expansion}]
Items (i) and (iii) of Theorem~\ref{thm:app-expansion}.
\end{proof}

\begin{remark}[No critic is required]
\label{rem:critic}
Item (iii) with $\psi=0$ gives:
\begin{equation*}
J^{(m)}=\mathbb{E}_\pi\Big[\sum_{|S|=m}\prod_{t\in S}(\rho_t-1)\,G_{t_S}\Big].
\end{equation*}
Any state-dependent $\psi$ keeps every term unbiased and changes only the
variance; \eqref{eq:app-coef} is the choice $\psi=V^\pi$.
\end{remark}

\begin{remark}[Overlap is needed only for estimation]
\label{rem:no-overlap}
For any $\pi'$, \eqref{eq:reward-form} and \eqref{eq:pf-exp-return} still
hold with $\Pr_\pi(\tau_t)\prod_{t'\in S}(\rho_{t'}-1)$, for $S\subseteq[t]$,
replaced by the signed measure:
\begin{equation*}
d_0(s_1)\prod_{t'\le t,\,t'\notin S}\pi_{t'}(a_{t'}\mid s_{t'})\prod_{t'\in S}\big(\pi'_{t'}(a_{t'}\mid s_{t'})-\pi_{t'}(a_{t'}\mid s_{t'})\big)\prod_{t'<t}P_{t'}(s_{t'+1}\mid s_{t'},a_{t'}).
\end{equation*}
The expansion thus holds for every pair of policies; overlap only makes it
an expectation under $\pi$, estimable from data of $\pi$.
\end{remark}

\begin{remark}[Relation to Taylor expansion policy optimization]
\label{rem:taypo}
For a single policy in a $\gamma$-discounted MDP, \citet{tang2020taylor}
expand $Q^{\pi'}$ around $Q^\pi$ in powers of $\pi'-\pi$ through the resolvent
of the transition operator of $\pi$; their second-order objective adds to the
first-order objective:
\begin{equation*}
\frac{\gamma}{(1-\gamma)^2}\,\mathbb{E}\Big[\big(\rho(x,a)-1\big)\big(\rho(x',a')-1\big)Q^\pi(x',a')\Big],
\end{equation*}
with $(x,a)$ and $(x',a')$ drawn from discounted visitation distributions of
$\pi$ and $(x',a')$ strictly later, the discounted analog of $J^{(2)}$.
Theorem~\ref{thm:app-expansion} differs as follows: (1) their series is
infinite and converges if
$\max_x\sum_a|\pi'(a\mid x)-\pi(a\mid x)|<(1-\gamma)/\gamma$, whereas
\eqref{eq:order-m} has at most $H$ terms and is exact under
Assumption~\ref{ass:overlap}; (2) its terms split by acting agents, giving the
pair terms of Appendix~\ref{app:hessian}; (3) the remainder after order $r$ is
bounded by visit counts (Theorem~\ref{thm:app-visit}), not by a radius on
$\pi'-\pi$; (4) the second-order estimate enters a high-confidence certificate
(Theorem~\ref{thm:certificate}).
\end{remark}

\section{Proof of Corollary~\ref{cor:curvature} and Interactions across Agents}
\label{app:hessian}

Each $J_{\mathbf{k}}$ with $|\mathbf{k}|\ge2$ averages products of
$|\mathbf{k}|$ factors $\rho_t-1$, which vanish at $\pi$.

\subsection{Gradient and Curvature}
\label{app:gradient}

\begin{theorem}[Gradient and curvature]
\label{thm:hessian}
Let each agent $i$ have parameters $\theta_i$, with $\pi^i_\theta$ depending
on $\theta$ only through $\theta_i$ and $\theta\mapsto\pi_\theta(a\mid s)$ in
$C^2$. Let $\pi'=\pi_\theta$ and let $\pi=\pi_{\theta^0}$ have full support.
At $\theta=\theta^0$,
$\nabla_{\theta_i}J(\pi_\theta)=\nabla_{\theta_i}L_i$, and for $i\neq i'$:
\begin{equation*}
\nabla^2_{\theta_i\theta_{i'}}J(\pi_\theta)
=\nabla^2_{\theta_i\theta_{i'}}\big(I_{i\to i'}+I_{i'\to i}\big),
\quad
\nabla^2_{\theta_i\theta_{i'}}J^{(1)}=0.
\end{equation*}
\end{theorem}

Throughout, $\theta=(\theta_i)_{i\in\mathcal{N}}$,
$\theta\mapsto\pi_\theta(a\mid s)$ is $C^2$, $\pi=\pi_{\theta^0}$ has full
support, and $\pi'=\pi_\theta$. Every coefficient takes the expectation and the
advantage under $\pi$, so $\theta$ enters only through:
\begin{equation*}
\rho_t(\theta)\coloneqq\frac{\pi_\theta(a_t\mid s_t)}{\pi(a_t\mid s_t)},\qquad
\nabla_{\theta_{i'}}\rho_t=0\ \ (i'\neq i_t).
\end{equation*}
We write $L_i(\theta)$, $I_{i\to i'}(\theta)$, and $J_{\mathbf{k}}(\theta)$ for
the quantities of Appendix~\ref{app:setup} and \eqref{eq:app-coef} with
$\pi'=\pi_\theta$.

\begin{lemma}[Products of vanishing factors]
\label{lem:vanishing}
Let $f=\prod_{\ell=1}^qf_\ell$ with $f_\ell\in C^2$ and
$f_\ell(\theta^0)=0$ for every $\ell$. Then $\nabla f(\theta^0)=0$ if
$q\ge2$, and $\nabla^2f(\theta^0)=0$ if $q\ge3$.
\end{lemma}

\begin{proof}
By the product rule:
\begin{align*}
\nabla f&=\sum_{\ell=1}^q\nabla f_\ell\prod_{\ell'\neq\ell}f_{\ell'},\\
\nabla^2f&=\sum_\ell\nabla^2f_\ell\prod_{\ell'\neq\ell}f_{\ell'}+\sum_{\ell\neq\ell''}\nabla f_\ell\,\nabla f_{\ell''}^\top\prod_{\ell'\neq\ell,\ell''}f_{\ell'}.
\end{align*}
For $q\ge2$ (resp.\ $q\ge3$), every summand of $\nabla f$ (resp.\
$\nabla^2f$) contains an undifferentiated factor $f_{\ell'}$, and
$f_{\ell'}(\theta^0)=0$.
\end{proof}

\begin{theorem}[Gradient and curvature, general form]
\label{thm:app-hessian}
Let $g_t\coloneqq\nabla_{\theta_{i_t}}\log\pi_\theta(a_t\mid s_t)$ at
$\theta^0$. At $\theta=\theta^0$:
\begin{enumerate}
\item[(a)] $\nabla_{\theta_i}J(\pi_\theta)=\nabla_{\theta_i}L_i$ for every agent $i$;
\item[(b)] $\nabla^2_{\theta_i\theta_{i'}}J^{(1)}=0$ for $i\neq i'$;
\item[(c)] for $i\neq i'$:
\begin{equation}
\nabla^2_{\theta_i\theta_{i'}}J(\pi_\theta)=\mathbb{E}_\pi\Big[\sum_{t'<t}\big(\mathbf{1}\{i_{t'}=i,i_t=i'\}g_{t'}g_t^\top+\mathbf{1}\{i_{t'}=i',i_t=i\}g_tg_{t'}^\top\big)A_t^\pi(s_t,a_t)\Big],
\label{eq:cross-hessian}
\end{equation}
which equals $\nabla^2_{\theta_i\theta_{i'}}(I_{i\to i'}+I_{i'\to i})$;
\item[(d)] for every agent $i$:
\begin{equation*}
\nabla^2_{\theta_i\theta_i}J(\pi_\theta)=\mathbb{E}_\pi\Big[\sum_{t:\,i_t=i}\big(\nabla^2_{\theta_i}\log\pi_\theta(a_t\mid s_t)+g_tg_t^\top\big)A_t^\pi(s_t,a_t)+\sum_{t'<t:\,i_{t'}=i_t=i}(g_{t'}g_t^\top+g_tg_{t'}^\top)A_t^\pi(s_t,a_t)\Big].
\end{equation*}
\end{enumerate}
\end{theorem}

\begin{proof}
Full support gives Assumption~\ref{ass:overlap} with
$W=\max_{t,s,a}\pi_\theta(a\mid s)/\pi(a\mid s)<\infty$, so by
Theorem~\ref{thm:app-expansion}(i):
\begin{equation*}
J(\pi_\theta)-J(\pi)=\sum_{\mathbf{k}\neq\mathbf{0}}J_{\mathbf{k}}(\theta),\qquad
J_{\mathbf{k}}(\theta)=\mathbb{E}_\pi\Big[\sum_{S:\,\mathbf{k}(S)=\mathbf{k}}\prod_{t\in S}\big(\rho_t(\theta)-1\big)A_{t_S}^\pi(s_{t_S},a_{t_S})\Big].
\end{equation*}
The expectation is a finite sum with $\theta$-free weights and
$\rho_t(\theta^0)-1=0$ $\forall t$, so $J_{\mathbf{k}}$ is a finite weighted
sum of products of $|\mathbf{k}|$ such factors, and Lemma~\ref{lem:vanishing}
gives:
\begin{equation}
\nabla J_{\mathbf{k}}(\theta^0)=0\ \text{ if }|\mathbf{k}|\ge2,\qquad
\nabla^2J_{\mathbf{k}}(\theta^0)=0\ \text{ if }|\mathbf{k}|\ge3.
\label{eq:pf-vanish}
\end{equation}

For (a):
\begin{equation*}
\nabla_{\theta_i}J(\pi_\theta)\overset{(a)}{=}\sum_{\ell\in\mathcal{N}}\nabla_{\theta_i}J_{\mathbf{e}_\ell}\overset{(b)}{=}\sum_{\ell\in\mathcal{N}}\nabla_{\theta_i}L_\ell\overset{(c)}{=}\nabla_{\theta_i}L_i,
\end{equation*}
where (a) uses \eqref{eq:pf-vanish}, (b) uses
Theorem~\ref{thm:app-expansion}(v), and (c) uses $\nabla_{\theta_i}L_\ell=0$
for $\ell\neq i$.

For (b), $L_\ell$ depends only on $\theta_\ell$, so
$\nabla^2_{\theta_i\theta_{i'}}J^{(1)}=\sum_\ell\nabla^2_{\theta_i\theta_{i'}}L_\ell=0$
for $i\neq i'$.

For (c) and (d), let $i_t=i$. Differentiating
$\nabla\log\pi_\theta=\nabla\pi_\theta/\pi_\theta$ gives:
\begin{equation*}
\nabla^2\log\pi_\theta=\frac{\nabla^2\pi_\theta}{\pi_\theta}-\frac{\nabla\pi_\theta}{\pi_\theta}\Big(\frac{\nabla\pi_\theta}{\pi_\theta}\Big)^\top.
\end{equation*}
With $\pi_{\theta^0}=\pi$, at $\theta^0$:
\begin{align}
\nabla_{\theta_i}\rho_t&=\frac{\nabla_{\theta_i}\pi_\theta(a_t\mid s_t)}{\pi_\theta(a_t\mid s_t)}\Big|_{\theta^0}=\nabla_{\theta_i}\log\pi_\theta(a_t\mid s_t)\big|_{\theta^0}=g_t,
\label{eq:pf-score}\\
\nabla^2_{\theta_i}\rho_t&=\frac{\nabla^2_{\theta_i}\pi_\theta(a_t\mid s_t)}{\pi_\theta(a_t\mid s_t)}\Big|_{\theta^0}=\nabla^2_{\theta_i}\log\pi_\theta(a_t\mid s_t)\big|_{\theta^0}+g_tg_t^\top.
\label{eq:pf-ratio-hess}
\end{align}

For (c), \eqref{eq:pf-vanish} leaves $|\mathbf{k}|\le2$, and
$J_{\mathbf{k}}$ depends on $\theta_j$ only if $k_j\ge1$, so only
$\mathbf{k}=\mathbf{e}_i+\mathbf{e}_{i'}$ remains:
\begin{equation}
\nabla^2_{\theta_i\theta_{i'}}J(\pi_\theta)=\nabla^2_{\theta_i\theta_{i'}}J_{\mathbf{e}_i+\mathbf{e}_{i'}}\overset{(a)}{=}\nabla^2_{\theta_i\theta_{i'}}(I_{i\to i'}+I_{i'\to i}),
\label{eq:pf-mixed}
\end{equation}
where (a) uses Theorem~\ref{thm:app-expansion}(v). By \eqref{eq:app-coef},
$\mathbf{k}(S)=\mathbf{e}_i+\mathbf{e}_{i'}\iff S=\{t',t\}$ with $t'<t$ and
$(i_{t'},i_t)\in\{(i,i'),(i',i)\}$, so:
\begin{equation*}
J_{\mathbf{e}_i+\mathbf{e}_{i'}}(\theta)=\mathbb{E}_\pi\Big[\sum_{t'<t}\big(\mathbf{1}\{i_{t'}=i,i_t=i'\}+\mathbf{1}\{i_{t'}=i',i_t=i\}\big)\big(\rho_{t'}(\theta)-1\big)\big(\rho_t(\theta)-1\big)A_t^\pi(s_t,a_t)\Big].
\end{equation*}
On $\{i_{t'}=i,\,i_t=i'\}$, $\rho_{t'}$ depends only on $\theta_i$ and
$\rho_t$ only on $\theta_{i'}$, so:
\begin{equation*}
\nabla_{\theta_i}\nabla_{\theta_{i'}}^\top\big[(\rho_{t'}-1)(\rho_t-1)\big]\Big|_{\theta^0}=\nabla_{\theta_i}\rho_{t'}\,\big(\nabla_{\theta_{i'}}\rho_t\big)^\top\overset{(a)}{=}g_{t'}g_t^\top,
\end{equation*}
where (a) uses \eqref{eq:pf-score}; on $\{i_{t'}=i',\,i_t=i\}$ the same
computation gives $g_tg_{t'}^\top$. Differentiating termwise gives
\eqref{eq:cross-hessian}.

For (d), \eqref{eq:pf-vanish} removes $|\mathbf{k}|\ge3$. For
$\mathbf{k}=\mathbf{e}_i+\mathbf{e}_\ell$ with $\ell\neq i$, two derivatives
in $\theta_i$ leave the factor $\rho-1$ of agent $\ell$, which is $0$ at
$\theta^0$. Hence only $L_i=J_{\mathbf{e}_i}$ and $I_{i\to i}=J_{2\mathbf{e}_i}$
contribute, with summands $(\rho_t-1)A_t^\pi$ over $\{t:i_t=i\}$ and
$(\rho_{t'}-1)(\rho_t-1)A_t^\pi$ over $\{t'<t:i_{t'}=i_t=i\}$. By the
product rule:
\begin{equation*}
\nabla^2_{\theta_i}\big[(\rho_{t'}-1)(\rho_t-1)\big]=\nabla^2_{\theta_i}\rho_{t'}\,(\rho_t-1)+\nabla_{\theta_i}\rho_{t'}\nabla_{\theta_i}\rho_t^\top+\nabla_{\theta_i}\rho_t\nabla_{\theta_i}\rho_{t'}^\top+(\rho_{t'}-1)\,\nabla^2_{\theta_i}\rho_t.
\end{equation*}
At $\theta^0$, $\rho_{t'}-1=\rho_t-1=0$, so only the middle terms remain:
\begin{align*}
\nabla^2_{\theta_i\theta_i}J(\pi_\theta)
&=\mathbb{E}_\pi\Big[\sum_{t:\,i_t=i}\nabla^2_{\theta_i}\rho_t\,A_t^\pi(s_t,a_t)+\sum_{t'<t:\,i_{t'}=i_t=i}\big(\nabla_{\theta_i}\rho_{t'}\nabla_{\theta_i}\rho_t^\top+\nabla_{\theta_i}\rho_t\nabla_{\theta_i}\rho_{t'}^\top\big)A_t^\pi(s_t,a_t)\Big]\\
&\overset{(a)}{=}\mathbb{E}_\pi\Big[\sum_{t:\,i_t=i}\big(\nabla^2_{\theta_i}\log\pi_\theta(a_t\mid s_t)+g_tg_t^\top\big)A_t^\pi(s_t,a_t)+\sum_{t'<t:\,i_{t'}=i_t=i}(g_{t'}g_t^\top+g_tg_{t'}^\top)A_t^\pi(s_t,a_t)\Big],
\end{align*}
where (a) uses \eqref{eq:pf-score} and \eqref{eq:pf-ratio-hess}.
\end{proof}

\begin{proof}[Proof of Theorem~\ref{thm:hessian}]
It is Theorem~\ref{thm:app-hessian}(a), (b), and (c).
\end{proof}

\begin{proof}[Proof of Corollary~\ref{cor:curvature}]
With all workers fixed, only the orchestrator has parameters, and full
support of $\pi$ gives Assumption~\ref{ass:overlap} for $\pi'=\pi_\theta$ with
$W=\max_{s,a}\pi_\theta(a\mid s)/\pi(a\mid s)<\infty$. By
Theorem~\ref{thm:app-expansion}(i):
\begin{equation*}
J(\pi_\theta)=J(\pi)+\sum_{m=1}^HJ^{(m)}(\theta).
\end{equation*}
$J(\pi)$ does not depend on $\theta$, and each $J^{(m)}(\theta)$ is a finite
sum of expectations under $\pi$ of products of $m$ factors $\rho_t(\theta)-1$
times $A_{t_m}^\pi$, which does not depend on $\theta$. Every factor vanishes
at $\theta^0$, so Lemma~\ref{lem:vanishing} gives:
\begin{equation*}
\nabla_\theta J^{(m)}(\theta^0)=0\ \ (m\ge2),\qquad
\nabla_\theta^2J^{(m)}(\theta^0)=0\ \ (m\ge3).
\end{equation*}
Differentiating the finite sum at $\theta^0$:
\begin{equation*}
\nabla_\theta J(\pi_\theta)=\nabla_\theta J^{(1)}(\theta),\qquad
\nabla_\theta^2J(\pi_\theta)=\nabla_\theta^2J^{(1)}(\theta)+\nabla_\theta^2J^{(2)}(\theta).
\end{equation*}
\end{proof}

By Theorem~\ref{thm:app-hessian}, for $v\to0$:
\begin{equation*}
J(\pi_{\theta^0+v})-J(\pi)=\sum_i\Big[\nabla L_i^\top v_i+\tfrac12v_i^\top\nabla^2(L_i+I_{i\to i})v_i\Big]
+\sum_{i<i'}v_i^\top\nabla^2_{\theta_i\theta_{i'}}(I_{i\to i'}+I_{i'\to i})\,v_{i'}+o(\|v\|^2).
\end{equation*}
The first-order objective $J^{(1)}=\sum_iL_i$ keeps $\nabla L_i^\top v_i$ and
$\frac12v_i^\top\nabla^2L_iv_i$ but misses every $I_{i\to i}$ and every cross
block; with frozen workers ($v_i=0$ for $i\neq0$) it misses $I_{0\to0}$.

\subsection{Per-Agent Rules Are Blind}
\label{app:blind}

For $\omega\in\{0,1\}^2$, let $\Delta(\omega)$ be the gain of the policy in
which agent $i$ adopts $\pi'^i$ if $\omega_i=1$ and keeps $\pi^i$ otherwise.

\begin{proposition}[Per-agent rules are blind]
\label{prop:blind}
There exist two problems, each with two agents that act once, with identical
$(L_i,\alpha_i,n_i)_{i=1,2}$ and gains:
\begin{equation*}
\Delta(\omega)=\omega_1L_1+\omega_2L_2+\omega_1\omega_2I_{1\to2},
\end{equation*}
where $I_{1\to2}>0$ in the first problem and $I_{1\to2}<0$ in the second.
Their best gains are $\Gamma>0$ and $0$. Every rule $\hat\omega$ that
depends only on $(L_i,\alpha_i,n_i)_{i=1,2}$ has expected regret
$\mathbb{E}[\max_\omega\Delta(\omega)-\Delta(\hat\omega)]\ge\Gamma/2$ on at
least one of the two problems that it cannot tell apart.
\end{proposition}

\begin{proof}
Fix $p\in(0,1)$, $\alpha\in(0,1-p)$, and $\zeta,\gamma>0$ with
$\gamma\alpha>2\zeta$, and set:
\begin{equation*}
\Gamma\coloneqq\alpha(\gamma\alpha-2\zeta)>0.
\end{equation*}
Agent $1$ draws $x_1\sim\mathrm{Bern}(\nu_1)$, then agent $2$ draws
$x_2\sim\mathrm{Bern}(\nu_2)$ without seeing $x_1$, and the terminal
reward is:
\begin{equation*}
-\zeta(x_1+x_2)+\gamma(x_1-p)(x_2-p)\ \text{ in }\mathcal{M}^+,\qquad
-\zeta(x_1+x_2)-\gamma(x_1-p)(x_2-p)\ \text{ in }\mathcal{M}^-.
\end{equation*}
By independence, $\mathbb{E}[(x_1-p)(x_2-p)]=(\nu_1-p)(\nu_2-p)$, so:
\begin{equation*}
J^\pm(\nu_1,\nu_2)=-\zeta(\nu_1+\nu_2)\pm\gamma(\nu_1-p)(\nu_2-p).
\end{equation*}
The current policy is $(p,p)$, the candidate is $(p+\alpha,p+\alpha)$, the
two problems of Proposition~\ref{prop:blind} are $\mathcal{M}^\pm$ with gains
$\Delta^\pm$, and $\pi_\omega=(p+\omega_1\alpha,\,p+\omega_2\alpha)$ switches
agent $i$ to the candidate iff $\omega_i=1$, so:
\begin{equation}
\Delta^\pm(\omega)=J^\pm(p+\omega_1\alpha,p+\omega_2\alpha)-J^\pm(p,p)
=-\zeta\alpha\,\omega_1-\zeta\alpha\,\omega_2\pm\gamma\alpha^2\,\omega_1\omega_2.
\label{eq:pf-blind-gains}
\end{equation}

Each agent acts once, so $n_1=n_2=1$ and
$\alpha_1=\alpha_2=\mathrm{TV}(\mathrm{Bern}(p),\mathrm{Bern}(p+\alpha))=\frac12(\alpha+\alpha)=\alpha$.
For $\omega=\mathbf{e}_1$, Theorem~\ref{thm:app-expansion}(ii) keeps only
$k_2=0$ and (iv) with $n_1=1$ keeps only $\mathbf{k}=\mathbf{e}_1$, so
$L_1=J_{\mathbf{e}_1}=\Delta^\pm(\mathbf{e}_1)=-\zeta\alpha$; likewise
$L_2=-\zeta\alpha$. Hence $(L_i,\alpha_i,n_i)_{i=1,2}$ is the same in
$\mathcal{M}^+$ and $\mathcal{M}^-$. For a general $\omega$,
Theorem~\ref{thm:app-expansion}(ii) and (iv) leave only
$\mathbf{k}\in\{\mathbf{e}_1,\mathbf{e}_2,\mathbf{e}_1+\mathbf{e}_2\}$, so:
\begin{equation*}
\Delta^\pm(\omega)=\omega_1L_1+\omega_2L_2+\omega_1\omega_2\big(I_{1\to2}+I_{2\to1}\big).
\end{equation*}
Agent $2$ never acts before agent $1$, so $I_{2\to1}=0$, and
\eqref{eq:pf-blind-gains} gives $I_{1\to2}=\pm\gamma\alpha^2$; hence
$\Delta^\pm(\omega)=J^{(1)}+J^{(2)}$ at $\pi_\omega$.

The profiles $\mathbf{0},\mathbf{e}_1,\mathbf{e}_2,(1,1)$ have gains
$0,-\zeta\alpha,-\zeta\alpha,-2\zeta\alpha\pm\gamma\alpha^2$, and:
\begin{equation*}
\Delta^+(1,1)=-2\zeta\alpha+\gamma\alpha^2=\alpha(\gamma\alpha-2\zeta)=\Gamma>0,\qquad
\Delta^-(1,1)=-\alpha(\gamma\alpha+2\zeta)<0.
\end{equation*}
Hence $\max_\omega\Delta^+(\omega)=\Delta^+(1,1)=\Gamma$ and
$\max_\omega\Delta^-(\omega)=\Delta^-(\mathbf{0})=0$.

Since $\hat\omega$ receives identical inputs in $\mathcal{M}^\pm$,
$\Pr(\hat\omega=(1,1))=\varphi$ in both. In $\mathcal{M}^+$,
$\Delta^+(\omega)\le0$ for $\omega\neq(1,1)$, so the regret is $\ge\Gamma$ on $\{\hat\omega\neq(1,1)\}$; in $\mathcal{M}^-$, $(1,1)$ has
regret $\alpha(\gamma\alpha+2\zeta)\ge\Gamma$. The expected regrets are at
least $(1-\varphi)\Gamma$ and $\varphi\Gamma$, so:
\begin{equation*}
\max_\pm\mathbb{E}\Big[\max_\omega\Delta^\pm(\omega)-\Delta^\pm(\hat\omega)\Big]
\ge\max\{(1-\varphi)\Gamma,\,\varphi\Gamma\}\ge\frac12\big((1-\varphi)\Gamma+\varphi\Gamma\big)=\frac\Gamma2.
\end{equation*}
\end{proof}

Per-agent inputs coincide for a complementary and a conflicting pair, which
differ only in the sign of $I_{1\to2}$; the order-two score sees this sign
and, being exact here, selects the best profile in both problems.

\section{Proof of Theorem~\ref{thm:visit} (Visit-Count Bound)}
\label{app:visit}

\subsection{A Binomial Identity}
\label{app:binomial}

For a count vector $\mathbf{m}=(m_i)_{i\in\mathcal{N}}$ and
$\mathbf{k}\in\mathbb{Z}^{\mathcal{N}}$, define:
\begin{equation*}
B(\mathbf{k};\mathbf{m})\coloneqq\prod_{i\in\mathcal{N}}\binom{m_i}{k_i}(2\alpha_i)^{k_i},
\end{equation*}
with $B(\mathbf{k};\mathbf{m})\coloneqq0$ if $k_i<0$ or $k_i>m_i$ for some
$i$, and $0^0\coloneqq1$.

\begin{lemma}[Properties of $B$]
\label{lem:B}
\begin{enumerate}
\item[(i)] If $m_i\ge1$, then
$B(\mathbf{k};\mathbf{m})=B(\mathbf{k};\mathbf{m}-\mathbf{e}_i)+2\alpha_i\,B(\mathbf{k}-\mathbf{e}_i;\mathbf{m}-\mathbf{e}_i)$.
\item[(ii)] If $\mathbf{m}'\le\mathbf{m}$ entrywise, then
$B(\mathbf{k};\mathbf{m}')\le B(\mathbf{k};\mathbf{m})$.
\item[(iii)] For every $r\ge1$,
$\sum_{|\mathbf{k}|=r}B(\mathbf{k};\mathbf{m})\le\frac1{r!}\big(2\sum_im_i\alpha_i\big)^r$.
\end{enumerate}
\end{lemma}

\begin{proof}
(i) Only the factor of agent $i$ differs, and Pascal's rule
$\binom mk=\binom{m-1}k+\binom{m-1}{k-1}$ gives:
\begin{equation*}
\binom{m_i}{k_i}(2\alpha_i)^{k_i}=\binom{m_i-1}{k_i}(2\alpha_i)^{k_i}+2\alpha_i\binom{m_i-1}{k_i-1}(2\alpha_i)^{k_i-1}.
\end{equation*}
Multiply by $\prod_{i'\neq i}\binom{m_{i'}}{k_{i'}}(2\alpha_{i'})^{k_{i'}}$; for
$k_i=0$ the last term is $0$.

(ii) $m\mapsto\binom mk$ is nondecreasing and every factor is $\ge0$.

(iii) With $x_i\coloneqq2m_i\alpha_i$:
\begin{equation*}
\sum_{|\mathbf{k}|=r}B(\mathbf{k};\mathbf{m})\overset{(a)}{\le}\sum_{|\mathbf{k}|=r}\prod_i\frac{(m_i)^{k_i}(2\alpha_i)^{k_i}}{k_i!}
=\frac1{r!}\sum_{|\mathbf{k}|=r}\frac{r!}{\prod_ik_i!}\prod_ix_i^{k_i}\overset{(b)}{=}\frac1{r!}\Big(\sum_ix_i\Big)^r,
\end{equation*}
where (a) uses $\binom mk\le m^k/k!$ and (b) is the multinomial theorem.
\end{proof}

\subsection{The Coefficient Bound}
\label{app:coef-bound}

\begin{trivlist}\item[]\textbf{Theorem~\ref{thm:visit} (Visit-count bound).}\ \itshape
Under Assumption~\ref{ass:overlap}, with workers fixed, for every integer
$r\ge1$:
\begin{equation*}
\Big|J(\pi')-J(\pi)-\sum_{m=1}^rJ^{(m)}\Big|\le\bar G\sum_{m>r}\frac{\beta^m}{m!}.
\tag{\ref{eq:tail}}
\end{equation*}
\end{trivlist}

We prove a stronger, per-coefficient bound.

\begin{theorem}[Visit-count bound, general form]
\label{thm:app-visit}
Under Assumption~\ref{ass:overlap}, for every $\mathbf{k}\neq\mathbf{0}$:
\begin{equation}
|J_{\mathbf{k}}|\le\min\Big\{\sum_{t=1}^H\bar r_t\prod_i\binom{n_{i,t}}{k_i}(2\alpha_i)^{k_i},\ \ \bar G\prod_i\binom{n_i}{k_i}(2\alpha_i)^{k_i}\Big\}.
\label{eq:coef-bound}
\end{equation}
Consequently, for every $r\ge1$:
\begin{equation}
\Big|J(\pi')-J(\pi)-\sum_{m=1}^rJ^{(m)}\Big|
\le\min\Big\{\sum_t\bar r_t\sum_{|\mathbf{k}|>r}B\big(\mathbf{k};(n_{i,t})_i\big),\ \ \bar G\sum_{|\mathbf{k}|>r}B\big(\mathbf{k};(n_i)_i\big)\Big\}
\le\bar G\sum_{m>r}\frac1{m!}\Big(2\sum_{i\in\mathcal N}n_i\alpha_i\Big)^m.
\label{eq:app-tail}
\end{equation}
\end{theorem}

\begin{proof}
Fix a reward step $t$. For $t'\in[t]$ define:
\begin{equation*}
\Phi_{t'}(\mathbf{k})\coloneqq\mathbb{E}_\pi\Big[\sum_{S\subseteq\{t',\dots,t\}:\,\mathbf{k}(S)=\mathbf{k}}\ \prod_{t''\in S}|\rho_{t''}-1|\ \Big|\ \tau_{t'-1},s_{t'}\Big],
\qquad\Phi_{t+1}(\mathbf{k})\coloneqq\mathbf{1}\{\mathbf{k}=\mathbf{0}\},
\end{equation*}
and let $m_i(s_{t'})$ be the maximum of
$|\{t''\in\{t',\dots,t\}:i_{t''}=i\}|$ over trajectories of $\pi$ through
$s_{t'}$. If $s_{t'}$ is absorbing, then $\rho_{t'}=1$, every $S\ni t'$
contributes $0$, $\Phi_{t'}(\mathbf{k})=\mathbb{E}_\pi[\Phi_{t'+1}(\mathbf{k})\mid\tau_{t'-1},s_{t'}]$,
and $\mathbf{m}(s_{t'})\ge\mathbf{m}(s_{t'+1})$, so the claim
\eqref{eq:pf-visit-claim} below passes from $t'+1$ to $t'$ by
Lemma~\ref{lem:B}(ii). Otherwise, let $i=i_{t'}$. Splitting by $t'\notin S$ ($S\subseteq\{t'+1,\dots,t\}$,
same counts) and $S=\{t'\}\cup S'$ ($S'\subseteq\{t'+1,\dots,t\}$,
$\mathbf{k}(S')=\mathbf{k}-\mathbf{e}_i$):
\begin{equation*}
\sum_{\substack{S\subseteq\{t',\dots,t\}\\\mathbf{k}(S)=\mathbf{k}}}\prod_{t''\in S}|\rho_{t''}-1|
=\sum_{\substack{S'\subseteq\{t'+1,\dots,t\}\\\mathbf{k}(S')=\mathbf{k}}}\prod_{t''\in S'}|\rho_{t''}-1|
+|\rho_{t'}-1|\sum_{\substack{S'\subseteq\{t'+1,\dots,t\}\\\mathbf{k}(S')=\mathbf{k}-\mathbf{e}_i}}\prod_{t''\in S'}|\rho_{t''}-1|.
\end{equation*}
Taking $\mathbb{E}_\pi[\,\cdot\mid\tau_{t'-1},s_{t'}]$, with the tower rule over
$(a_{t'},s_{t'+1})$ and $|\rho_{t'}-1|$ a function of $(s_{t'},a_{t'})$:
\begin{equation}
\Phi_{t'}(\mathbf{k})=\mathbb{E}_\pi\big[\Phi_{t'+1}(\mathbf{k})\mid\tau_{t'-1},s_{t'}\big]+\mathbb{E}_\pi\big[|\rho_{t'}-1|\,\Phi_{t'+1}(\mathbf{k}-\mathbf{e}_i)\mid\tau_{t'-1},s_{t'}\big].
\label{eq:pf-visit-recursion}
\end{equation}
Since states determine their history, every trajectory of $\pi$ through a next
state $s_{t'+1}$ passes through $s_{t'}$ and has $i_{t'}=i$, so its counts on
$\{t',\dots,t\}$ are its counts on $\{t'+1,\dots,t\}$ plus $\mathbf{e}_i$.
Maximizing:
\begin{equation}
\mathbf{m}(s_{t'})\ge\mathbf{m}(s_{t'+1})+\mathbf{e}_i,\qquad\text{in particular }m_i(s_{t'})\ge1.
\label{eq:pf-visit-counts}
\end{equation}

We show by backward induction on $t'$ that, for all $t'$ and $\mathbf{k}$:
\begin{equation}
\Phi_{t'}(\mathbf{k})\le B\big(\mathbf{k};\mathbf{m}(s_{t'})\big).
\label{eq:pf-visit-claim}
\end{equation}
At $t'=t+1$, $\mathbf{m}=\mathbf{0}$ and
$B(\mathbf{k};\mathbf{0})=\mathbf{1}\{\mathbf{k}=\mathbf{0}\}=\Phi_{t+1}(\mathbf{k})$.
Given \eqref{eq:pf-visit-claim} at $t'+1$, \eqref{eq:pf-visit-counts} and
Lemma~\ref{lem:B}(ii) give, for every next state $s_{t'+1}$ of a trajectory of
$\pi$ through $s_{t'}$:
\begin{equation*}
\Phi_{t'+1}(\mathbf{k})\le B\big(\mathbf{k};\mathbf{m}(s_{t'+1})\big)\le B\big(\mathbf{k};\mathbf{m}(s_{t'})-\mathbf{e}_i\big),\qquad
\Phi_{t'+1}(\mathbf{k}-\mathbf{e}_i)\le B\big(\mathbf{k}-\mathbf{e}_i;\mathbf{m}(s_{t'})-\mathbf{e}_i\big).
\end{equation*}
These bounds are free of $(a_{t'},s_{t'+1})$, so \eqref{eq:pf-visit-recursion}
gives:
\begin{align*}
\Phi_{t'}(\mathbf{k})
&\le B\big(\mathbf{k};\mathbf{m}(s_{t'})-\mathbf{e}_i\big)+B\big(\mathbf{k}-\mathbf{e}_i;\mathbf{m}(s_{t'})-\mathbf{e}_i\big)\,\mathbb{E}_\pi\big[|\rho_{t'}-1|\mid\tau_{t'-1},s_{t'}\big]\\
&\overset{(a)}{\le}B\big(\mathbf{k};\mathbf{m}(s_{t'})-\mathbf{e}_i\big)+2\alpha_iB\big(\mathbf{k}-\mathbf{e}_i;\mathbf{m}(s_{t'})-\mathbf{e}_i\big)
\overset{(b)}{=}B\big(\mathbf{k};\mathbf{m}(s_{t'})\big),
\end{align*}
where (a) uses Lemma~\ref{lem:center} and (b) is Lemma~\ref{lem:B}(i), valid
since $m_i(s_{t'})\ge1$ by \eqref{eq:pf-visit-counts}.

By \eqref{eq:pf-exp-return} and \eqref{eq:pf-exp-q}, which hold for each
$\mathbf{k}$ separately, the reward form \eqref{eq:reward-form} holds termwise:
\begin{equation*}
J_{\mathbf{k}}=\sum_{t=1}^H\mathbb{E}_\pi\Big[\sum_{S\subseteq[t]:\,\mathbf{k}(S)=\mathbf{k}}\ \prod_{t''\in S}(\rho_{t''}-1)\ r_t(s_t,a_t)\Big].
\end{equation*}
First bound in \eqref{eq:coef-bound}:
\begin{equation*}
|J_{\mathbf{k}}|\overset{(a)}{\le}\sum_{t=1}^H\bar r_t\,\mathbb{E}_\pi\big[\Phi_1(\mathbf{k})\big]
\overset{(b)}{\le}\sum_{t=1}^H\bar r_t\,\mathbb{E}_\pi\big[B(\mathbf{k};\mathbf{m}(s_1))\big]
\overset{(c)}{\le}\sum_{t=1}^H\bar r_t\prod_i\binom{n_{i,t}}{k_i}(2\alpha_i)^{k_i},
\end{equation*}
where (a) uses $|r_t|\le\bar r_t$, (b) is \eqref{eq:pf-visit-claim} at $t'=1$
for reward step $t$, and (c) uses $m_i(s_1)\le n_{i,t}$ and
Lemma~\ref{lem:B}(ii).

Second bound in \eqref{eq:coef-bound}: let
$X_t\coloneqq\sum_{S\subseteq[t],\,\mathbf{k}(S)=\mathbf{k}}\prod_{t''\in S}|\rho_{t''}-1|$.
The summands are $\ge0$ and $\{S\subseteq[t]\}\subseteq\{S\subseteq[t+1]\}$, so
$X_t\le X_H$, and the integrand of the reward form satisfies:
\begin{equation*}
\sum_{t=1}^H|r_t(s_t,a_t)|\,X_t\le X_H\sum_{t=1}^H|r_t(s_t,a_t)|\le\bar G\,X_H.
\end{equation*}
By this display, \eqref{eq:pf-visit-claim} at $t'=1$ with reward step $H$,
and Lemma~\ref{lem:B}(ii):
\begin{equation*}
|J_{\mathbf{k}}|\le\bar G\,\mathbb{E}_\pi[X_H]=\bar G\,\mathbb{E}_\pi[\Phi_1(\mathbf{k})]\le\bar G\prod_i\binom{n_i}{k_i}(2\alpha_i)^{k_i}.
\end{equation*}

Tail: by Theorem~\ref{thm:app-expansion}(i) and \eqref{eq:order-by-counts},
$J(\pi')-J(\pi)-\sum_{m\le r}J^{(m)}=\sum_{|\mathbf{k}|>r}J_{\mathbf{k}}$, and
the triangle inequality with \eqref{eq:coef-bound} gives both terms of the
minimum in \eqref{eq:app-tail}. For the second term:
\begin{equation*}
\bar G\sum_{|\mathbf{k}|>r}B\big(\mathbf{k};(n_i)_i\big)=\bar G\sum_{m>r}\sum_{|\mathbf{k}|=m}B\big(\mathbf{k};(n_i)_i\big)\overset{(a)}{\le}\bar G\sum_{m>r}\frac1{m!}\Big(2\sum_in_i\alpha_i\Big)^m,
\end{equation*}
where (a) is Lemma~\ref{lem:B}(iii) with $\mathbf{m}=(n_i)_i$.
\end{proof}

\begin{proof}[Proof of Theorem~\ref{thm:visit} and of \eqref{eq:lower-bound}]
With workers fixed, $\alpha_j=0$ for $j\in[M]$, so
$2\sum_in_i\alpha_i=2n_0\alpha_0\le\beta$ by $n_0\le K+1$. The last
inequality of \eqref{eq:app-tail} and the monotonicity of
$x\mapsto\sum_{m>r}x^m/m!$ on $x\ge0$ give Theorem~\ref{thm:visit}. For
$r=2$, $\sum_{m>2}\beta^m/m!=e^\beta-1-\beta-\beta^2/2$, so \eqref{eq:tail}
gives
$J(\pi')-J(\pi)\ge J^{(1)}+J^{(2)}-\bar G(e^\beta-1-\beta-\beta^2/2)$, which
is \eqref{eq:lower-bound}.
\end{proof}

\begin{remark}[Two agents that never meet]
If $i\neq i'$ never act on a common trajectory of $\pi$, no $S$ has
$\mathbf{k}(S)=\mathbf{e}_i+\mathbf{e}_{i'}$, so $I_{i\to i'}+I_{i'\to i}=0$ by
Theorem~\ref{thm:app-expansion}(iv). Otherwise, Theorem~\ref{thm:app-expansion}(v)
and \eqref{eq:coef-bound} with
$\mathbf{k}=\mathbf{e}_i+\mathbf{e}_{i'}$ gives
$|I_{i\to i'}+I_{i'\to i}|\le\bar G\,n_i(2\alpha_i)\,n_{i'}(2\alpha_{i'})=4\bar Gn_in_{i'}\alpha_i\alpha_{i'}$.
\end{remark}

\subsection{Depth of the Expansion}
\label{app:depth}

\begin{corollary}[Depth]
\label{cor:depth}
If every trajectory of $\pi$ contains at most $r$ steps of changed agents,
then $J^{(m)}=0$ for all $m>r$.
\end{corollary}

\begin{proof}
On any trajectory of $\pi$, at most $r$ steps belong to changed agents, so
$|S|>r$ implies that some $t'\in S$ has
$\pi'^{i_{t'}}=\pi^{i_{t'}}$, hence $\rho_{t'}=1$ and
$\prod_{t\in S}(\rho_t-1)=0$. So every term of $J^{(m)}$ in
\eqref{eq:order-m} with $m>r$ vanishes.
\end{proof}

By Corollary~\ref{cor:depth}, a one-shot router or workflow designer has $r=1$
($J^{(1)}$ is exact), a fixed pipeline with each trained agent acting once
has $r=|\{\text{trained agents}\}|$ and only terms across agents, and a
multi-turn orchestrator has $r=K+1$.

\section{Proofs for the Algorithm}
\label{app:algorithm}

Throughout, $\pi'_{\theta,\eta}$ applies \eqref{eq:bounded-policy} to the
orchestrator and keeps $\pi^1,\dots,\pi^M$, so $\rho_t=1$ whenever $i_t\neq0$.

\subsection{The Bounded Update}
\label{app:bounded}

\begin{trivlist}\item[]\textbf{Proposition~\ref{prop:bounded} (Bounded ratios).}\ \itshape
For any $\theta$ and $\eta\in[0,1]$,
\eqref{eq:bounded-policy} defines a valid
action distribution at every orchestrator
state $s$, satisfying
$\pi'_{\theta,\eta}(a\mid s)\ge\pi(a\mid s)^2$.
Moreover, on the support of $\pi$, its
likelihood ratio satisfies
$|\rho_t-1|\le\eta$, and its total variation
distance from $\pi(\cdot\mid s)$ is at most
$\eta/4$.
\end{trivlist}

\begin{proof}
Fix an orchestrator state $s$ and write:
\begin{equation*}
p_a\coloneqq\pi(a\mid s),\qquad f_a\coloneqq f_\theta(s,a)\in[-1,1],\qquad
\bar f\coloneqq\sum_{a'}p_{a'}f_{a'},\qquad u_a\coloneqq\tfrac12(f_a-\bar f).
\end{equation*}
Then $\pi'_{\theta,\eta}(a\mid s)=p_a(1+\eta u_a)$ and:
\begin{equation*}
\sum_ap_au_a=\frac12\Big(\sum_ap_af_a-\bar f\sum_ap_a\Big)=\frac12(\bar f-\bar f)=0,\qquad
\sum_a\pi'_{\theta,\eta}(a\mid s)=\sum_ap_a+\eta\sum_ap_au_a=1.
\end{equation*}
Since $f_a-\bar f=(1-p_a)f_a-\sum_{a'\neq a}p_{a'}f_{a'}$ with
$|(1-p_a)f_a|\le1-p_a$ and $|\sum_{a'\neq a}p_{a'}f_{a'}|\le1-p_a$, and
$\eta\le1$:
\begin{equation*}
|f_a-\bar f|\le2(1-p_a),\qquad|u_a|\le1-p_a\le1,\qquad
\pi'_{\theta,\eta}(a\mid s)=p_a(1+\eta u_a)\ge p_a\big(1-(1-p_a)\big)=p_a^2.
\end{equation*}
Hence $\pi'_{\theta,\eta}(\cdot\mid s)$ is a distribution on $\mathcal{A}(s)$,
positive wherever $\pi(\cdot\mid s)$ is. If $(s_t,a_t)=(s,a)$ with $i_t=0$,
then $\rho_t=\pi'_{\theta,\eta}(a\mid s)/p_a=1+\eta u_a$, so
$|\rho_t-1|=\eta|u_a|\le\eta$. Finally:
\begin{align*}
\mathrm{TV}&=\frac12\sum_a|p_a\,\eta u_a|=\frac\eta4\sum_ap_a|f_a-\bar f|
\overset{(a)}{\le}\frac\eta4\Big(\sum_ap_a(f_a-\bar f)^2\Big)^{1/2}\\
&=\frac\eta4\Big(\sum_ap_af_a^2-\bar f^2\Big)^{1/2}\overset{(b)}{\le}\frac\eta4,
\end{align*}
where (a) is the Cauchy--Schwarz inequality with weights $p_a$ and (b) uses
$f_a^2\le1$.
\end{proof}

Uniformly over $\theta$, every ratio is a known function of $(\theta,\eta)$
and the logged probabilities of $\pi$, Assumption~\ref{ass:overlap} holds with
$W=1+\eta$, and $\alpha_0\le\eta/4$, $\alpha_j=0$ for $j\in[M]$.

\subsection{The First Two Orders and the Remainder}
\label{app:quadratic}

\begin{lemma}[Quadratic form and remainder]
\label{lem:quadratic}
For the bounded update with parameters $(\theta,\eta)$, with $\mu_1,\mu_2$ as
in \eqref{eq:quadratic}:
\begin{equation*}
J^{(1)}=\eta\,\mu_1(\theta),\qquad J^{(2)}=\eta^2\mu_2(\theta),\qquad
\big|J(\pi'_{\theta,\eta})-J(\pi)-\eta\mu_1(\theta)-\eta^2\mu_2(\theta)\big|\le\tilde\epsilon(\eta)\le\epsilon(\eta),
\end{equation*}
where $\epsilon(\eta)$ is \eqref{eq:remainder} and:
\begin{equation*}
\tilde\epsilon(\eta)\coloneqq\min\Big\{\sum_{t=1}^H\bar r_t\sum_{m\ge3}\binom{n_{0,t}}m\Big(\frac\eta2\Big)^m,\ \ \bar G\sum_{m\ge3}\binom{K+1}m\Big(\frac\eta2\Big)^m\Big\}.
\end{equation*}
Moreover, $\epsilon(\eta)\le\bar G(e^x-1-x-x^2/2)$ with $x=(K+1)\eta/2$.
\end{lemma}

\begin{proof}
For all $t$, $\rho_t-1=\eta u_t$, with $u_t=u_\theta(s_t,a_t)$ if $i_t=0$ and
$\rho_t-1=u_t=0$ if $i_t\neq0$.
Theorem~\ref{thm:app-expansion}(iii) with $\psi=0$ (Remark~\ref{rem:critic})
for $|S|\in\{1,2\}$, with $S=\{t',t\}$, $t'<t$, $t_S=t$ when $|S|=2$, gives:
\begin{equation}
\label{eq:pf-quad-orders}
\begin{aligned}
J^{(1)}&=\mathbb{E}_\pi\Big[\sum_{t=1}^H(\rho_t-1)G_t\Big]=\eta\,\mathbb{E}_\pi\Big[\sum_{t=1}^Hu_tG_t\Big]=\eta\mu_1(\theta),\\
J^{(2)}&=\mathbb{E}_\pi\Big[\sum_{t'<t}(\rho_{t'}-1)(\rho_t-1)G_t\Big]=\eta^2\,\mathbb{E}_\pi\Big[\sum_{t'<t}u_{t'}u_tG_t\Big]=\eta^2\mu_2(\theta).
\end{aligned}
\end{equation}
By Proposition~\ref{prop:bounded}, $\alpha_0\le\eta/4$ and $\alpha_j=0$ for
$j\in[M]$, so $B(\mathbf{k};\cdot)=0$ unless $\mathbf{k}=m\,\mathbf{e}_0$, with
orchestrator visit counts $n_{0,t}$ and $n_0\le K+1$. Hence:
\begin{align*}
\big|J(\pi'_{\theta,\eta})-J(\pi)-J^{(1)}-J^{(2)}\big|
&\overset{(a)}{\le}\min\Big\{\sum_t\bar r_t\sum_{m\ge3}\binom{n_{0,t}}m\Big(\frac\eta2\Big)^m,\ \bar G\sum_{m\ge3}\binom{n_0}m\Big(\frac\eta2\Big)^m\Big\}\\
&\overset{(b)}{\le}\tilde\epsilon(\eta),
\end{align*}
where (a) is Theorem~\ref{thm:app-visit} with $r=2$ and
$2\alpha_0\le\eta/2$, and (b) uses $n_0\le K+1$ and Lemma~\ref{lem:B}(ii). The
second entry of $\tilde\epsilon(\eta)$ equals $\epsilon(\eta)$, as
$\binom{K+1}m=0$ for $m>K+1$. By $\binom{K+1}m\le(K+1)^m/m!$:
\begin{equation*}
\epsilon(\eta)\le\bar G\sum_{m\ge3}\frac{x^m}{m!}=\bar G\Big(e^x-1-x-\frac{x^2}2\Big).
\end{equation*}
\end{proof}

Since the proofs below use only the inequality of Lemma~\ref{lem:quadratic},
$\tilde\epsilon(\eta)$ can replace $\epsilon(\eta)$ in \eqref{eq:certificate}
with every guarantee intact.

\subsection{One-Pass Computation}
\label{app:onepass}

Lemma~\ref{lem:onepass} verifies Algorithm~\ref{alg:onepass}.

\begin{algorithm}[!ht]
\caption{One pass over a trajectory}
\label{alg:onepass}
\begin{algorithmic}[1]
\REQUIRE trajectory $(s_1,a_1,\dots,s_H,a_H)$, parameters $\theta$, step size $\eta$, baseline $b$
\STATE $G_{H+1}\gets0$; $G_t\gets r_t(s_t,a_t)+G_{t+1}$ for $t=H,\dots,1$
\STATE $U,\nabla U,Z_2,\nabla Z_2,Z_{\mathrm{IS}}\gets0$; $w\gets1$
\FOR{$t=1,\dots,H$}
\IF{$i_t=0$}
\STATE $\rho\gets1+\eta\,u_\theta(s_t,a_t)$; $\nabla\rho\gets\eta\nabla_\theta u_\theta(s_t,a_t)$
\ELSE
\STATE $\rho\gets1$; $\nabla\rho\gets0$
\ENDIF
\STATE $Z_2\gets Z_2+(\rho-1)(1+U)(G_t-b)$
\STATE $\nabla Z_2\gets\nabla Z_2+\big((1+U)\nabla\rho+(\rho-1)\nabla U\big)(G_t-b)$
\STATE $U\gets U+(\rho-1)$; $\nabla U\gets\nabla U+\nabla\rho$
\STATE $w\gets w\rho$; $Z_{\mathrm{IS}}\gets Z_{\mathrm{IS}}+(w-1)r_t(s_t,a_t)$
\ENDFOR
\STATE $Z_{\mathrm{IS}}\gets Z_{\mathrm{IS}}-b(w-1)$
\STATE \textbf{return} $Z_2,\nabla Z_2,Z_{\mathrm{IS}}$
\end{algorithmic}
\end{algorithm}

\begin{lemma}[One-pass statistic and gradient]
\label{lem:onepass}
Let $U_1=0$ and $U_{t+1}=U_t+(\rho_t-1)$. Then:
\begin{enumerate}
\item[(a)] $Z_2=\sum_t(\rho_t-1)(G_t-b)+\sum_{t'<t}(\rho_{t'}-1)(\rho_t-1)(G_t-b)$,
and $\mathbb{E}_\pi[Z_2]=\eta\mu_1(\theta)+\eta^2\mu_2(\theta)$ for every fixed
$b$;
\item[(b)] $\nabla_\theta Z_2=\sum_t\big[(1+U_t)\nabla_\theta\rho_t+(\rho_t-1)\nabla_\theta U_t\big](G_t-b)$
with $\nabla_\theta U_{t+1}=\nabla_\theta U_t+\nabla_\theta\rho_t$, and
$\nabla_\theta\rho_t=\frac\eta2\big[\nabla_\theta f_\theta(s_t,a_t)-\sum_{a'}\pi(a'\mid s_t)\nabla_\theta f_\theta(s_t,a')\big]$
at orchestrator steps and $\nabla_\theta\rho_t=0$ at worker steps;
\item[(c)] Algorithm~\ref{alg:onepass} returns $Z_2$, $\nabla_\theta Z_2$, and
$Z_{\mathrm{IS}}$, in time linear in $H$.
\end{enumerate}
\end{lemma}

\begin{proof}
By induction, $U_t=\sum_{t'<t}(\rho_{t'}-1)$, since $U_1=0$ and
$U_{t+1}=U_t+(\rho_t-1)=\sum_{t'\le t}(\rho_{t'}-1)$.

(a) By this identity and \eqref{eq:z2}:
\begin{equation*}
Z_2=\sum_t(\rho_t-1)(1+U_t)(G_t-b)=\sum_t(\rho_t-1)(G_t-b)+\sum_t\sum_{t'<t}(\rho_{t'}-1)(\rho_t-1)(G_t-b).
\end{equation*}
Since $1+U_t$ is a function of $\tau_{t-1}$, Lemma~\ref{lem:center} gives:
\begin{equation*}
\mathbb{E}_\pi\big[(\rho_t-1)(1+U_t)\,b\big]=b\,\mathbb{E}_\pi\big[(1+U_t)\,\mathbb{E}_\pi[\rho_t-1\mid\tau_{t-1},s_t]\big]=0,
\end{equation*}
so $\mathbb{E}_\pi[Z_2]=J^{(1)}+J^{(2)}=\eta\mu_1+\eta^2\mu_2$ by
\eqref{eq:pf-quad-orders}. For $\{t:i_t=0\}=\{1,3,5\}$ and $\rho_t-1=\eta u_t$:
\begin{equation*}
Z_2=\eta\big(u_1G_1+u_3G_3+u_5G_5\big)+\eta^2\big(u_1u_3G_3+u_1u_5G_5+u_3u_5G_5\big)
\end{equation*}
for $b=0$, which excludes the third-order term $\eta^3u_1u_3u_5G_5$.

(b) Only the ratios depend on $\theta$, so by the product rule:
\begin{equation}
\label{eq:pf-onepass-grad}
\nabla_\theta Z_2=\sum_t\big[(1+U_t)\nabla_\theta\rho_t+(\rho_t-1)\nabla_\theta U_t\big](G_t-b),
\qquad\nabla_\theta U_t=\sum_{t'<t}\nabla_\theta\rho_{t'},
\end{equation}
whence $\nabla_\theta U_{t+1}=\nabla_\theta U_t+\nabla_\theta\rho_t$. If
$i_t=0$, then $\rho_t=1+\eta u_\theta(s_t,a_t)$ and the weights
$\pi(a'\mid s_t)$ of the mean $\sum_{a'}\pi(a'\mid s_t)f_\theta(s_t,a')$ do not
depend on $\theta$, so:
\begin{equation}
\label{eq:pf-onepass-rho}
\nabla_\theta\rho_t=\eta\nabla_\theta u_\theta(s_t,a_t)=\frac\eta2\Big[\nabla_\theta f_\theta(s_t,a_t)-\sum_{a'}\pi(a'\mid s_t)\nabla_\theta f_\theta(s_t,a')\Big].
\end{equation}

(c) Line 1 of Algorithm~\ref{alg:onepass} computes $G_t$ backward from
$G_{H+1}=0$. At the start of iteration $t$ of the loop:
\begin{equation*}
U=U_t,\quad\nabla U=\nabla_\theta U_t,\quad w=\prod_{t'<t}\rho_{t'},\quad
Z_2=\sum_{t'<t}(\rho_{t'}-1)(1+U_{t'})(G_{t'}-b),\quad
Z_{\mathrm{IS}}=\sum_{t'<t}\Big(\prod_{t''\le t'}\rho_{t''}-1\Big)r_{t'}(s_{t'},a_{t'}).
\end{equation*}
This holds at $t=1$ by the initialization. If it holds at $t$, iteration $t$
sets $(\rho,\nabla\rho)=(\rho_t,\nabla_\theta\rho_t)$ by
\eqref{eq:pf-onepass-rho}, adds the $t$-th terms of \eqref{eq:z2} and
\eqref{eq:pf-onepass-grad} with $(U,\nabla U)=(U_t,\nabla_\theta U_t)$, sets
$(U,\nabla U)=(U_{t+1},\nabla_\theta U_{t+1})$, and sets
$w=\prod_{t'\le t}\rho_{t'}$ before adding the $t$-th reward term of
\eqref{eq:zis}; so it holds at $t+1$. At $t=H+1$, $w=\prod_{t'\le H}\rho_{t'}$, and
the last line subtracts the baseline term of \eqref{eq:zis}. Each iteration
costs $O(1)$, so the time is $O(H)$.
\end{proof}

\subsection{Improving Steps Exist}
\label{app:improving}

\begin{trivlist}\item[]\textbf{Theorem~\ref{thm:improving} (Existence of improving steps).}\ \itshape
Fix $\theta$ and write $\mu_1=\mu_1(\theta)$ and $\mu_2=\mu_2(\theta)$.
(i) If $\mu_1>0$, then any $\eta\in(0,1]$ with
$\eta(|\mu_2|+\epsilon(1))\le\mu_1/2$ guarantees
$J(\pi'_{\theta,\eta})-J(\pi)\ge\eta\mu_1/2$.
(ii) If $\mu_1=0<\mu_2$, then any $\eta\in(0,1]$ with
$\eta\epsilon(1)\le\mu_2/2$ guarantees
$J(\pi'_{\theta,\eta})-J(\pi)\ge\eta^2\mu_2/2$.\par
Furthermore, if the score class can represent
$f_\theta(s_t,a)=Q_t^\pi(s_t,a)/\bar G$, then:
\begin{equation*}
\mu_1=\frac1{2\bar G}
\mathbb E_\pi\Big[
\sum_{t:\,i_t=0}
A_t^\pi(s_t,a_t)^2
\Big],
\end{equation*}
which is positive whenever the advantage is nonzero with positive probability
at an orchestrator decision.
\end{trivlist}

\begin{proof}
By Lemma~\ref{lem:quadratic}:
\begin{equation*}
J(\pi'_{\theta,\eta})-J(\pi)\ge\eta\mu_1+\eta^2\mu_2-\epsilon(\eta).
\end{equation*}
For $\eta\in[0,1]$ and $m\ge3$, $(\eta/2)^m=\eta^m2^{-m}\le\eta^32^{-m}$, so by
\eqref{eq:remainder}:
\begin{equation}
\label{eq:pf-improving-cubic}
\epsilon(\eta)\le\eta^3\,\bar G\sum_{m=3}^{K+1}\binom{K+1}m2^{-m}=\eta^3\epsilon(1).
\end{equation}
If $\mu_1>0$ and $\eta(|\mu_2|+\epsilon(1))\le\mu_1/2$:
\begin{equation*}
\eta\mu_1+\eta^2\mu_2-\epsilon(\eta)\overset{(a)}{\ge}\eta\mu_1-\eta^2|\mu_2|-\eta^2\epsilon(1)
=\eta\Big[\mu_1-\eta\big(|\mu_2|+\epsilon(1)\big)\Big]\overset{(b)}{\ge}\frac{\eta\mu_1}2,
\end{equation*}
where (a) uses $\eta^2\mu_2\ge-\eta^2|\mu_2|$, \eqref{eq:pf-improving-cubic},
and $\eta^3\le\eta^2$, and (b) uses $\eta(|\mu_2|+\epsilon(1))\le\mu_1/2$.

If $\mu_1=0<\mu_2$ and $\eta\epsilon(1)\le\mu_2/2$, then
\eqref{eq:pf-improving-cubic} gives:
\begin{equation*}
\eta^2\mu_2-\epsilon(\eta)\ge\eta^2\mu_2-\eta^3\epsilon(1)=\eta^2\big(\mu_2-\eta\epsilon(1)\big)\ge\frac{\eta^2\mu_2}2.
\end{equation*}

By Lemma~\ref{lem:bellman}(c), $|Q_t^\pi|\le\bar G$, so
$f_\theta(s_t,a)=Q_t^\pi(s_t,a)/\bar G\in[-1,1]$, a function of $(s,a)$ as $s$
determines $t$. By Lemma~\ref{lem:bellman}(b),
$\sum_{a'}\pi(a'\mid s_t)f_\theta(s_t,a')=V_t^\pi(s_t)/\bar G$, so if $i_t=0$:
\begin{equation*}
u_t=\frac12\Big(\frac{Q_t^\pi(s_t,a_t)}{\bar G}-\frac{V_t^\pi(s_t)}{\bar G}\Big)=\frac{A_t^\pi(s_t,a_t)}{2\bar G}.
\end{equation*}
Since $u_t$ is a function of $\tau_t$:
\begin{equation}
\label{eq:pf-improving-tower}
\mathbb{E}_\pi[u_tG_t]\overset{(a)}{=}\mathbb{E}_\pi\big[u_t\,Q_t^\pi(s_t,a_t)\big]\overset{(b)}{=}\mathbb{E}_\pi\big[u_tA_t^\pi(s_t,a_t)\big],
\end{equation}
where (a) uses the tower rule and Lemma~\ref{lem:bellman}(a), and (b) uses
$Q_t^\pi=A_t^\pi+V_t^\pi$ and, by Lemma~\ref{lem:bellman}(b):
\begin{equation*}
\mathbb{E}_\pi\big[A_t^\pi(s_t,a_t)\,V_t^\pi(s_t)\mid s_t\big]=V_t^\pi(s_t)\sum_a\pi_t(a\mid s_t)A_t^\pi(s_t,a)=0.
\end{equation*}
Summing \eqref{eq:pf-improving-tower} over $t$, with
$u_t=A_t^\pi(s_t,a_t)/(2\bar G)$ if $i_t=0$ and $u_t=0$ otherwise:
\begin{equation*}
\mu_1=\sum_t\mathbb{E}_\pi\big[u_tA_t^\pi(s_t,a_t)\big]=\frac1{2\bar G}\,\mathbb{E}_\pi\Big[\sum_{t:\,i_t=0}A_t^\pi(s_t,a_t)^2\Big],
\end{equation*}
which is positive unless $A_t^\pi(s_t,a_t)=0$ almost surely at every
orchestrator step.
\end{proof}

\subsection{Means and Ranges of the Statistics}
\label{app:ranges}

\begin{lemma}[Means and ranges]
\label{lem:ranges}
For the candidate $\pi'_\eta$, let $\Delta_\eta\coloneqq J(\pi'_\eta)-J(\pi)$,
and let $b$ be fixed. Then:
\begin{enumerate}
\item[(a)] $\mathbb{E}_\pi[Z_{\mathrm{IS}}]=\Delta_\eta$ and
$|\mathbb{E}_\pi[Z_2]-\Delta_\eta|\le\epsilon(\eta)$;
\item[(b)] $|Z_2|\le B_{2,\eta}$ and $|Z_{\mathrm{IS}}|\le B_{\mathrm{IS},\eta}$,
where:
\begin{equation*}
B_{2,\eta}\coloneqq(\bar G+|b|)\Big((K+1)\eta+\binom{K+1}2\eta^2\Big),\qquad
B_{\mathrm{IS},\eta}\coloneqq(\bar G+|b|)\big((1+\eta)^{K+1}-1\big).
\end{equation*}
\end{enumerate}
\end{lemma}

\begin{proof}
For (a), Lemma~\ref{lem:reweight} with $f=r_t(s_t,a_t)$ and with $f=1$ gives:
\begin{equation*}
\mathbb{E}_\pi\Big[\Big(\prod_{t'\le t}\rho_{t'}-1\Big)r_t(s_t,a_t)\Big]=\mathbb{E}_{\pi'_\eta}[r_t(s_t,a_t)]-\mathbb{E}_\pi[r_t(s_t,a_t)],\qquad
\mathbb{E}_\pi\Big[\prod_{t\le H}\rho_t-1\Big]=1-1=0.
\end{equation*}
Summing over $t$, Lemma~\ref{lem:prefix} gives
$\mathbb{E}_\pi[Z_{\mathrm{IS}}]=J(\pi'_\eta)-J(\pi)=\Delta_\eta$. For $Z_2$:
\begin{equation*}
\big|\mathbb{E}_\pi[Z_2]-\Delta_\eta\big|\overset{(a)}{=}\big|\eta\mu_1+\eta^2\mu_2-\Delta_\eta\big|\overset{(b)}{\le}\epsilon(\eta),
\end{equation*}
where (a) uses Lemma~\ref{lem:onepass}(a) and (b) uses Lemma~\ref{lem:quadratic}.

For (b), on every trajectory $|G_t-b|\le\bar G+|b|$,
$|\{t:\rho_t\neq1\}|\le K+1$ (one per orchestrator step), and
$|\rho_t-1|\le\eta$ by Proposition~\ref{prop:bounded}. By
Lemma~\ref{lem:onepass}(a), $Z_2$ has at most $K+1$ nonzero single terms and
$\binom{K+1}2$ nonzero pair terms, so:
\begin{equation*}
|Z_2|\le(\bar G+|b|)\Big((K+1)\eta+\binom{K+1}2\eta^2\Big)=B_{2,\eta}.
\end{equation*}
Since at most $K+1$ ratios differ from $1$ and
$\rho_{t'}\in[1-\eta,1+\eta]$,
$(1-\eta)^{K+1}\le\prod_{t'\le t}\rho_{t'}\le(1+\eta)^{K+1}$. By convexity of
$x\mapsto x^{K+1}$, $(1+\eta)^{K+1}+(1-\eta)^{K+1}\ge2$, so $|\prod_{t'\le t}\rho_{t'}-1|\le(1+\eta)^{K+1}-1$
and:
\begin{equation*}
|Z_{\mathrm{IS}}|\le\Big(\sum_t|r_t|+|b|\Big)\big((1+\eta)^{K+1}-1\big)\le(\bar G+|b|)\big((1+\eta)^{K+1}-1\big)=B_{\mathrm{IS},\eta}.
\end{equation*}
\end{proof}

\subsection{Certified Selection}
\label{app:certificate}

\begin{trivlist}\item[]\textbf{Theorem~\ref{thm:certificate} (Certified selection).}\ \itshape
With probability at least $1-\delta$, the following statements hold
simultaneously. (i) For every $\eta\in\Lambda$,
$L_\eta\le J(\pi'_\eta)-J(\pi)$; consequently, $J(\hat\pi)\ge J(\pi)$.
(ii) The selected policy satisfies:
\begin{equation*}
J(\hat\pi)-J(\pi)\ge\max\Big\{0,\,\max_{\eta\in\Lambda}\big[J(\pi'_\eta)-J(\pi)-2c_\eta\big]\Big\}.
\end{equation*}
(iii) If some $\eta\in\Lambda$ satisfies $J(\pi'_\eta)>J(\pi)$ and
$n-1>256B_{\mathrm{IS},\eta}^2\log(8|\Lambda|/\delta)/[J(\pi'_\eta)-J(\pi)]^2$,
then $\hat\pi\neq\pi$.
\end{trivlist}

\begin{lemma}[Empirical Bernstein bound \citep{maurer2009empirical}]
\label{lem:eb}
Let $Y_1,\dots,Y_n$, $n\ge2$, be i.i.d.\ copies of $Y\in[-B,B]$ with
sample mean $\bar Y$ and sample variance
$\hat\sigma^2=\frac1{n-1}\sum_{\ell=1}^n(Y_\ell-\bar Y)^2$. For every
$\delta'\in(0,1)$, with probability $\ge1-\delta'$:
\begin{equation*}
\mathbb{E}[Y]-\bar Y\le\hat\sigma\sqrt{\frac{2\log(2/\delta')}{n}}+\frac{14B\log(2/\delta')}{3(n-1)}.
\end{equation*}
\end{lemma}

This is Theorem~4 of \citet{maurer2009empirical} applied to
$(Y_\ell+B)/(2B)\in[0,1]$ and rescaled by $2B$.

\begin{proof}[Proof of Theorem~\ref{thm:certificate}]
Let $\iota\coloneqq\log(8|\Lambda|/\delta)$, so \eqref{eq:radius} reads
$\xi_\eta(Z)=\hat\sigma_Z\sqrt{2\iota/n}+14B\iota/(3(n-1))$ with $B=B_{Z,\eta}$. Set $\pi'_0\coloneqq\pi$
and $\Delta_0=L_0\coloneqq0$, so $\hat\pi=\pi'_{\hat\eta}$ with
$\hat\eta=\arg\max_{\eta\in\Lambda\cup\{0\}}L_\eta$.

The candidates are fixed before the $n$ validation trajectories are drawn
i.i.d.\ from $\pi$, so each statistic is i.i.d.\ with bound $B$ from
Lemma~\ref{lem:ranges}(b). Lemma~\ref{lem:eb}, applied to $Z$
and $-Z$ with $\delta'=2e^{-\iota}$ (so $\log(2/\delta')=\iota$), gives:
\begin{equation*}
\Pr\big(|\bar Z-\mathbb{E}[Z]|>\xi_\eta(Z)\big)\le2\delta'=4e^{-\iota}.
\end{equation*}
By the union bound over the $2|\Lambda|$ statistics, all intervals hold except
with probability:
\begin{equation*}
2|\Lambda|\cdot4e^{-\iota}=8|\Lambda|\cdot\frac{\delta}{8|\Lambda|}=\delta.
\end{equation*}
On this event:
\begin{equation*}
\bar Z_2-\xi_\eta(Z_2)-\epsilon(\eta)\overset{(a)}{\le}\mathbb{E}[Z_2]-\epsilon(\eta)\overset{(b)}{\le}\Delta_\eta\overset{(b)}{\le}\mathbb{E}[Z_2]+\epsilon(\eta)\overset{(a)}{\le}\bar Z_2+\xi_\eta(Z_2)+\epsilon(\eta),
\end{equation*}
where (a) uses the event and (b) uses Lemma~\ref{lem:ranges}(a). Hence:
\begin{equation}
0\le\Delta_\eta-\big(\bar Z_2-\xi_\eta(Z_2)-\epsilon(\eta)\big)\le2\xi_\eta(Z_2)+2\epsilon(\eta).
\label{eq:pf-cert-gap-z2}
\end{equation}
Likewise, by $\mathbb{E}[Z_{\mathrm{IS}}]=\Delta_\eta$
(Lemma~\ref{lem:ranges}(a)):
\begin{equation}
0\le\Delta_\eta-\big(\bar Z_{\mathrm{IS}}-\xi_\eta(Z_{\mathrm{IS}})\big)\le2\xi_\eta(Z_{\mathrm{IS}}).
\label{eq:pf-cert-gap-is}
\end{equation}

For (i), $L_\eta$ is the larger of the two lower bounds in
\eqref{eq:pf-cert-gap-z2} and \eqref{eq:pf-cert-gap-is}, so
$L_\eta\le\Delta_\eta$ and:
\begin{equation}
\Delta_\eta-L_\eta\le2\min\{\xi_\eta(Z_2)+\epsilon(\eta),\,\xi_\eta(Z_{\mathrm{IS}})\}.
\label{eq:pf-cert-gap}
\end{equation}
Since $L_{\hat\eta}\ge L_0=0$:
\begin{equation}
J(\hat\pi)-J(\pi)=\Delta_{\hat\eta}\ge L_{\hat\eta}\ge0.
\label{eq:pf-cert-gain}
\end{equation}

For (ii), $J(\hat\pi)-J(\pi)\ge0$ by (i), and for every $\eta\in\Lambda$:
\begin{equation*}
J(\hat\pi)-J(\pi)\overset{(a)}{\ge}L_{\hat\eta}\overset{(b)}{\ge}L_\eta\overset{(c)}{\ge}\Delta_\eta-2\min\{\xi_\eta(Z_2)+\epsilon(\eta),\,\xi_\eta(Z_{\mathrm{IS}})\},
\end{equation*}
where (a) uses \eqref{eq:pf-cert-gain}, (b) the definition of $\hat\eta$, and
(c) \eqref{eq:pf-cert-gap}. This is $\Delta_\eta-2c_\eta$; maximizing over
$\eta$ and with $0$ gives (ii).

For (iii), let $Y_1,\dots,Y_n\in[-B,B]$ be the validation values of $Z$, with
mean $\bar Z$, so $Y_\ell^2\le B^2$. Averaging over $\ell$:
\begin{equation}
\frac1n\sum_{\ell=1}^n(Y_\ell-\bar Z)^2=\frac1n\sum_{\ell=1}^nY_\ell^2-\bar Z^2\le B^2,
\qquad\text{so}\qquad\hat\sigma^2\le\frac{n}{n-1}B^2.
\label{eq:pf-cert-var}
\end{equation}
Put $x\coloneqq\sqrt{\iota/(n-1)}$ and assume $n-1>\iota$, so $x<1$ and:
\begin{align}
\xi_\eta(Z)&=\hat\sigma\sqrt{\frac{2\iota}n}+\frac{14B\iota}{3(n-1)}
\overset{(a)}{\le}B\sqrt{\frac{2\iota}{n-1}}+\frac{14}3Bx^2\nonumber\\
&=\sqrt2\,Bx+\frac{14}3Bx^2
\overset{(b)}{\le}\Big(\sqrt2+\frac{14}3\Big)Bx\le8Bx,
\label{eq:pf-cert-radius}
\end{align}
where (a) uses \eqref{eq:pf-cert-var} and (b) uses $x<1$. Since
$|\Delta_\eta|=|\mathbb{E}[Z_{\mathrm{IS}}]|\le B$,
$n-1>256B^2\iota/\Delta_\eta^2\Rightarrow n-1>256\iota>\iota$, so
\eqref{eq:pf-cert-radius} applies with $B=B_{\mathrm{IS},\eta}$ and:
\begin{equation*}
\Delta_\eta-L_\eta\overset{(a)}{\le}2\xi_\eta(Z_{\mathrm{IS}})\overset{(b)}{\le}16B\sqrt{\frac{\iota}{n-1}}\overset{(c)}{<}\Delta_\eta,
\end{equation*}
where (a) uses \eqref{eq:pf-cert-gap}, (b) \eqref{eq:pf-cert-radius}, and (c)
$n-1>256B^2\iota/\Delta_\eta^2$. Hence
$L_{\hat\eta}\ge L_\eta>0=L_0$, so $\hat\eta\neq0$
and $\hat\pi\neq\pi$.
\end{proof}

\begin{corollary}[Every update improves]
\label{cor:rounds}
If round $\kappa$ of Algorithm~\ref{alg:iapo} uses fresh trajectories and
level $\delta/2^\kappa$, then with probability at least $1-\delta$, every
update made by Algorithm~\ref{alg:iapo} strictly improves $J$.
\end{corollary}

\begin{proof}
In round $\kappa$, Theorem~\ref{thm:certificate} at level $\delta/2^\kappa$
gives, with probability at least $1-\delta/2^\kappa$,
$L_\eta>0\Rightarrow\Delta_\eta\ge L_\eta>0$ for every
candidate, and updates require $L_{\hat\eta}>0$. A union bound over
rounds bounds the failure probability by
$\sum_{\kappa\ge1}\delta/2^\kappa=\delta$.
\end{proof}

Theorem~\ref{thm:certificate} needs only candidates fixed before the
validation data (the split $\mathcal{T},\mathcal{V}$ in
Algorithm~\ref{alg:iapo}) and known bounds (Lemma~\ref{lem:ranges}(b)), not a
good optimizer of $Z_2$.

\subsection{A Betting Confidence Bound}
\label{app:betting}

Item (i) of Theorem~\ref{thm:certificate} also holds with $\bar Z-\xi_\eta(Z)$
replaced by a betting lower bound \citep{waudbysmith2024estimating}, often
tighter for skewed statistics such as $Z_{\mathrm{IS}}$. For a statistic
$|Z|\le B$ with validation values $Y_1,\dots,Y_n$, a finite set
$F\subset(0,1]$ of bet sizes, and $z\in(-B,B]$, define:
\begin{equation*}
\mathcal{W}(z)\coloneqq\frac1{|F|}\sum_{\nu\in F}\prod_{\ell=1}^n\Big(1+\nu\Big(\frac{Y_\ell+B}{z+B}-1\Big)\Big),
\end{equation*}
and:
\begin{equation}
\bar Z^{\mathrm{bet}}\coloneqq\sup\Big\{z\in(-B,B]:\ \mathcal{W}(z)\ge\frac{2|\Lambda|}\delta\Big\},
\label{eq:betting}
\end{equation}
with $\bar Z^{\mathrm{bet}}\coloneqq-B$ if the set is empty.

\begin{proposition}[Validity of the betting bound]
\label{prop:betting}
For each statistic, $\Pr(\bar Z^{\mathrm{bet}}>\mathbb{E}[Z])\le\delta/(2|\Lambda|)$.
Hence item (i) of Theorem~\ref{thm:certificate} holds with
$\bar Z^{\mathrm{bet}}$ in place of $\bar Z-\xi_\eta(Z)$.
\end{proposition}

\begin{proof}
Let $\mu_Z\coloneqq\mathbb{E}[Z]$, fix $z\ge\mu_Z$ with $z>-B$, and let
$X_\ell\coloneqq(Y_\ell+B)/(z+B)\ge0$. For $\nu\in(0,1]$:
\begin{equation*}
\mathbb{E}[X_\ell]=\frac{\mu_Z+B}{z+B}\le1,\qquad 1+\nu(X_\ell-1)\ge1-\nu\ge0.
\end{equation*}
Hence, for each $\nu\in F$,
$\big(\prod_{\ell\le\ell'}(1+\nu(X_\ell-1))\big)_{\ell'=0}^n$ is a nonnegative
supermartingale starting at $1$, and so is its average over $F$. By Ville's
inequality \citep{ville1939etude} with threshold $2|\Lambda|/\delta$:
\begin{equation*}
\Pr\Big(\mathcal{W}(z)\ge\frac{2|\Lambda|}\delta\Big)\le\frac\delta{2|\Lambda|}.
\end{equation*}
Each factor is nonincreasing in $z$, so $\mathcal{W}$ is nonincreasing and
$\bar Z^{\mathrm{bet}}>\mu_Z\Rightarrow\exists z>\mu_Z:\mathcal{W}(z)\ge2|\Lambda|/\delta\Rightarrow\mathcal{W}(\mu_Z)\ge2|\Lambda|/\delta$.
If $\mu_Z=-B$, then $Y_\ell=-B$ almost surely, every factor equals $1-\nu$, and
$\mathcal{W}\le1<2|\Lambda|/\delta$. In both cases,
$\Pr(\bar Z^{\mathrm{bet}}>\mu_Z)\le\Pr(\mathcal{W}(\mu_Z)\ge2|\Lambda|/\delta)\le\delta/(2|\Lambda|)$.

By the union bound over the $2|\Lambda|$ statistics, with probability at least
$1-\delta$ the lower sides of
\eqref{eq:pf-cert-gap-z2} and \eqref{eq:pf-cert-gap-is} hold with
$\bar Z^{\mathrm{bet}}$ in place of $\bar Z-\xi_\eta(Z)$, so
$L_\eta\le\Delta_\eta$ for every $\eta$. Items (ii) and (iii) use the
two-sided radius, so they are stated for the empirical Bernstein bound only, and the betting bound is not claimed to satisfy them.
\end{proof}

\subsection{A Stationary Point of the First-Order Objective}
\label{app:trap}

Policies at which $J^{(1)}$ has no improving direction (the second case of
Theorem~\ref{thm:improving}) arise from complementary decisions.

\begin{proposition}[Complementarity]
\label{prop:trap}
Two agents independently choose $1$ with probabilities
$\varphi_1,\varphi_2$, and $J=3\varphi_1\varphi_2-\varphi_1-\varphi_2$. Start
from $(p,p)$ with $p\in(0,\frac13)$, and let both agents move to $p+\alpha$ with
$\alpha\in(0,1-p)$. Then:
\begin{enumerate}
\item[(a)] each single move lowers $J$, and projected gradient flow on
$[0,1]^2$ from $(p,p)$ reaches the strict local maximum $(0,0)$;
\item[(b)] $J^{(1)}=2\alpha(3p-1)<0$ and $J^{(2)}=3\alpha^2$, and the joint
move gains if and only if $3(2p+\alpha)>2$;
\item[(c)] with probability at least $1-\delta$, the sign of $J^{(2)}$ is
identified from $n\ge1/[9\delta p^2(1-p)^2]$ trajectories.
\end{enumerate}
\end{proposition}

\begin{proof}
With $r(1,1)=1$, $r(0,0)=0$, $r(1,0)=r(0,1)=-1$ and independence:
\begin{equation*}
J=\varphi_1\varphi_2\cdot1+\varphi_1(1-\varphi_2)\cdot(-1)+(1-\varphi_1)\varphi_2\cdot(-1)=3\varphi_1\varphi_2-\varphi_1-\varphi_2.
\end{equation*}

For (a), since $p<\frac13$, a single move of agent $1$ gives:
\begin{equation*}
J(p+\alpha,p)-J(p,p)=\big(3(p+\alpha)p-(p+\alpha)-p\big)-\big(3p^2-2p\big)=\alpha(3p-1)<0,
\end{equation*}
and agent $2$ is symmetric. Since $\partial_{\varphi_1}J=3\varphi_2-1$ and
$\partial_{\varphi_2}J=3\varphi_1-1$, the flow from $(p,p)$ stays on
$\varphi_1=\varphi_2=\varphi$ with $\dot\varphi=3\varphi-1<0$ for
$\varphi<\frac13$, so $\varphi$ reaches $0$ in finite time, and the projection
keeps it there since $\partial_{\varphi_i}J=-1<0$ at $(0,0)$. Near $(0,0)$ in $[0,1]^2$,
$J=3\varphi_1\varphi_2-\varphi_1-\varphi_2<0=J(0,0)$ unless
$\varphi_1=\varphi_2=0$, so $(0,0)$ is a strict local maximum of $J$ on $[0,1]^2$.

For (b), a single move changes one decision per trajectory, so by
Corollary~\ref{cor:depth} its gain is $L_i=\alpha(3p-1)$ and
$J^{(1)}=L_1+L_2=2\alpha(3p-1)$. The joint move changes two decisions per
trajectory, so $J^{(1)}+J^{(2)}$ is exact by Corollary~\ref{cor:depth}:
\begin{align*}
J^{(2)}&=J(p+\alpha,p+\alpha)-J(p,p)-2\alpha(3p-1)
=\big(3(p+\alpha)^2-2(p+\alpha)\big)-(3p^2-2p)-6p\alpha+2\alpha=3\alpha^2,\\
J^{(1)}+J^{(2)}&=2\alpha(3p-1)+3\alpha^2=\alpha(6p-2+3\alpha)=\alpha\big(3(2p+\alpha)-2\big),
\end{align*}
so $J^{(1)}+J^{(2)}>0\iff3(2p+\alpha)>2$.

For (c), under $\pi$, $\rho-1=\frac{p+\alpha}p-1=\frac\alpha p$
with probability $p$ and $\rho-1=\frac{1-p-\alpha}{1-p}-1=-\frac\alpha{1-p}$
with probability $1-p$, so:
\begin{equation*}
\mathbb{E}\big[(\rho-1)^2\big]=p\cdot\frac{\alpha^2}{p^2}+(1-p)\cdot\frac{\alpha^2}{(1-p)^2}=\frac{\alpha^2}{p(1-p)}.
\end{equation*}
For $Z\coloneqq(\rho_1-1)(\rho_2-1)G_2$ with $|G_2|\le1$, independence gives
$\mathbb{E}[Z]=J^{(2)}=3\alpha^2$ and:
\begin{equation*}
\mathbb{E}[Z^2]\le\mathbb{E}\big[(\rho_1-1)^2\big]\,\mathbb{E}\big[(\rho_2-1)^2\big]=\frac{\alpha^4}{p^2(1-p)^2}.
\end{equation*}
The sample mean $\bar Z$ of $n$ trajectories satisfies:
\begin{equation*}
\Pr(\bar Z\le0)\le\Pr\big(|\bar Z-3\alpha^2|\ge3\alpha^2\big)\overset{(a)}{\le}\frac{\mathbb{E}[Z^2]}{9n\alpha^4}\le\frac1{9np^2(1-p)^2},
\end{equation*}
where (a) is Chebyshev's inequality, and the right side is at most $\delta$ for
$n\ge1/[9\delta p^2(1-p)^2]$.
\end{proof}

At $p=\frac13$, $J^{(1)}=0$ for every $\alpha$ while $J^{(2)}=3\alpha^2>0$,
the case $\mu_1=0<\mu_2$ of Theorem~\ref{thm:improving}; for an orchestrator
with frozen workers, the same holds for two complementary decisions, with
$I_{0\to0}$ in place of $I_{1\to2}$.
\section{Certified Leaps}
\label{app:select}

Throughout, for $p,q\in(0,1)$, $\mathrm{kl}(p\|q)$ is the Kullback--Leibler
divergence between $\mathrm{Bern}(p)$ and $\mathrm{Bern}(q)$. We use two
standard facts \citep{tsybakov2009introduction}:
\begin{itemize}
\item \emph{chi-square bound}: $\mathrm{kl}(p\|q)\le\chi^2(p\|q)=(p-q)^2/(q(1-q))$;
\item \emph{data processing}: for laws $\mathbb{P},\mathbb{Q}$ of the data and
any event $A$, $\mathrm{kl}(\mathbb{P}(A)\|\mathbb{Q}(A))\le\mathrm{KL}(\mathbb{P}\|\mathbb{Q})$,
and $\mathrm{KL}(\mathbb{P}^{\otimes n}\|\mathbb{Q}^{\otimes n})=n\,\mathrm{KL}(\mathbb{P}\|\mathbb{Q})$;
\end{itemize}
For $p\ge q$, $\mathrm{kl}(p\|q)$ is nondecreasing in $p$ and nonincreasing in
$q$. Hence, if $\mathbb{P}(A)\ge\frac12$ and $\mathbb{Q}(A)\le\delta\le\frac12$:
\begin{equation}
\mathrm{KL}(\mathbb{P}\|\mathbb{Q})\ge\mathrm{kl}\big(\tfrac12\big\|\delta\big)
=\frac12\log\frac1{4\delta(1-\delta)}\ge\frac12\log\frac1{4\delta}.
\label{eq:pf-sel-dp}
\end{equation}

\subsection{Certified Leaps}
\label{app:leap}

Fix a \emph{plan}, a sequence of orchestrator actions, and let:
\begin{equation*}
\nu=\Pr_\pi(\tau\text{ follows the plan}),\qquad
R(\tau)\sim\mathrm{Bern}\big(v+g\,\mathbf{1}\{\tau\text{ follows the plan}\}\big),\qquad
0<v<v+g\le1,
\end{equation*}
with return $G_1=R$. Write $F(\tau)=\mathbf{1}\{\tau\text{ follows the plan}\}$,
so that $J(\pi)=v+g\nu$.

\begin{proposition}[What a trust region cannot certify]
\label{prop:leap}
In this setting:
\begin{enumerate}
\item[(i)] a rule that certifies a gain with probability at most
$\delta\le\frac12$ when $g=0$, and with probability at least $\frac12$ here, needs at least
$v(1-v)\log\frac1{4\delta}/(2\nu g^2)$ trajectories of $\pi$;
\item[(ii)] a step of \eqref{eq:bounded-policy} gains at most
$g\nu\big((1+\eta)^{K+1}-1\big)$;
\item[(iii)] if the actions of the plan are always legal, the policy that
follows it gains $g(1-\nu)$, and for every $\nu<1$ it is certified from $n'$ of
its own trajectories, paired with trajectories of $\pi$ on the same initial
states, once $n'-1>256\log(4/\delta)/(g(1-\nu))^2$.
\end{enumerate}
\end{proposition}

\begin{proof}
For (i), let $\mathbb{P}_g$ be the law of one trajectory of $\pi$ in this
setting and $\mathbb{P}_0$ the law with $g=0$. The two laws share the law of
the actions and states, and differ only in the law of $R$ when $F=1$:
\begin{equation*}
\mathrm{KL}(\mathbb{P}_g\|\mathbb{P}_0)=\nu\,\mathrm{kl}(v+g\|v)\overset{(a)}{\le}\frac{\nu g^2}{v(1-v)},
\end{equation*}
where (a) is the chi-square bound. Let $A$ be the event that the rule
certifies a gain from $n$ trajectories. Then $\mathbb{P}_g^{\otimes n}(A)\ge\frac12$
and $\mathbb{P}_0^{\otimes n}(A)\le\delta$, so by \eqref{eq:pf-sel-dp} and data
processing:
\begin{equation*}
\frac12\log\frac1{4\delta}\le n\,\mathrm{KL}(\mathbb{P}_g\|\mathbb{P}_0)\le\frac{n\nu g^2}{v(1-v)}
\quad\Longrightarrow\quad n\ge\frac{v(1-v)}{2\nu g^2}\log\frac1{4\delta}.
\end{equation*}

For (ii), let $\pi'=\pi'_{\theta,\eta}$ and $\nu'=\Pr_{\pi'}(F=1)$. A trajectory
has at most $K+1$ orchestrator steps, and $\rho_t\le1+\eta$ at each of them by
Proposition~\ref{prop:bounded}. Hence, by Lemma~\ref{lem:reweight}:
\begin{equation*}
J(\pi')-J(\pi)=g(\nu'-\nu)=g\,\mathbb{E}_\pi\Big[F\Big(\prod_{t\le H}\rho_t-1\Big)\Big]
\le g\,\mathbb{E}_\pi\big[F\big((1+\eta)^{K+1}-1\big)\big]=g\nu\big((1+\eta)^{K+1}-1\big).
\end{equation*}

For (iii), the policy $\tilde\pi$ that follows the plan has $F=1$ almost surely,
so $J(\tilde\pi)-J(\pi)=(v+g)-(v+g\nu)=g(1-\nu)$. For $i\le n'$, $\tau_i$ and
$\tilde\tau_i$ share $s_1\sim d_0$, and the pairs are i.i.d. Hence $d_i=G_1(\tilde\tau_i)-G_1(\tau_i)\in[-1,1]$ are i.i.d.\ with:
\begin{equation*}
\mathbb{E}[d_i]=\mathbb{E}_{s_1\sim d_0}\big[\mathbb{E}_{\tilde\pi}[G_1\mid s_1]-\mathbb{E}_\pi[G_1\mid s_1]\big]=J(\tilde\pi)-J(\pi)=g(1-\nu).
\end{equation*}
Let $\iota=\log(4/\delta)$ and let $\xi(d)$ be \eqref{eq:radius} with $B=1$,
$n'$ samples, and $\iota$ in place of $\log(8|\Lambda|/\delta)$. Lemma~\ref{lem:eb}
applied to $d$ and $-d$ with $\delta'=2e^{-\iota}$ gives
$|\bar d-\mathbb{E}[d]|\le\xi(d)$ except with probability $4e^{-\iota}=\delta$. On
this event, $L_{\tilde\pi}=\bar d-\xi(d)\le\mathbb{E}[d]$, and, since
$n'-1>256\iota/(g(1-\nu))^2\ge256\iota>\iota$, \eqref{eq:pf-cert-radius} with
$B=1$ gives:
\begin{equation*}
L_{\tilde\pi}\ge\mathbb{E}[d]-2\xi(d)\ge g(1-\nu)-16\sqrt{\frac{\iota}{n'-1}}>0.
\end{equation*}
As $\nu\to0$, this threshold stays bounded, while the bound in (i) grows as
$1/\nu$.
\end{proof}

A \emph{leap} is a policy $\tilde\pi$ that need not be close to $\pi$. The
set $\tilde\Pi$ of leaps is fixed before $\tau_1,\dots,\tau_n\sim\pi$ are
drawn. For a leap $\tilde\pi$ and $i\le n'\le n$, $\tilde\tau_i\sim\tilde\pi$
starts at the initial state of $\tau_i$, and the certificate of $\tilde\pi$
applies \eqref{eq:radius} to the paired returns:
\begin{equation}
L_{\tilde\pi}=\bar d-\xi(d),\qquad d_i=G_1(\tilde\tau_i)-G_1(\tau_i),
\label{eq:leap}
\end{equation}
with $\bar d$ and $\xi(d)$ over $i\le n'$ and $B=2\bar G\ge|d_i|$. Plans are
chosen by successive halving \citep{karnin2013almost} on separate
trajectories. In practice, a leap plays $\pi$ with small probability, so
that bounded steps from it can still reach every action of $\pi$.

\begin{corollary}[Certified selection with leaps]
\label{cor:leap}
Theorem~\ref{thm:certificate} holds when the leaps are added to the candidates
with the certificates \eqref{eq:leap}, $8|\Lambda|$ is replaced by
$4(2|\Lambda|+|\tilde\Pi|)$ in \eqref{eq:radius} and in item (iii), and
$c_{\tilde\pi}=\xi(d)$ in item (ii).
\end{corollary}

\begin{proof}
Let $\iota=\log(4(2|\Lambda|+|\tilde\Pi|)/\delta)$. The leaps are fixed before
the validation data, and each $d$ is i.i.d.\ over $i\le n'$ with
$\mathbb{E}[d]=J(\tilde\pi)-J(\pi)$, as in the proof of
Proposition~\ref{prop:leap}(iii). As in the proof of
Theorem~\ref{thm:certificate}, each of the $2|\Lambda|+|\tilde\Pi|$ statistics
violates its two-sided interval with probability at most $4e^{-\iota}$, so all
intervals hold except with probability:
\begin{equation*}
(2|\Lambda|+|\tilde\Pi|)\cdot4e^{-\iota}=\delta.
\end{equation*}
On this event, $0\le\big(J(\tilde\pi)-J(\pi)\big)-L_{\tilde\pi}\le2\xi(d)$ for
every leap, which is \eqref{eq:pf-cert-gap-is} with $d$ in place of
$Z_{\mathrm{IS}}$. The proof of Theorem~\ref{thm:certificate} then applies with
the leaps among the candidates.
\end{proof}

\section{Controlled Experiments}
\label{app:exp-controlled}

Each controlled task shares one probability $q$ over $N$ decisions with frozen binary workers, and the reward arrives at the end: Complementarity ($N{=}2$, $p{=}1/3$, reward $3x_0x_1-x_0-x_1$), Redundancy ($N{=}2$, $p{=}0.2$, reward $-x_0x_1+0.3(x_0+x_1)$) and Evidence threshold ($N{=}6$, $p{=}0.3$, reward $\mathbf 1\{\sum_tx_t\ge3\}-0.05\sum_tx_t$). A candidate with step $s\in[-1,1]$ plays $q=p+s\,p(1-p)$, which is \eqref{eq:bounded-policy} with $\eta=|s|$, and all $2^N$ trajectories are enumerated, with 500 repetitions at $\delta=0.05$. On Complementarity the linear term vanishes at $p=\tfrac13$ and a step gains $4s^2/27$, so the first-order score keeps the reference while the second-order score reaches $0.148$. On Redundancy a step gains $0.032s-0.0256s^2$: the first-order score takes $s=1$ and gains $0.0064$, whereas the second-order score takes $s=\tfrac12$ on the grid and gains $0.0096$. On Evidence threshold the truncation bound $\epsilon=1.728$ exceeds the second-order gain $0.423$ at $s=1$, so the order-two certificate rejects that step at every sample size and accepts only a smaller one (gain $0.087$); the importance-sampling certificate accepts the oracle step from $n=4096$ (gain $0.356$). Across the 6{,}000 validation runs, no candidate with negative gain was accepted.

\section{Environments, Roles and Protocol}
\label{app:envs}

\paragraph{Benchmarks.} Each problem is stored as a target unit and a question in AFlow's format \citep{zhang2025aflow}. Its long input shuffles the target unit among $D$ units of other problems of the same split, using a generator seeded by a hash of the problem text, so that the document is a fixed function of the problem. Units stay whole and numbered, which is what the evidence finder retrieves. The setting follows GSM-IC \citep{shi2023large}, carried to document length. The baselines read the whole document, and the team reaches it only through the evidence finder or the generalist; both receive the same question text, including the opening words of the target unit.

\begin{table}[h]
\centering\footnotesize
\caption{The five environments. Prompt tokens are the mean over test of a single call that reads the whole input.}
\label{tab:envs}
\setlength{\tabcolsep}{3.5pt}
\begin{tabular}{@{}lccccc@{}}
\toprule
& GSM8K & MATH & DROP & MBPP & HumanEval \\
\midrule
Target unit & statements & statements & passage & specification & specification \\
Distractors $D$ & 80 & 90 & 30 & 60 & 60 \\
Units per document & 81 & 91 & 31 & 61 & 61 \\
Retrieved units $k$ & 3 & 3 & 2 & 3 & 3 \\
Whole-input prompt tokens & 3.6k & 4.9k & 9.1k & 7.4k & 8.4k \\
\midrule
IAPO training problems & 2{,}000 & 1{,}131 & 1{,}000 & 447 & 577 \\
\quad proposal / validation & 200 / 1{,}800 & 200 / 931 & 200 / 800 & 200 / 247 & 200 / 377 \\
Baseline training problems & 2{,}000 & 1{,}131 & 200 & 447 & 447 (MBPP) \\
Development problems & 252 & 96 & 200 & 100 & 33 \\
Test problems & 790 & 220 & 800 & 333 & 131 \\
\midrule
Orchestrator decisions $K{+}1$ & 5 & 5 & 5 & 4 & 4 \\
Declared plans & 2{,}016 & 2{,}016 & 2{,}016 & 432 & 432 \\
Scorer & last number & MATH equality & token F1 & hidden asserts & \texttt{check} \\
\bottomrule
\end{tabular}
\end{table}

Training problems come from the official training splits and test problems from AFlow's test splits. GSM8K uses the first 2{,}000 official training problems in SHA-256 order and 800 of AFlow's 1{,}055 test problems (790 with a statement before the question). MATH uses 1{,}131 official level 4 to 5 problems of MaAS's four subjects \citep{zhang2025multi} and 336 test problems (220 with at least two sentences). Because AFlow's DROP training split has only 200 questions, IAPO trains on 1{,}000 questions of the official split whose passages are disjoint from AFlow's splits. MBPP keeps the 333 test problems whose first assertion names the function. HumanEval has no training split, so IAPO trains on 577 problems, AFlow's 33 validation problems and 544 MBPP problems rewritten as HumanEval prompts \citep{cassano2023multiple}; none of the test problems is in this training pool.

MATH answers are scored by a matcher that takes the last balanced \verb|\boxed{...}| from prediction and reference and applies MaAS's MATH equality (identical strings, numbers within $10^{-3}$, or a SymPy difference that simplifies to zero). It replaced a matcher that misread some answers, and every MATH arm was rerun with it.

\paragraph{Roles.} Question answering and mathematics use eight roles and code adds Test (Table~\ref{tab:roles}). Teams run in a pipeline regime with typed interfaces: a worker's product reaches only its designated consumer at the next decision, and candidates are combined only when the selector is chosen. The evidence finder ranks the numbered units with BM25 \citep{robertson2009probabilistic}, and Test is MaAS's public-test operator as released, with up to three repairs. Which roles are legal depends on the decision index and a few flags, and at the last decision only stopping is legal, which leaves 2{,}016 open-loop plans at five decisions and 432 at four. The orchestrator observes a MiniLM embedding of the task, the role signals, the previous action, the position and the share of the token budget spent, but never a score.

\begin{table}[h]
\centering\footnotesize
\caption{The roles. ``Right after $X$'' means at the decision immediately following role $X$.}
\label{tab:roles}
\begin{tabular}{@{}l>{\raggedright\arraybackslash}p{0.10\linewidth}>{\raggedright\arraybackslash}p{0.38\linewidth}>{\raggedright\arraybackslash}p{0.32\linewidth}@{}}
\toprule
role & model calls & reads & produces \\
\midrule
Decompose (planner) & 1 & task text only & at most four sub-questions (QA, math) or requirements (code); they only extend the evidence finder's query \\
Extract (evidence finder) & 0 & task text (and the planner's lines right after Decompose); BM25 over the document's units & the top-$k$ units as evidence \\
Reason (analyst / programmer) & 1 & task text; the evidence right after Extract; a solution and its review right after Check (QA, math) & a candidate \\
ReasonAlt (generalist) & 1 & the whole document and the task text & a candidate \\
Calculate & 1 & QA, math: as Reason; code: task text only & QA, math: an arithmetic expression evaluated by a tool; code: a function written from its name alone \\
Check (checker / reviewer) & 1 & task text, the evidence if any, the current solution (code: only the candidate's answer key) & a verdict (supported / unsupported) and review text \\
Test (code only) & 0--3 & the current program, its public tests, task text and evidence & the program after up to three repairs \\
ScEnsemble (selector) & 0 & all candidates & the most common answer key becomes current \\
EarlyStop / FinishEnsemble & 0--1 & -- & submits the current candidate \\
\bottomrule
\end{tabular}
\end{table}

\paragraph{Protocol.} Every call is a chat completion of Qwen2.5-7B-Instruct in bfloat16 on vLLM, one engine per H200, with a 24{,}576-token context window, at most 4{,}096 completion tokens and temperature 1, and both arms run through one metered environment that follows MaAS's protocol. Tokens are the prompt and completion tokens of every request, retries included. IAPO's trajectories are capped at $32{,}768$ tokens and the baselines' at $131{,}072$ unless Appendix~\ref{app:baselines} says otherwise. A trajectory that exhausts its budget or fails with known usage scores $0$ at its paid cost, as in MaAS's evaluator; this happened only on MBPP among IAPO's test runs (at most 3 of 1{,}332 episodes per run). The reward is $s-\lambda N/32{,}768$ for score $s$ and $N$ tokens, with $\lambda=0.076$ (config~A, IAPO), a value fixed in an early pilot so that one extra revision step of about 1{,}500 tokens breaks even, or $\lambda=1$ (CW1, IAPO-Lite).

\paragraph{Baselines.}\label{app:baselines} Table~\ref{tab:baseline-setup} lists how each system was run.

\begin{table*}[t]
\centering\footnotesize
\caption{How each baseline was run. Budget is the trajectory cap in tokens; ``Rep.'' is attempts per test problem.}
\label{tab:baseline-setup}
\setlength{\tabcolsep}{4pt}
\begin{tabular}{@{}l>{\raggedright\arraybackslash}p{0.43\textwidth}>{\raggedright\arraybackslash}p{0.2\textwidth}c>{\raggedright\arraybackslash}p{0.17\textwidth}@{}}
\toprule
System & Setting & Budget & Rep. & Disclosed issue \\
\midrule
Vanilla, CoT & one call with MaAS's solving prompt for the task type & 131k & 4 & -- \\
ComplexCoT & CoT with the 8 most complex training examples & 131k & 4 & not on DROP \\
SC & five CoT calls and MaAS's \texttt{ScEnsemble} & 131k & 4 & invalid ensemble choices \\
MultiPersona & CoT prefixed by the SPP instruction & 131k & 4 & demonstrations omitted \\
LLM-Debate & 3 agents, 2 rounds, majority vote & 131k & 4 & -- \\
LLM-Blender & 5 CoT candidates, PairRM ranking, fusion of the top 3 & 131k & 4 & executor fuses; PairRM unmetered \\
DyLAN & released MATH setting (GSM8K, MATH) or HumanEval setting (code) & 131k; 393k on MBPP, HumanEval, MATH s2 & 4 & not on DROP \\
AgentVerse & upstream \texttt{mgsm} (math) and \texttt{humaneval} (code) tasks & 131k (math); code uncapped & 4 / 1 & 12 of 131 and 17 of 333 code crashes \\
MacNet & Mesh topology, 4 agent nodes; MASLab prompts for math & 131k; 393k on code & 1 & strict parser empties code \\
AutoAgents & upstream explorer, manager and group & 131k & 1 & most problems fail on format \\
GPTSwarm & 4 math solvers or 5 code writers; edges by REINFORCE, 200 iterations & 131k; 393k HumanEval; 33k MBPP & 4 & DROP solver is our adaptation \\
AFlow & upstream optimizer, 7B optimizer, at most 20 rounds & 131k & 4 & few working rounds \\
ADAS & 30 generations from the seed archive & 131k; 33k MBPP & 4 & no agent beat the seeds \\
MaAS & upstream optimizer, 4 layers, threshold $0.3$, 200 training problems & 131k; 33k MBPP & 4 & two evaluator defects \\
MAS-Zero & per-problem design by a 7B meta-agent & $10^{8}$ ledger & 1 & design fails in 76--94\% of iterations \\
\bottomrule
\end{tabular}
\end{table*}

Each system runs as released, adapted only where the shared protocol or the 7B executor forces it, and is then frozen. Systems that search or train read only the benchmark's training split (MBPP's for HumanEval), and their search cost is not counted in tokens per problem. On DROP the baselines' training file, AFlow's validation split, shares 120 of its 161 passages with the test split. For LLM-Blender both the fused output and the top-ranked candidate are kept and the higher is reported. MaAS's solving prompt for MATH reaches the model with \verb|\boxed| as a backspace and ``oxed'', so prompting baselines on MATH run with the backslash restored. The strongest baseline of each environment received a second seed where one was run (Debate on GSM8K and DROP, DyLAN on MATH and MBPP, LLM-Blender on DROP and HumanEval, the benchmarks where it was the strongest).

\section{IAPO Settings, Certificates and Statistics}
\label{app:exp-cert}

All 42 IAPO runs (21 pairs of IAPO and its first-order variant) share the settings of Table~\ref{tab:app-hyper} within each environment family, apart from the draw salt and the interaction switch.

\begin{table}[h]
\centering\footnotesize
\caption{IAPO settings of every reported run.}
\label{tab:app-hyper}
\begin{tabular}{@{}>{\raggedright\arraybackslash}p{0.2\textwidth}>{\raggedright\arraybackslash}p{0.37\textwidth}>{\raggedright\arraybackslash}p{0.37\textwidth}@{}}
\toprule
Setting & QA / math (GSM8K, MATH, DROP) & Code (MBPP, HumanEval) \\
\midrule
Orchestrator decisions $K{+}1$ & 5 & 4 \\
Rounds & 1, from the uniform reference & 1 \\
Proposal sample & 200 problems $\times$ 32 repeats $=6{,}400$ & same \\
Validation sample & 8{,}000 i.i.d.\ draws from the validation pool & same \\
Score model & MLP on the whole observation: Linear($405\!\to\!64$), tanh, Linear($64\!\to\!10$), tanh & MLP on the last 24 (process) features: Linear($24\!\to\!64$), tanh, Linear($64\!\to\!11$), tanh \\
Score range & $f_\theta(s,a)\in[-1,1]$, one output per action, shared across decisions & same \\
Step grid $\Lambda$ & $\{0.25,0.5,1\}$, one fit per step & same \\
Fit & Adam, learning rate 0.01, 400 full-batch steps on $-\overline{Z_2}$ & same \\
Baseline $b$ & mean proposal return & same \\
Certificate per step & $L_\eta=\max\{L(Z_2)-\epsilon(\eta),\,L(Z_{\mathrm{IS}})\}$, betting bounds & same \\
Bet sizes & $\{\frac18,\frac14,\frac12,1\}$ & same \\
Validation baseline & predictable per-problem mean & same \\
Total level $\delta$ & 0.05; round 1 uses $\delta_1=\delta/2=0.025$ & same \\
Leaps $L$, mixing $\varepsilon$ & 3, 0.001 & 3, 0.001 \\
Leap ranking & mean $-$ 2\,SE of $Z_2$ at the plan's leap policy ($J^{(1)}$ in the first-order variant) & same \\
Screen & top $S=50$, 1{,}000 fresh episodes each; the 3 with the largest mean net return become leaps & none; the top 3 by the ranking \\
Leap certification & 4{,}000 episodes per leap, paired with the first 4{,}000 validation draws & same \\
Deploy rule & largest certificate among the reference ($Q=0$), 3 steps and 3 leaps & same \\
Test & deployed policy, every test problem $\times$ 4 & same \\
\bottomrule
\end{tabular}
\end{table}

\subsection{Certificates}
\label{app:exp-cert-details}

With three steps and three leaps there are $M=9$ statistics per run, each at level $\delta_1/9\approx2.8\times10^{-3}$ for $\delta_1=0.025$. A step is certified by $L_\eta=\max\{L(Z_2)-\epsilon(\eta),L(Z_{\mathrm{IS}})\}$ with the betting bound of Appendix~\ref{app:betting}, found by 48 bisection steps and rounded down; the interval lengths and $\epsilon(\eta)$ are in Table~\ref{tab:app-bounds}. A leap is certified by the betting bound on the paired differences $d_i=G_1(\tilde\tau_i)-G_1(\tau_i)$ for $i\le4{,}000$, where $\tilde\tau_i$ starts from the problem of validation draw $\tau_i$ with its own seeds. The deployed policy is the reference, step or leap with the largest certificate, and a candidate with a positive certificate was deployed in all 42 runs. One MBPP run spends about $2.9\times10^{7}$ tokens on proposal trajectories, $3.6\times10^{7}$ on validation and $1.07\times10^{8}$ on the three leaps, about $1.7\times10^{8}$ in total, which Table~\ref{tab:main} does not include.

\begin{table}[h]
\centering\footnotesize
\caption{Known interval lengths $b$ of $Z_2$ and $Z_{\mathrm{IS}}$ (with the per-problem baseline) and the remainder $\epsilon(\eta)$, per family and per cost weight of the objective.}
\label{tab:app-bounds}
\begin{tabular}{@{}ll rrr rrr@{}}
\toprule
& & \multicolumn{3}{c}{config A ($\lambda=0.076$)} & \multicolumn{3}{c}{CW1 ($\lambda=1$)} \\
\cmidrule(lr){3-5}\cmidrule(l){6-8}
Family & $\eta$ & $b(Z_2)$ & $b(Z_{\mathrm{IS}})$ & $\epsilon$ & $b(Z_2)$ & $b(Z_{\mathrm{IS}})$ & $\epsilon$ \\
\midrule
QA / math ($K{+}1=5$) & 0.25 & 4.04 & 4.42 & 0.0208 & 7.50 & 8.21 & 0.0208 \\
 & 0.5 & 10.76 & 14.19 & 0.1768 & 20.00 & 26.38 & 0.1768 \\
 & 1 & 32.28 & 66.71 & 1.5938 & 60.00 & 124.00 & 1.5938 \\
Code ($K{+}1=4$) & 0.25 & 2.96 & 3.10 & 0.0081 & 5.50 & 5.77 & 0.0081 \\
 & 0.5 & 7.53 & 8.74 & 0.0664 & 14.00 & 16.25 & 0.0664 \\
 & 1 & 21.52 & 32.28 & 0.5625 & 40.00 & 60.00 & 0.5625 \\
\bottomrule
\end{tabular}
\end{table}

\paragraph{Per-problem baseline.} A constant baseline removes the mean return but not the spread between problems, so in validation trajectory $i$ we use $b_i\in[g_-,g_+]$, the mean return of the earlier validation trajectories on the same problem, or $b$ if there is none. Because $b_i$ depends only on trajectories $1,\dots,i-1$ and the initial state of trajectory $i$, and $\mathbb E_\pi[\prod_{j\le t}\rho_j-1\mid s_1]=\mathbb E_\pi[\rho_t-1\mid\tau_{t-1},s_t]=0$, the $b_i$ terms have conditional mean zero and each statistic keeps its mean given the earlier trajectories. The payoff then lies in $[g_--g_+,\,g_+-g_-]$, which widens the intervals of Lemma~\ref{lem:ranges}. The values are martingale differences rather than i.i.d.; the proof of Proposition~\ref{prop:betting} needs only that the conditional mean is at most $z$, so it holds with $[z_-,z_+]$ in place of $[-B,B]$, whereas the empirical Bernstein bound would need independence. On competition mathematics with a seven-operator workflow this baseline lowers the standard deviation of $Z_{\mathrm{IS}}$ by about a quarter at every step of the grid.

\paragraph{Leaps.} Every declared plan is scored by the proposal statistic at its leap policy, the plan mixed with $\varepsilon=0.001$ of the reference, as the mean over the $6{,}400$ proposal trajectories minus two standard errors. On question answering and mathematics the 50 top-ranked plans each run on $1{,}000$ fresh validation draws and the three with the largest mean net return become the leaps; on code the three top-ranked plans become the leaps directly. The certificate never reads the screening draws. The expansion serves here as a proposal heuristic for distant plans, so the remainder bound does not apply to the leap ratios, while their certificates stay valid because they use paired draws of their own.

\paragraph{Matched budget.} The frozen deployed policy is rerun on the test split with a fresh salt and 10 to 200 repeats per problem. On answer tasks a vote takes the most common non-empty answer key over the first $K$ episodes, with ties to the earliest; on code the episodes split into groups of $K$, each reduced to one program by MaAS's HumanEval-template selector with its tokens added. $K$ is the largest count whose tokens per vote fit the strongest baseline's tokens per episode, for every listed seed. The budgets are 28.6k (GSM8K), 53.8k (MATH), 57.1k (DROP) and 26.5k tokens (HumanEval); on MBPP it is 35.4k, below DyLAN's 121.9k.

\paragraph{Statistics.} Comparisons are paired by problem on the full test split: per-problem differences of arm means give $\bar d$, $\mathrm{SE}=s_d/\sqrt n$ and $z=\bar d/\mathrm{SE}$, called significant when $z>2$. The ablation is scored on net return $s-\lambda\,\text{tokens}/32{,}768$ with each configuration's own $\lambda$. An arm's score is averaged over its runs per problem before the differences are taken, and no run of a reported configuration is left out.

\section{Full Results}
\label{app:results}

Table~\ref{tab:app-arms} gives one episode of each deployed policy and Table~\ref{tab:app-paired} the paired comparisons. Against every baseline, IAPO's single-episode accuracy on MBPP is higher by at least 8.8 points (best baseline DyLAN, 72.3\%), and on HumanEval, apart from MaAS (83.8\%), by at least 8.1 points (next best Debate, 75.6\%).

\begin{table}[h]
\centering\footnotesize
\caption{Single episode of the deployed policy, mean over runs: accuracy (\%) and tokens per problem (thousands) of IAPO and of its first-order variant (FO).}
\label{tab:app-arms}
\begin{tabular}{@{}lcccc@{\hspace{1.5em}}cccc@{}}
\toprule
& \multicolumn{4}{c}{config A} & \multicolumn{4}{c}{CW1} \\
\cmidrule(lr){2-5}\cmidrule(l){6-9}
& IAPO acc. & tokens & FO acc. & tokens & IAPO acc. & tokens & FO acc. & tokens \\
\midrule
GSM8K & 78.1 & 5.6 & 77.4 & 9.7 & 79.3 & 4.0 & 66.4 & 4.3 \\
MATH & 37.7 & 5.6 & 37.6 & 15.2 & 39.0 & 1.0 & 36.4 & 1.2 \\
DROP & 69.2 & 11.5 & 67.9 & 20.1 & 59.8 & 1.0 & 60.3 & 0.8 \\
MBPP & 81.1 & 5.1 & 63.8 & 15.7 & 78.2 & 1.1 & 75.3 & 1.6 \\
HumanEval & 83.7 & 9.9 & 76.5 & 18.1 & 77.0 & 5.6 & 64.1 & 7.3 \\
\bottomrule
\end{tabular}
\end{table}

\begin{table}[h]
\centering\footnotesize
\caption{Paired comparisons, with $z$ in parentheses and bold for $z>2$. (a) Matched budget against the strongest baseline (mean over runs): accuracy (\%) and tokens per problem, and the difference in points. $^{\ddagger}$Against DyLAN, at the smaller budget of Appendix~\ref{app:exp-cert}. (b) IAPO minus its first-order variant on net return and accuracy.}
\label{tab:app-paired}
\setlength{\tabcolsep}{3.5pt}
\begin{tabular}{@{}llccccc@{}}
\toprule
\multicolumn{7}{@{}l}{\textbf{(a) Matched budget}}\\
& Strongest baseline & \multicolumn{2}{c}{IAPO} & \multicolumn{2}{c}{IAPO-Lite} & \\
\cmidrule(lr){3-4}\cmidrule(lr){5-6}
& acc.\ (tokens) & acc.\ (tokens) & diff. & acc.\ (tokens) & diff. & \\
\midrule
GSM8K & Debate 79.5 (28.6k) & 81.8 (25.1k) & $\mathbf{+2.3}$ (2.2) & 85.1 (25.8k) & $\mathbf{+5.6}$ (5.5) & \\
MATH & DyLAN 48.3 (54.3k) & 48.6 (50.8k) & $+0.3$ (0.2) & 54.3 (51.1k) & $\mathbf{+6.0}$ (2.6) & \\
DROP & Debate 70.6 (57.1k) & 74.2 (46.0k) & $\mathbf{+3.6}$ (3.3) & 66.1 (51.7k) & $-4.5$ ($-2.9$) & \\
MBPP$^{\ddagger}$ & DyLAN 71.2 (121.9k) & 80.9 (33.1k) & $\mathbf{+9.7}$ (6.3) & 78.3 (29.0k) & $\mathbf{+7.2}$ (4.6) & \\
HumanEval & MaAS 83.8 (26.5k) & 84.3 (20.4k) & $+0.5$ (0.4) & 75.7 (21.6k) & $-8.1$ ($-4.4$) & \\
\midrule
\multicolumn{7}{@{}l}{\textbf{(b) Interaction ablation}}\\
& & \multicolumn{2}{c}{IAPO} & \multicolumn{2}{c}{IAPO-Lite} & \\
\cmidrule(lr){3-4}\cmidrule(lr){5-6}
& & net & acc. & net & acc. & \\
\midrule
GSM8K & & $+1.6$ (2.2) & $+0.6$ (0.9) & $\mathbf{+13.9}$ (14.4) & $\mathbf{+13.0}$ (13.4) & \\
MATH & & $+2.3$ (1.8) & $+0.0$ (0.0) & $\mathbf{+3.1}$ (2.5) & $\mathbf{+2.7}$ (2.2) & \\
DROP & & $\mathbf{+3.4}$ (4.5) & $+1.4$ (1.8) & $-1.3$ ($-3.7$) & $-0.5$ ($-1.5$) & \\
MBPP & & $\mathbf{+19.7}$ (12.1) & $\mathbf{+17.3}$ (10.6) & $\mathbf{+4.3}$ (4.7) & $\mathbf{+2.9}$ (3.1) & \\
HumanEval & & $\mathbf{+9.1}$ (5.9) & $\mathbf{+7.2}$ (4.7) & $\mathbf{+18.0}$ (10.2) & $\mathbf{+12.9}$ (7.4) & \\
\bottomrule
\end{tabular}
\end{table}

\begin{figure}[h]
\centering
\includegraphics[width=\textwidth]{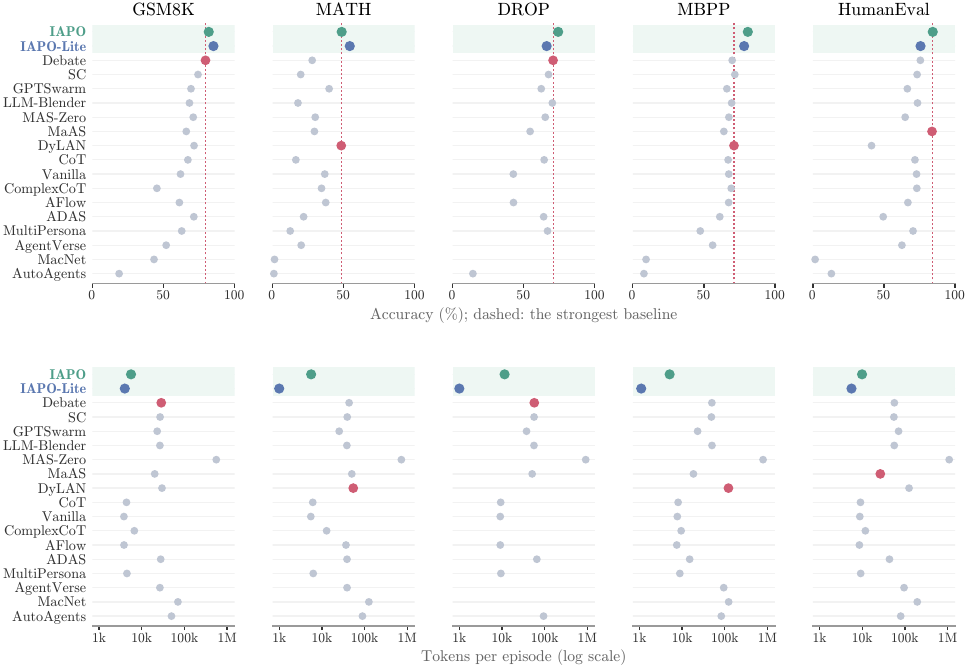}
\caption{Every system on each benchmark. Top: accuracy, with the strongest baseline's accuracy
dashed; bottom: tokens per episode (log scale). IAPO and IAPO-Lite are shaded; as in Figure~\ref{fig:cost},
their accuracy is that of the voted plan within the strongest baseline's budget.}
\label{fig:app-methods}
\end{figure}

\begin{figure}[h]
\centering
\includegraphics[width=\textwidth]{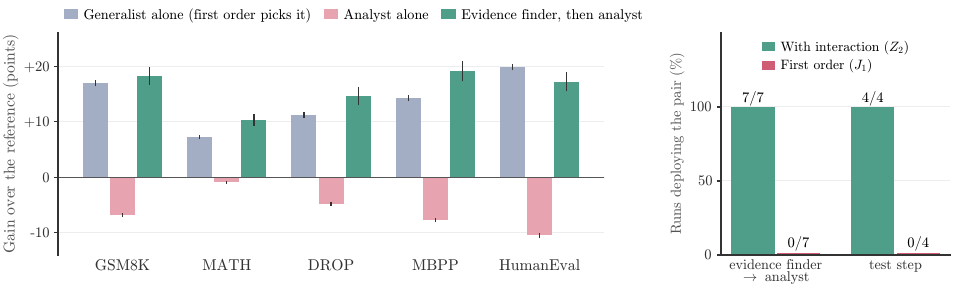}
\caption{Left: gain over the reference of the generalist alone (the choice of the
first-order score), the analyst alone, and the evidence finder followed by the analyst.
Right: runs that deploy the pair, with the interaction term and with first order only, for each benchmark.}
\label{fig:lockout}
\end{figure}

\begin{figure}[h]
\centering
\includegraphics[width=\textwidth]{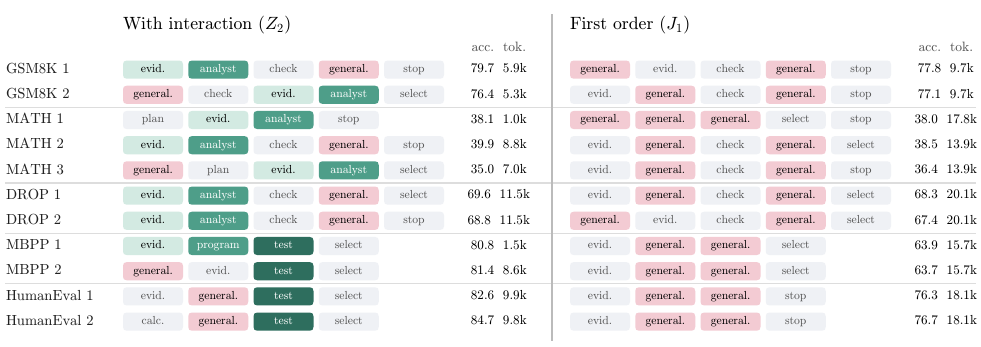}
\caption{Plan deployed by each IAPO run, with the interaction term (left) and with first order only
(right): roles in call order, with accuracy and tokens of a single episode. Each row is a separately
trained run. evid.: evidence finder; general.: second answerer reading the whole document; calc.: calculator; select: selector.}
\label{fig:app-plans}
\end{figure}

\end{document}